\documentclass{tlp}
\usepackage{aopmath}

\usepackage{latexsym}
\usepackage{times}
\usepackage{epsfig}
\usepackage{alltt}

\newcommand{\comment}[1]{}

\long\def\COMMENT#1\ENDCOMMENT{\message{(Commented text...)}\par}

\def\no{{\bf not}\;}

\def\beq{\begin{equation}}
\def\eeq#1{\label{#1}\end{equation}}

\def\oor{\; \emph{or} \; }

\def\calS0{{\cal S}_0}

\newtheorem{definition}{Definition}
\newtheorem{example}{Example}
\newtheorem{condition}{Condition}

\newcommand{\qed}[0]{\hspace*{0mm}\hfill $\Box $\vspace{3mm}}

\def\st{\noindent}

\def\ni{\noindent}

\def\no{ not \;}

\long\def\COMMENT#1\ENDCOMMENT{\message{(Commented text...)}\par}

\begin{document}

\title{Probabilistic reasoning with answer sets}

\author[C. Baral, M. Gelfond and N. Rushton]
{\vspace*{0.1in} Chitta Baral $\dag$, Michael Gelfond $\sharp$, and Nelson Rushton $\sharp$\\ $\dag$ Department of
Computer Science and Engineering,\\ Arizona State University,\\ Tempe, AZ 85287-8809, USA.\\ \vspace*{0.1in} {\it
chitta@asu.edu}\\ $\sharp$ Department of Computer Science \\ Texas Tech University\\ Lubbock, Texas 79409\\ {\it
\{mgelfond,nrushton\}@cs.ttu.edu }}

\bibliographystyle{acmtrans}

%\pagerange{\pageref{firstpage}--\pageref{lastpage}} \volume{\textbf{10} (3):} \jdate{September 2005}
%\setcounter{page}{1} \pubyear{2005}

\submitted {22 September 2005}
\revised {21 June 2007,  20 June 2008}
\accepted {2 December 2008}

\maketitle

\label{firstpage}

\begin{abstract}
{\bf To appear in Theory and Practice of Logic Programming (TPLP)}

\medskip \noindent
This paper develops a declarative language, P-log, that combines logical and probabilistic arguments in its reasoning.
Answer Set Prolog is used as the logical foundation, while causal Bayes nets serve as a probabilistic foundation. We
give several non-trivial examples and illustrate the use of P-log for knowledge representation and updating of
knowledge. We argue that our approach to updates is more appealing than existing approaches. We give sufficiency
conditions for the coherency of P-log programs and show that Bayes nets can be easily mapped to coherent P-log
programs.
\end{abstract}


\section{Introduction}\label{sec1}
The goal of this paper is to define a knowledge representation language allowing natural, elaboration tolerant
representation of commonsense knowledge involving logic and probabilities. The result of this effort is a language
called {\emph P-log}.

\st By a knowledge representation language, or KR language, we mean a formal language $L$ with an entailment relation
$E$ such that (1) statements of $L$ capture the meaning of some class of sentences of natural language, and (2) when a
set $S$ of natural language sentences is translated into a set $T(S)$ of statements of $L$, the formal consequences of
$T(S)$ under $E$ are translations of the informal, commonsense consequences of $S$.

\st One of the best known KR languages is predicate calculus, and this example can be used to illustrate several
points. First, a KR language is committed to an entailment relation, but it is not committed to a particular inference
algorithm. Research on inference mechanisms for predicate calculus, for example, is still ongoing while predicate
calculus itself remains unchanged since the 1920's.

\st Second, the merit of a KR language is partly determined by the class of statements representable in it. Inference in
predicate calculus, e.g., is very expensive, but it is an important language because of its ability to formalize a broad
class of natural language statements, arguably including mathematical discourse.

\st Though representation of mathematical discourse is a problem solved to the satisfaction of many, representation of
other kinds of discourse remains an area of active research, including work on defaults, modal reasoning, temporal
reasoning, and varying degrees of certainty.

\st Answer Set Prolog (ASP) is a successful KR language with a large history of literature and an active community of
researchers. In the last decade ASP was shown to be a powerful tool capable of representing recursive definitions,
defaults, causal relations, special forms of self-reference, and other language constructs which occur frequently in
various non-mathematical domains \cite{baral03:book}, and are difficult or impossible to express in classical logic and
other common formalisms. ASP is based on the answer set/stable models semantics \cite{gel88} of logic programs with
default negation (commonly written as $\no$), and has its roots in research on non-monotonic logics. In addition to the
default negation the language contains ``classical'' or ``strong'' negation (commonly written as $\neg$) and ``epistemic disjunction'' (commonly written as \emph{or}).

\st Syntactically, an ASP program is a collection of rules of the form:
$$l_0 \oor \dots \oor l_k \leftarrow l_{k+1},\dots,l_m,\no l_{m+1},\dots,\no l_n$$
where $l$'s are literals, i.e. expressions of the form $p$ and $\neg p$ where $p$ is an atom. A rule with variables is
viewed as a schema - a shorthand notation for the set of its ground instantiations. Informally, a ground program $\Pi$
can be viewed as a specification for the sets of beliefs which could be held by a rational reasoner associated with
$\Pi$. Such sets are referred to as \emph{answer sets}. An answer set is represented by a collection of ground literals. In forming answer sets the reasoner must be guided by the following informal principles:

\st 1. One should satisfy the rules of $\Pi$. In other words, if one believes in the body of a rule, one must also
believe in its head.

\st 2. One should not believe in contradictions.

\st 3. One should adhere to the rationality principle, which says: ``Believe nothing you are not
forced to believe.''

\st An answer set $S$ of a program satisfies a literal $l$ if $l \in S$; $S$ satisfies $\no l$ if $l \not\in S$; $S$
satisfies a disjunction if it satisfies at least one of its members. We often say that if $p \in S$ then $p$ is
\emph{believed to be true} in $S$, if $\neg p \in S$ then $p$ is \emph{believed to be false} in $S$. Otherwise $p$ is
\emph{unknown} in $S$. Consider, for instance, an ASP program $P_1$ consisting of rules:\\

\st 1. $p(a)$.\\ 2. $\neg p(b)$.\\ 3. $q(c) \leftarrow \no p(c), \no \neg p(c)$.\\ 4. $\neg q(c) \leftarrow p(c)$.\\ 5.
$\neg q(c) \leftarrow \neg p(c)$.\\

\st The first two rules of the program tell the agent associated with $P_1$ that he must believe that $p(a)$ is true and
$p(b)$ is false. The third rule tells the agent to believe $q(c)$ if he believes neither truth nor falsity of $p(c)$.
Since the agent has reason to believe neither truth nor falsity of $p(c)$ he must believe $q(c)$. The last two rules
require the agent to include $\neg q(c)$ in an answer set if this answer set contains either $p(c)$ or $\neg p(c)$.
Since there is no reason for either of these conditions to be satisfied, the program will have unique answer set $S_0 =
\{p(a),\neg p(b), q(c)\}$. As expected the agent believes that $p(a)$ and $q(c)$ are true and that $p(b)$ is false,
and simply does not consider truth or falsity of $p(c)$.

\st If $P_1$ were expanded by another rule:\\

\st 6. $p(c) \oor \neg p(c)$\\

\st the agent will have two possible sets of beliefs represented by answer sets $S_1 = \{p(a),\neg p(b), p(c),\neg
q(c)\}$ and $S_2 = \{p(a),\neg p(b), \neg p(c),\neg q(c)\}$.

\st Now $p(c)$ is not ignored. Instead the agent considers two possible answer sets, one containing $p(c)$ and another
containing $\neg p(c)$. Both, of course, contain $\neg q(c)$.

\st The example illustrates that the disjunction (6), read as ``believe $p(c)$ to be true or believe $p(c)$ to be
false'', is certainly not a tautology. It is often called the \emph{awareness axiom} (for $p(c)$). The axiom prohibits
the agent from removing truth of falsity of $p(c)$ from consideration. Instead it forces him to consider the
consequences of believing $p(c)$ to be true as well as the consequences of believing it to be false.

\st The above intuition about the meaning of logical connectives of ASP\footnote{It should be noted that the connectives of Answer Set Prolog are different from those of Propositional Logic.} and that of the rationality principle is
formalized in the definition of an answer set of a logic program (see Appendix III). There is a substantial amount of
literature on the methodology of using the language of ASP for representing various types of (possibly incomplete)
knowledge \cite{baral03:book}.

\st There are by now a large number of inference engines designed for various subclasses of ASP programs. For example, a number of recently developed systems, called \emph{answer set solvers},
\cite{niemela97,simons02,dlvsystem,leone06,lierler05,lz04,gebser07} compute answer sets of logic programs with finite
Herbrand universes. \emph{Answer set programming}, a programming methodology which consists in reducing a
computational problem to computing answer sets of a program associated with it, has been successfully applied to
solutions of various classical AI and CS tasks including planning, diagnostics, and configuration \cite{baral03:book}.
As a second example, more traditional query-answering algorithms of logic programming including SLDNF based Prolog
interpreter and its variants \cite{apt94z,csw95} are sound with respect to stable model semantics of programs without $\neg$ and \emph{or}.

\st However, ASP recognizes only three truth values: true, false, and unknown. This paper discusses an augmentation of
ASP with constructs for representing varying degrees of belief. The objective of the resulting language is
to allow elaboration tolerant representation of commonsense knowledge involving logic and probabilities. P-log was first
introduced in \cite{baral04:lpnmr1}, but much of the material here is new, as discussed in the concluding section of
this paper.

\st A prototype implementation of P-log exists and has been used in promising experiments comparing its performance with
existing approaches \cite{gel06b}. However, the focus of this paper is not on algorithms, but on precise declarative
semantics for P-log, basic mathematical properties of the language, and illustrations of its use. Such semantics are
prerequisite for serious research in algorithms related to the language, because they give a definition with respect to
which correctness of algorithms can be judged. As a declarative language, P-log stands ready to borrow and combine
existing and future algorithms from fields such as answer set programming, satisfiability solvers, and Bayesian
networks.

\st {\em P-log} extends ASP by adding probabilistic constructs, where probabilities are understood as a measure of the
degree of an agent's belief. This extension is natural because the intuitive semantics of an ASP program is given in
terms of the beliefs of a rational agent associated with it. In addition to the usual ASP statements, the P-log
programmer may declare ``random attributes'' (essentially random variables) of the form $a(X)$ where $X$ and the value
of $a(X)$ range over \emph{finite domains}. Probabilistic information about possible values of $a$ is given through {\em
causal probability atoms}, or $pr$-atoms. A $pr$-atom takes roughly the form
$$pr_r(a(t)=y |_c\ B) = v$$
where $a(t)$ is a random attribute, $B$ a set of literals, and $v \in [0,1]$. The statement says that {\em if the value
of $a(t)$ is fixed by experiment $r$, and $B$ holds, then the probability that $r$ causes $a(\overline{t})=y$ is $v$.}

\st A P-log program consists of its {\emph logical part} and its {\emph probabilistic part}. The logical part represents
knowledge which determines the possible worlds of the program, including ASP rules and declarations of random
attributes, while the probabilistic part contains pr-atoms which determine the probabilities of those worlds. If $\Pi$
is a P-log program, the semantics of P-log associates the logical part of $\Pi$ with a ``pure'' ASP program $\tau(\Pi)$.
The semantics of a ground $\Pi$ is then given by

\st (i) a collection of answer sets of $\tau(\Pi)$ viewed as the possible sets of beliefs of a rational agent associated
with $\Pi$, and

\st (ii) a measure over the possible worlds defined by the collection of the probability atoms of $\Pi$ and the
\emph{principle of indifference} which says that possible values of random attribute $a$ are assumed to be equally
probable if we have no reason to prefer one of them to any other.

\st As a simple example, consider the program

\st $a : \{1,2,3\}$.\\ $random(a)$.\\ $pr(a=1) = 1/2$.

\st This program defines a random attribute $a$ with possible values $1,2$, and $3$. The program's possible worlds are
$W_1 = \{a=1\}$, $W_2 = \{a=2\}$, and $W_3 = \{a=3\}$. In accordance with the probability atom of the program, the
probability measure $\mu(W_1) = 1/2$. By the principle of indifference $\mu(W_2) = \mu(W_3) = 1/4$.

\st This paper is concerned with defining the syntax and semantics of P-log, and a methodology of its use for knowledge
representation. Whereas much of the current research in probabilistic logical languages focuses on learning, our main
purpose, by contrast, is to elegantly and straightforwardly represent knowledge requiring subtle logical and
probabilistic reasoning. A limitation of the current version of P-log is that we limit the discussion to models with finite Herbrand domains. This is common for ASP and its extensions. A related limitation prohibits programs containing infinite number of random selections (and hence an uncountable number of possible worlds). This means P-log cannot be used, for example, to describe stochastic processes whose time domains are infinite. However, P-log can be used to describe initial finite segments of such processes, and this paper gives two small examples of such descriptions (Sections~\ref{robot} and ~\ref{sec-squirrel}) and discusses one large example in Section~\ref{shuttleRCS}. We believe the techniques used by \cite{sato95} can be used to extend the semantics of P-log to account for programs with infinite Herbrand domains. The resulting language would, of course, allow representation of processes with infinite time domains. Even though such extension is theoretically not difficult, its implementation requires further research in ASP solvers. This matter is a subject of future work. In this paper we do not emphasize P-log inference algorithms even for programs
with finite Herbrand domains, though this is also an obvious topic for future work. However, our prototype implementation of P-log, based on an answer set solver Smodels \cite{niemela97}, already works rather efficiently for programs with large and complex logical component and a comparatively small number of random attributes.

\st The existing implementation of P-log was successfully used for instance in an industrial size application for
diagnosing faults in the reactive control system (RCS) of the space shuttle \cite{bal01,bgnw02}. The RCS is the Shuttle's system that has primary responsibility for maneuvering the aircraft while it is in space. It consists of fuel and oxidizer tanks, valves, and other plumbing needed to provide propellant to the maneuvering jets of the Shuttle. It also includes electronic circuitry: both to control the valves in the fuel lines and to prepare the jets to receive firing commands. Overall, the system is rather complex, in that it includes $12$ tanks, $44$ jets, $66$ valves, $33$ switches, and around $160$ computer commands (computer-generated signals).

\st We believe that P-log has some distinctive features which can be of interest to those who use probabilities. First,
P-log probabilities are defined by their relation to a knowledge base, represented in the form of a P-log program. Hence
we give an account of the relationship between probabilistic models and the background knowledge on which they are
based. Second, P-log gives a natural account of how degrees of belief change with the addition of new knowledge. For
example, the standard definition of conditional probability in our framework becomes a \emph{theorem}, relating degrees
of belief computed from two different knowledge bases, in the special case where one knowledge base is obtained from the
other by the addition of observations which eliminate possible worlds. Moreover, P-log can accommodate updates which add
rules to a knowledge base, including defaults and rules introducing new terms.

\st Another important feature of P-log is its ability to distinguish between conditioning on observations and on
deliberate actions. The distinction was first explicated in \cite{pearl99b}, where, among other things, the author
discusses relevance of the distinction to answering questions about desirability of various actions (Simpson paradox
discussed in section \ref{sp} gives a specific example of such a situation). In Pearl's approach the effect of a
deliberate action is modeled by an operation on a graph representing causal relations between random variables of a
domain. In our approach, the semantics of
conditioning on actions is axiomatized using ASP's default negation, and these axioms are included as part of the
translation of programs from P-log to ASP. Because Pearl's theory of causal Bayesian nets (CBN's) acts as the
probabilistic foundation of P-log, CBN's are defined precisely in Appendix II, where it is shown that each CBN maps in a
natural way to a P-log program.

\st The last characteristic feature of P-log we would like to mention here is its \emph{probabilistic non-monotonicity}
--- that is, the ability of the reasoner to change his probabilistic model as a result of new information. Normally any solution
of a probabilistic problem starts with construction of probabilistic model of a domain. The model consists of a
collection of possible worlds and the corresponding probability measure, which together determine the degrees of the
reasoner's beliefs. In most approaches to probability, new information can cause a reasoner to abandon some of his
possible worlds. Hence, the effect of update is monotonic, i.e. it can only eliminate possible worlds. Formalisms in
which an update can cause creation of new possible worlds are called ``probabilistically non-monotonic''. We claim that
non-monotonic probabilistic systems such as P-log can nicely capture changes in the reasoner's probabilistic models.

\st To clarify the argument let us informally consider the following P-log program (a more elaborate example involving
 a Moving Robot will be given in Section \ref{robot}).

\st $a : \{1,2,3\}$.\\ $a=1 \leftarrow \no abnormal$.\\ $random(a) \leftarrow abnormal$.

\st Here $a$ is an attribute with possible values $1$, $2$, and $3$. The second rule of the program says that normally
the value of $a$ is $1$. The third rule tells us that under abnormal circumstances $a$ will randomly take on one of its
possible values. Since the program contains no atom $abnormal$ the second rule concludes $a=1$. This is the only
possible world of the program, $\mu(a=1) =1$, and hence the value of $a$ is $1$ with probability $1$. Suppose, however,
that the program is expanded by an atom $abnormal$. This time the second rule is not applicable, and the program has
three possible worlds: $W_1 = \{a=1\}$, $W_2 = \{a=2\}$, and $W_3 = \{a=3\}$. By the principle of indifference $\mu(W_1)
= \mu(W_2) = \mu(W_3) = 1/3$ -- attribute $a$ takes on value $1$ with probability $1/3$.

\st The rest of the paper is organized as follows. In Section~\ref{sec2} we give the syntax of P-log and in
Section~\ref{semantics-sec} we give its semantics. In Section~\ref{sec4} we discuss updates of P-log programs.
Section~\ref{kr-sec} contains a number of examples of the use of P-log for knowledge representation and reasoning. The
emphasis here is on demonstrating the power of P-log and the methodology of its use. In Section~\ref{property-sec} we
present sufficiency conditions for consistency of P-log programs and use it to show how Bayes nets are special cases of
consistent P-log programs. Section~\ref{other_work} contains a discussion of the relationship between P-log and other
languages combining probability and logic programming. Section 8 discusses conclusions and future work. Appendix I
contains the proofs of the major theorems, and appendix II contains background material on causal Bayesian networks.
Appendix III contains the definition and a short discussion of the notion of an answer set of a logic program.

\section{Syntax of P-log} \label{sec2}
A {\em probabilistic logic program} (P-log program) $\Pi$ consists of (i) a {\em sorted signature}, (ii) a {\em
declaration}, (iii) a {\em regular part}, (iv) a set of {\em random selection rules}, (v) a {\em probabilistic
information} part, and (vi) a set of {\em observations} and {\em actions}. Every statement of P-log must be ended by a
period.

\st {\bf (i)\ Sorted Signature}: The sorted signature $\Sigma$ of $\Pi$ contains a set $O$ of objects and a set $F$ of
function symbols. The set $F$ is a union of two disjoint sets, $F_r$ and $F_a$. Elements of $F_r$ are called {\em term
building functions}. Elements of $F_a$ are called {\em attributes}.

\st Terms of P-log are formed in a usual manner using function symbols from $F_r$ and objects from $O$. Expressions of
the form $a(\overline{t})$, where $a$ is an attribute and $\overline{t}$ is a vector of terms of the sorts required by
$a$, will be referred to as {\em attribute terms}. (Note that attribute terms are not terms). Attributes with the range
$\{true,false\}$ are referred to as Boolean attributes or {\em relations}. {\em We assume that the number of terms and attributes over $\Sigma$ is finite}. Note that, since our signature is sorted, this does not preclude the use of function symbols. The example in Section~\ref{shuttleRCS} illustrates such a use.

\st Atomic statements are of the form $a(\overline{t})=t_0$, where $t_0$ is a term, $\overline{t}$ is a vector of terms, and $a$ is an attribute (we assume that $t$ and $\overline{t}$ are of the sorts required by $a$). An atomic statement,
$p$, or its negation, $\neg p$ is referred to as a {\em literal} (or $\Sigma$-literal, if $\Sigma$ needs to be emphasized); literals $p$ and $\neg p$ are called {\em contrary}; by $\overline{l}$ we denote the literal contrary to $l$; expressions $l$ and $\no l$ where $l$ is a literal and $\no$ is the default negation of Answer Set Prolog are called {\em extended literals}. Literals of the form $a(\overline{t})=true$, $a(\overline{t})=false$, and $\neg (a(\overline{t})=t_0)$ are often written as $a(\overline{t})$, $\neg a(\overline{t})$, and $a(\overline{t})\not=t_0$ respectively. If $p$ is a unary relation and $X$ is a variable then an expression of the form $\{X : p(X)\}$ will be called a {\em set-term}. Occurrences of $X$ in such an expression are referred to as {\em bound}.

\st Terms and literals are normally denoted by (possibly indexed) letters $t$ and $l$ respectively. The letters $c$
and $a$, possibly with indices, are used as generic names for sorts and attributes. Other lower case letters denote
objects. Capital letters normally stand for variables.

\st Similar to Answer Set Prolog, a P-log statement containing unbound variables is considered a shorthand for the set
of its ground instances, where a ground instance is obtained by replacing unbound occurrences of variables with properly
sorted ground terms. Sorts in a program are indicated by the declarations of attributes (see below). In defining
semantics of our language we limit our attention to finite programs with no unbound occurrences of variables. We
sometimes refer to programs without unbound occurrences of variables as {\em ground}.

\st {\bf (ii)\ Declaration}: The declaration of a P-log program is a collection of definitions of sorts and sort
declarations for attributes.

\st A sort $c$ can be defined by explicitly listing its elements,
\begin{equation}\label{e-1}
c = \{x_1, \ldots, x_n\}.
\end{equation}
or by a logic program $T$ with a unique answer set $A$. In the latter case $x \in c$ iff $c(x) \in A$.

\st The domain and range of an attribute $a$ are given by a statement of the form:
\begin{equation}\label{e0}
a : c_1 \times \dots \times c_n \rightarrow c_0.
\end{equation}
For attributes without parameters we simply write $a:c_0$.

\st The following example will be used throughout this section.
\begin{example}\label{dice1}[Dice Example: program component $D_1$]\\
{\rm Consider a domain containing two dice owned by Mike and John respectively. Each of the dice will be rolled once. A
P-log program $\Pi_0$ modeling the domain will have a signature $\Sigma$ containing the names of the two dice, $d_1$ and
$d_2$, an attribute $roll$ mapping each die to the value it indicates when thrown, which is an integer from $1$ to $6$,
an attribute $owner$ mapping each die to a person, relation $even(D)$, where $D$ ranges over $dice$, and ``imported'' or
``predefined'' arithmetic functions $+$ and $mod$. The corresponding declarations, $D_1$, will be as follows:

\st $dice = \{d_1,d_2\}.$\\ $score = \{1, 2, 3, 4, 5, 6\}.$\\ $person = \{mike,john\}.$\\ $roll : dice \rightarrow
score.$\\ $owner : dice \rightarrow person.$\\ $even : dice \rightarrow Boolean.$ } \hfill $\Box$
\end{example}

\st {\bf (iii)\ Regular part}: The regular part of a P-log program consists of a collection of rules of Answer Set
Prolog (without disjunction) formed using literals of $\Sigma$.
\begin{example}\label{dice1a}[Dice Example (continued): program
component $D_2$]\\ {\rm For instance, the regular part $D_2$ of program $\Pi_0$ may contain the following rules:

\st $owner(d_1) = mike.$\\ $owner(d_2) = john.$\\ $even(D) \leftarrow roll(D) = Y, Y \ mod \ 2 = 0.$\\ $\neg even(D)
\leftarrow \no even(D).$

\st Here $D$ and $Y$ range over $dice$ and $score$ respectively. } \hfill $\Box$
\end{example}

\st {\bf (iv)\ Random Selection}: This section contains rules describing possible values of random attributes. More
precisely a {\em random selection} is a rule of the form
\begin{equation}\label{e1}
[\ r \ ]\ random(a(\overline{t}): \{X : p(X)\}) \leftarrow B.
\end{equation}
where $r$ is a term used to name the rule and $B$ is a collection of extended literals of $\Sigma$. The name $[\ r\ ]$
is optional and can be omitted if the program contains exactly one random selection for $a(\overline{t})$. Sometimes we
refer to $r$ as an \emph{experiment}. Statement (\ref{e1}) says that {\em if $B$ holds, the value of $a(\overline{t})$
is selected at random from the set $\{X : p(X)\} \cap range(a)$ by experiment $r$, unless this value is fixed by a
deliberate action}. If $B$ in (\ref{e1}) is empty we simply write
\begin{equation}\label{e1a}
[\ r \ ]\ random(a(\overline{t}) : \{X : p(X)\}).
\end{equation}
If $\{X : p(X)\}$ is equal to the $range(a)$ then rule (\ref{e1}) may be written as
\begin{equation}\label{e1a1}
[\ r \ ] \ random(a(\overline{t})) \leftarrow B.
\end{equation}
Sometimes we refer to the attribute term $a(\overline{t})$ as {\em random} and to $\{X : p(X)\} \cap range(a)$ as the
{\em dynamic range} of $a(\overline{t})$ via rule $r$. We also say that a literal $a(\overline{t}) = y$ {\em occurs in
the head} of (\ref{e1}) for every $y \in range(a)$, and that any ground instance of $p(X)$ and literals occurring in $B$
{\em occur in the body} of (\ref{e1}).

\begin{example}\label{dice1b}[Dice Example (continued)]\\
{\rm The fact that values of attribute $roll : dice \rightarrow score$ are random is expressed by the statement

\st $[\ r(D) \ ] \ random(roll(D))$. } \hfill $\Box$
\end{example}

\st {\bf (v)\ Probabilistic Information:} Information about probabilities of random attributes taking particular values
is given by {\em probability atoms} (or simply {\em pr-atoms}) which have the form:
\begin{equation}\label{e2}
pr_r(a(\overline{t})=y \ |_c \ B) = v.
\end{equation}
where $v \in [0,1]$, $B$ is a collections of extended literals, $pr$ is a special symbol not belonging to $\Sigma$, $r$
is the name of a random selection rule for $a(\overline{t})$, and $pr_r(a(\overline{t})=y \ |_c \ B)=v$ says that {\em
if the value of $a(\overline{t})$ is fixed by experiment $r$, and $B$ holds, then the probability that $r$ causes
$a(\overline{t})=y$ is $v$.} (Note that here we use `cause' in the sense that $B$ is an immediate or proximate cause of
$a(\overline{t})=y$, as opposed to an indirect cause.) If $W$ is a possible world of a program containing (\ref{e2}) and $W$ satisfies both $B$ and the body of rule $r$, then we will refer to $v$ as the {\em causal probability} of the atom
$a(\overline{t})=y $ in $W$.

\st
We say that a literal $a(\overline{t}) = y$ {\em occurs in the head} of (\ref{e2}), and that literals occurring in $B$ {\em occur in the body} of (\ref{e2}).

\st If $B$ is empty we simply write
\begin{equation}\label{e2a}
pr_r(a(\overline{t})=y) = v.
\end{equation}
If the program contains exactly one rule generating values of $a(\overline{t})=y$ the index $r$ may be omitted.
\begin{example}\label{dice1c}[Dice Example (continued): program component $D_3$]\\
{\rm For instance, the dice domain may include $D_3$ consisting of the random declaration of
$roll(D)$ given in Example \ref{dice1b} and the following probability atoms:

\st $pr(roll(D)=Y \ |_c \ owner(D) = john) = 1/6.$\\ $pr(roll(D)=6 \ |_c \ owner(D) = mike) = 1/4$.\\ $pr(roll(D)=Y \
|_c \ Y \not= 6, owner(D) = mike) = 3/20$.

\st The above probability atoms convey that the die owned by John is fair, while the die owned by Mike is biased to roll $6$ at a probability of $.25$. } \hfill $\Box$
\end{example}

\st {\bf (vi)\ Observations and actions}: Observations and actions are statements of the respective forms
$$obs(l). \ \ \ \ \ \ \ \ \ \ \ \ do(a(\overline{t})=y)).$$
where $l$ is a literal. Observations are used to record the outcomes of random events, i.e., random attributes, and
attributes dependent on them. The dice domain may, for instance, contain $\{obs(roll(d_1)=4)\}$ recording the outcome of rolling die $d_1$. The statement $do(a(\overline{t})=y)$ indicates that $a(\overline{t})=y$ is made true as a result of
a deliberate (non-random) action. For instance, $\{do(roll(d_1)=4)\}$ may indicate that $d_1$ was simply put on the
table in the described position. Similarly, we may have $obs(even(d_1))$. Here, even though $even(d_1)$ is not a random
attribute, it is dependent on the random attribute $roll(d_1)$. If $B$ is a collection of literals $obs(B)$ denotes the
set $\{obs(l) \ | \ l \in B\}$. Similarly for $do$.

\st The precise meaning of $do$ and $obs$ is captured by axioms (\ref{intervene1} -- \ref{action-eq}) in the next section and discussed in Example \ref{rat}, and in connection with Simpson's Paradox in section \ref{sp}. More discussion of the difference between actions and observations in the context of probabilistic reasoning can be found in \cite{pearl99b}.

\noindent Note that limiting observable formulas to literals is not essential. It is caused by the syntactic restriction of Answer Set Prolog which prohibits the use of arbitrary formulas. The restriction could be lifted if instead of Answer Set Prolog we were to consider, say, its dialect from \cite{ltt99}. For the sake of simplicity we decided to stay with
the original definition of Answer Set Prolog.

\st A P-log program $\Pi$ can be viewed as consisting of two parts. The {\em logical part}, which is formed by
declarations, regular rules, random selections, actions and observations, defines possible worlds of $\Pi$. The {\em
probabilistic part} consisting of probability atoms defines a measure over the possible worlds, and hence defines the
probabilities of formulas. (If no probabilistic information on the number of possible values of a random attribute is
available we assume that all these values are equally probable).

\section{Semantics of P-log} \label{semantics-sec}
The semantics of a ground P-log program $\Pi$ is given by a collection of the possible sets of beliefs of a rational
agent associated with $\Pi$, together with their probabilities. We refer to these sets as possible worlds of $\Pi$. We
will define the semantics in two stages. First we will define a mapping of the logical part of $\Pi$ into its Answer Set Prolog counterpart, $\tau(\Pi)$. The answer sets of $\tau(\Pi)$ will play the role of possible worlds of $\Pi$. Next we
will use the probabilistic part of $\Pi$ to define a measure over the possible worlds, and the probabilities of
formulas.

\subsection{Defining possible worlds:} \label{def-pos-worlds}
The logical part of a P-log program $\Pi$ is translated into an Answer Set Prolog program $\tau(\Pi)$ in the following
way.
\begin{enumerate}

\item Sort declarations: For every sort declaration $c = \{x_1,\dots,x_n\}$ of $\Pi$, $\tau(\Pi)$ contains $c(x_1),
    \dots, c(x_n)$.

\st For all sorts that are defined using an Answer Set Prolog program $T$ in $\Pi$, $\tau(\Pi)$ contains $T$.

\st \item Regular part:

\st In what follows (possibly indexed) variables $Y$ are free variables. A rule containing these variables will be
viewed as shorthand for a collection of its ground instances with respect to the appropriate typing.

\begin{enumerate}

\item For each rule $r$ in the regular part of $\Pi$, $\tau(\Pi)$ contains the rule obtained by replacing each
    occurrence of an atom $a(\overline{t}) = y$ in $r$ by $a(\overline{t},y)$.

\item For each attribute term $a(\overline{t})$, $\tau(\Pi)$ contains the rule:
\begin{equation}\label{r0}
\neg a(\overline{t},Y_1) \leftarrow a(\overline{t},Y_2), Y_1\not=Y_2.
\end{equation}
\end{enumerate}
which guarantees that in each answer set $a(\overline{t})$ has at most one value.

\st \item Random selections:

\begin{enumerate}

\item For an attribute $a$, we have the rule:
\begin{equation} \label{intervene1}
intervene(a(\overline{t})) \leftarrow do(a(\overline{t},Y)).
\end{equation}
Intuitively, $intervene(a(\overline{t}))$ means that the value of $a(\overline{t})$ is fixed by a deliberate
action. Semantically, $a(\overline{t})$ will not be considered random in possible worlds which satisfy
$intervene(a(\overline{t}))$.

\item Each random selection rule of the form
$$[\ r \ ]\ random(a(\overline{t}): \{Z : p(Z)\}) \leftarrow B.$$
with $range(a) = \{y_1,\dots,y_k\}$ is translated to the following rules in Answer Set Prolog\footnote{Our P-log
implementation uses an equivalent rule $1\{a(\overline{t},Z) : c_0(Z) : p(Z)\}1 \leftarrow B, \no
intervene(a(\overline{t}))$ from the input language of Smodels.}
\begin{equation}\label{r1.1}
a(\overline{t},y_1) \mbox{ or } \dots \mbox{ or } a(\overline{t},y_k) \leftarrow B, \no
intervene(a(\overline{t})).
\end{equation}
If the dynamic range of $a$ in the selection rule is not equal to its static range, i.e. expression $\{Z :
p(Z)\}$ is not omitted, then we also add the rule
\begin{equation}\label{r1.1a}
\leftarrow a(\overline{t},y), \no p(y), B, \no intervene(a(\overline{t})).
\end{equation}
Rule (\ref{r1.1}) selects the value of $a(\overline{t})$ from its range while rule (\ref{r1.1a}) ensures that
the selected value satisfies $p$.

\end{enumerate}

\item $\tau(\Pi)$ contains actions and observations of $\Pi$.

\item For each $\Sigma$-literal $l$, $\tau(\Pi)$ contains the rule:
\begin{equation} \label{obs-eq}
\leftarrow obs(l), \no l.
\end{equation}
\item For each atom $a(\overline{t})=y$, $\tau(\Pi)$ contains the rule:
\begin{equation}\label{action-eq}
a(\overline{t},y) \leftarrow do(a(\overline{t},y)).
\end{equation}

The rule (\ref{obs-eq}) guarantees that no possible world of the program fails to satisfy observation $l$. The rule
(\ref{action-eq}) makes sure the atoms that are made true by the action are indeed true.

\end{enumerate}
This completes our definition of $\tau(\Pi)$.

\st Before we proceed with some additional definitions let us comment on the difference between rules
\ref{obs-eq} and \ref{action-eq}. Since the P-log programs $T \cup obs(l)$ and $T \cup \{\leftarrow \no l\}$ have possible worlds which are identical except for possible occurrences of $obs(l)$, the new observation simply eliminates some of the possible worlds of $T$. This reflects understanding of observations in classical probability theory. In contrast, due to the possible non-monotonicity of the regular part of $T$, possible worlds of $T \cup do(l)$ can be substantially different from those of $T$ (as opposed to merely fewer in number); as we will illustrate in Section ~\ref{robot}.

\begin{definition}{[Possible worlds]}\\
{\rm An answer set of $\tau(\Pi)$ is called a {\em possible world} of $\Pi$. } \hfill $\Box$
\end{definition}
The set of all possible worlds of $\Pi$ will be denoted by $\Omega(\Pi)$. When $\Pi$ is clear from context we will
simply write $\Omega$. Note that due to our restriction on the signature of P-log programs possible worlds
of $\Pi$ are always finite.

\begin{example}\label{dice-a-prolog}[Dice example continued: P-log program
$T_1$]\\ {\rm Let $T_1$ be a P-log program consisting of $D_1$, $D_2$ and $D_3$ described in Examples
~\ref{dice1}, \ref{dice1a}, \ref{dice1b} and \ref{dice1c}. The Answer Set Prolog counterpart $\tau(T_{1})$ of $T_1$ will
consist of the following rules:

\st $dice(d_1)$. $dice(d_2)$. $score(1)$. $score(2)$. \\ $score(3)$. $score(4)$. $score(5)$. $score(6)$.\\
$person(mike)$. $person(john)$.\\ $owner(d_1,mike)$. $owner(d_2,john)$.

\st $even(D) \leftarrow roll(D,Y), Y \ mod \ 2 = 0$.

\st $\neg even(D) \leftarrow \no even(D)$.

\st $intervene(roll(D)) \leftarrow do(roll(D,Y))$.

\st $roll(D,1) \mbox{ or } \dots \mbox{ or } roll(D,6) \leftarrow B, \no intervene(roll(D))$.

\st $\neg roll(D,Y_1) \leftarrow roll(D,Y_2), Y_1\not=Y_2$.

\st $\neg owner(D,P_1) \leftarrow owner(D,P_2), P_1\not=P_2$.

\st $\neg even(D,B_1) \leftarrow even(D,B_2), B_1 \not= B_2$.

\st $\leftarrow obs(roll(D,Y)), \no roll(D,Y)$.

\st $\leftarrow obs(\neg roll(D,Y)), \no \neg roll(D,Y)$.

\st $roll(D,Y)) \leftarrow do(roll(D,Y))$.

\st The translation also contains similar $obs$ and $do$ axioms for other attributes which have been omitted here.

\st The variables $D$, $P$, $B$'s, and $Y$'s range over $dice$, $person$, $boolean$, and $score$ respectively. (In the
input language of Lparse used by Smodels\cite{niemela97} and several other answer set solving systems this typing can be expressed by the statement

\st $\#domain \ dice(D), person(P), score(Y)$.

\st Alternatively $c(X)$ can be added to the body of every rule containing variable $X$ with domain $c$. In the rest of
the paper we will ignore these details and simply use Answer Set Prolog with the typed variables as needed.)

\st It is easy to check that $\tau(T_{1})$ has $36$ answer sets which are possible worlds of P-log program $T_1$. Each
such world contains a possible outcome of the throws of the dice, e.g. $roll(d_1, 6), roll(d_2,3)$. } \hfill $\Box$
\end{example}

\subsection{Assigning measures of probability:} \label{assign-prob}
There are certain reasonableness criteria which we would like our programs to satisfy. These are normally easy to check
for P-log programs. However, the conditions are described using quantification over possible worlds, and so cannot be
axiomatized in Answer Set Prolog. We will state them as meta-level conditions, as follows (from this point forward we
will limit our attention to programs satisfying these criteria):
\begin{condition}\label{l1}[Unique selection rule]\\
{\rm If rules
$$[\ r_1 \ ]\ random(a(\overline{t}):\{Y:p_1(Y)\}) \leftarrow B_1. $$
$$[\ r_2 \ ]\ random(a(\overline{t}):\{Y:p_2(Y)\}) \leftarrow B_2. $$
belong to $\Pi$ then no possible world of $\Pi$ satisfies both $B_1$ and $B_2$. } \hfill $\Box$
\end{condition}

\st The above condition follows from the intuitive reading of random selection rules. In particular, there cannot be two different random experiments each of which determines the value of the same attribute.

\begin{condition}\label{l2}[Unique probability assignment]\\
{\rm If $\Pi$ contains a random selection rule
$$[\ r \ ]\ random(a(\overline{t}):\{Y:p(Y)\}) \leftarrow B.$$
along with two different probability atoms
$$pr_r(a(\overline{t})\ |_c \ B_1) = v_1 \mbox{ and } pr_r(a(\overline{t})\ |_c \ B_2) = v_2.$$
then no possible world of $\Pi$ satisfies $B$, $B_1$, and $B_2$. } \hfill $\Box$
\end{condition}
The justification of Condition 2 is as follows: If the conditions $B_1$ and $B_2$ can possibly both hold, and we do not
have $v_1 = v_2$, then the intuitive readings of the two {\em pr}-atoms are contradictory. On the other hand if $v_1 =
v_2$, the same information is represented in multiple locations in the program which is bad for maintenance and extension
of the program.

\st Note that we can still represent situations where the value of an attribute is determined by multiple possible
causes, as long as the attribute is not explicitly random. To illustrate this point let us consider a simple example
from \cite{joost06}.

\begin{example}\label{guns}[Multiple Causes: Russian roulette with two guns]\\
{\rm Consider a game of Russian roulette with two six-chamber guns. Each of the guns is loaded with a single bullet.
What is the probability of the player dying if he fires both guns?

\st Note that in this example pulling the trigger of the first gun and pulling the trigger of the second gun are two
independent causes of the player's death. That is, the mechanisms of death from each of the two guns are separate and do
not influence each other.

\st The logical part of the story can be encoded by the following P-log program $\Pi_g$:

\st $gun = \{1,2\}$. \\ $pull\_trigger : gun \rightarrow boolean$. \% $pull\_trigger(G)$ says that the player pulls the
trigger of gun $G$.\\ $fatal : gun \rightarrow boolean$. \% $fatal(G)$ says that the bullet from
gun $G$ is sufficient to kill the player.\\
$is\_dead : boolean$. \% $is\_dead$ says that the player is dead.\\ $[r(G)]\ :\ random(fatal(G)) \leftarrow
pull\_trigger(G)$. \\ $is\_dead \leftarrow fatal(G)$.\\ $\neg is\_dead \leftarrow \no is\_dead$.\\ $pull\_trigger(G)$.

\st Here the value of the random attribute $fatal(1)$, which stands for ``Gun 1 causes a wound sufficient to kill the
player'' is generated at random by rule $r(1)$. Similarly for $fatal(2)$. The attribute $is\_dead$, which stands for the
death of the player, is described in terms of $fatal(G)$ and hence is not explicitly random. To define the probability
of $fatal(G)$ we will assume that when the cylinder of each gun is spun, each of the six chambers is equally likely to
fall under the hammer. Thus,

\st $pr_{r(1)}(fatal(1)) = 1/6$.\\
    $pr_{r(2)}(fatal(2)) = 1/6$.

\st Intuitively the probability of the player's death will be $11/36$. At the end of this section we will learn how to
compute this probability from the program.

\st Suppose now that due to some mechanical defect the probability of the first gun firing its bullet (and therefore
killing the player) is not $1/6$ but, say, $11/60$. Then the probability atoms above will be replaced by

\st $pr_{r(1)}(fatal(1)) = 11/60$.\\
    $pr_{r(2)}(fatal(2)) = 1/6$.

\st The probability of the player's death defined by the new program will be $0.32$. Obviously, both programs satisfy
Conditions \ref{l1} and \ref{l2} above.

\st Note however that the somewhat similar program

\st $gun = \{1,2\}$.\\ $pull\_trigger : gun \rightarrow boolean$.\\ $is\_dead : boolean$.\\ $[r(G)]\ :\ random(is\_dead)
\leftarrow pull\_trigger(G)$. \\ $pull\_trigger(G)$.

\st does not satisfies Condition \ref{l1} and hence will not be allowed in P-log.} \hfill $\Box$
\end{example}

\noindent
The next example presents a slightly different version of reasoning with multiple causes.
\begin{example}\label{roulette}[Multiple Causes: The casino story]\\
{\rm A roulette wheel has 38 slots, two of which are green. Normally, the ball falls into one of these slots at random.
However, the game operator and the casino owner each have buttons they can press which ``rig'' the wheel so that the
ball falls into slot 0, which is green, with probability 1/2, while the remaining slots are all equally likely. The game is rigged in the same way no matter which button is pressed, or if both are pressed. In this example, the rigging of the game can be viewed as having two causes. Suppose in this particular game both buttons were pressed. What is the
probability of the ball falling into slot $0$?

\st The story can be represented in P-log as follows:

\st $slot = \{zero,double\_zero,1..36\}$.\\ $button = \{1,2\}$.\\ $pressed : button \rightarrow boolean$.\\ $rigged:
boolean$. \\ $falls\_in : slot$.\\ $[r] : random(falls\_in)$.\\ $rigged \leftarrow pressed(B)$.\\ $\neg rigged
\leftarrow \no rigged$.\\ $pressed(B)$.\\ $pr_r(falls\_in = zero |_c rigged ) = 1/2$.

\st Intuitively, the probability of the ball falling into slot zero is $1/2$. The same result will be obtained by our
formal semantics. Note that the program obviously satisfies Conditions \ref{l1} and \ref{l2}. However the following
similar program violates Condition \ref{l2}.

\st $slot = \{zero,double\_zero,1..36\}$.\\ $button = \{1,2\}$.\\ $pressed : button \rightarrow boolean$.\\ $falls\_in :
slot$.\\ $[r] : random(falls\_in)$.\\ $pressed(B)$.\\ $pr_r(falls\_in = zero |_c pressed(B) ) = 1/2$.

\st Condition \ref{l2} is violated here because two separate pr-atoms each assign probability to the literal
$falls\_in=zero$. Some other probabilistic logic languages allow this, employing various systems of ``combination rules'' to compute the overall probabilities of literals whose probability values are multiply assigned. The study of
combination rules is quite complex, and so we avoid it here for simplicity. }\hfill $\Box$
\end{example}

\begin{condition}\label{l3}[No probabilities assigned outside
of dynamic range]\\ {\rm If $\Pi$ contains a random selection rule
$$[\ r \ ]\ random(a(\overline{t}):\{Y:p(Y)\}) \leftarrow B_1. $$
along with probability atom
$$pr_r(a(\overline{t}) =y \ |_c \ B_2) = v.$$
then no possible world $W$ of $\Pi$ satisfies $B_1$ and $B_2$ and $\no intervene(a(\overline{t}))$ but fails to satisfy
$p(y)$. } \hfill $\Box$
\end{condition}
The condition ensures that probabilities are only assigned to logically possible outcomes of random selections. It
immediately follows from the intuitive reading of statements (\ref{e1}) and (\ref{e2}).

\st To better understand the intuition behind our definition of probabilistic measure it may be useful to consider an
intelligent agent in the process of constructing his possible worlds. Suppose he has already constructed a part $V$ of a (not yet completely constructed) possible world $W$, and suppose that $V$ satisfies the precondition of some random
selection rule $r$. The agent can continue his construction by considering a random experiment associated with $r$. If
$y$ is a possible outcome of this experiment then the agent may continue his construction by adding the atom
$a(\overline{t}) = y$ to $V$. To define the probabilistic measure $\mu$ of the possible world $W$ under construction, we
need to know the likelihood of $y$ being the outcome of $r$, which we will call the {\em causal probability} of the atom
$a(\overline{t}) = y$ in $W$. This information can be obtained from a pr-atom $pr_r(a(\overline{t}) =y) = v$ of our
program or computed using the principle of indifference. In the latter case we need to consider the collection $R$ of
possible outcomes of experiment $r$. For example if $y \in R$, there is no probability atom assigning probability to
outcomes of $R$, and $|R| = n$, then the causal probability of $a(\overline{t}=y)$ in $W$ will be $1/n$.

\st Let $v$ be the causal probability of $a(\overline{t}) = y$. The atom $a(\overline{t}) = y$ may be dependent, in the
usual probabilistic sense, with other atoms already present in the construction. However $v$ is not read as the
probability of $a(\overline{t}) = y$, but the probability that, given what the agent knows about the possible world at
this point in the construction, the experiment determining the value of $a(\overline{t})$ will have a certain result.
Our assumption is that these experiments are independent, and hence it makes sense that $v$ will have a multiplicative
effect on the probability of the possible world under construction. (This approach should be familiar to those
accustomed to working with Bayesian nets.) This intuition will be captured by the following definitions.

\begin{definition}\label{PossVal}{[Possible outcomes]}\\
{\rm Let $W$ be a consistent set of literals of $\Sigma$, $\Pi$ be a P-log program, $a$ be an attribute, and $y$ belong
to the range of $a$. We say that the atom $a(\overline{t}) = y$ is {\em possible} in $W$ with respect to $\Pi$ if $\Pi$
contains a random selection rule $r$ for $a(\overline{t})$, where if $r$ is of the form (\ref{e1}) then $p(y) \in W$ and $W$ satisfies $B$, and if $r$ is of the form (\ref{e1a1}) then $W$ satisfies $B$. We also say that $y$ is a {\em possible outcome} of $a(\overline{t})$ in $W$ with respect to $\Pi$ via rule $r$, and that $r$ is a {\em generating rule} for the atom $a(\overline{t}) = y$. \hfill $\Box$ }
\end{definition}

\st Recall that, based on our convention, if the range of $a$ is boolean then we can just say that $a(\overline{t})$ and
$\neg a(\overline{t})$ are possible in $W$. (Note that by Condition \ref{l1}, if $W$ is a possible world of $\Pi$ then
each atom possible in $W$ has exactly one generating rule.)

\st Note that, as discussed above, there is some subtlety here because we are describing $a(\overline{t}) = y$ as
possible, though not necessarily true, with respect to a particular set of literals and program $\Pi$.

\st For every $W \in \Omega(\Pi)$ and every atom $a(\overline{t}) = y$ possible in $W$ we will define the corresponding
causal probability $P(W, a(\overline{t}) = y)$. Whenever possible, the probability of an atom $a(\overline{t}) = y$ will
be directly assigned by pr-atoms of the program and denoted by $PA(W, a(\overline{t}) = y)$. To define probabilities of
the remaining atoms we assume that by default, all values of a given attribute which are not assigned a probability are
equally likely. Their probabilities will be denoted by $PD(W, a(\overline{t}) = y)$. ($PA$ stands for {\em assigned
probability} and $PD$ stands for {\em default probability}).

\st For each atom $a(\overline{t}) = y$ possible in $W$:
\begin{enumerate}
\item Assigned probability:

\noindent If $\Pi$ contains $pr_r(a(\overline{t})=y \ |_c \ B) = v$ where $r$ is the generating rule of
$a(\overline{t}) = y$, $B \subseteq W$, and $W$ does not contain $intervene(a(\overline{t}))$, then
$$PA(W, a(\overline{t}) = y) = v$$
\item Default probability:

For any set $S$, let $|S|$ denote the cardinality of $S$. Let $A_{a(\overline{t})}(W) =\{ y \ | \ PA(W,
a(\overline{t})= y)\ \mbox{is defined} \}$, and $a(\overline{t}) = y$ be possible in $W$ such that $y \not\in
A_{a(\overline{t})}(W)$. Then let
$$\alpha_{a(\overline{t})}(W) = \sum_{y \in
A_{a(\overline{t})}(W)} PA(W, a(\overline{t}) = y)$$
$$\beta_{a(\overline{t})}(W) = |\{ y \ : a(\overline{t}) = y \mbox{ is
possible in } W \mbox{ and } y \not\in A_{a(\overline{t})}(W)\}|$$
$$PD(W, a(\overline{t}) = y) = \frac{1 - \alpha_{a(\overline{t})}(W)}
{\beta_{a(\overline{t})}(W) }$$

\st \item Finally, the causal probability $P(W, a(\overline{t}) = y)$ of $a(\overline{t}) = y$ in $W$ is defined
by:
\[P(W, a(\overline{t}) = y) = \left\{\begin{array}{ll}
PA(W, a(\overline{t}) = y) & \mbox{ if } y \in A_{a(\overline{t})}(W)\\ PD(W, a(\overline{t}) = y) & \mbox{
otherwise}.
\end{array}
\right.\]
\end{enumerate}

\begin{example} \label{dice-a-prolog-b} [Dice example continued: P-log program
$T_1$]\\ {\rm Recall the P-log program $T_1$ from Example~\ref{dice-a-prolog}. The program contains the following
probabilistic information:

\st $pr(roll(d_1)=i \ |_c \ owner(d_1) = mike) = 3/20, \mbox{ for each } i \mbox{ such that }1 \leq i \leq 5.$\\
$pr(roll(d_1)=6 \ |_c \ owner(d_1) = mike) = 1/4.$\\ $pr(roll(d_2)=i \ |_c \ owner(d_2) = john) = 1/6, \mbox{ for each }
$i$ \mbox{ such that } 1 \leq i \leq 6.$

\st We now consider a possible world
$$W=\{owner(d_1,mike),owner(d_2,john),roll(d_1,6), roll(d_2,3),\dots\}$$
of $T_1$ and compute $P(W,roll(d_i)=j)$ for every die $d_i$ and every possible score $j$.

\st According to the above definition, $PA(W, roll(d_i)= j)$ and $P(W, roll(d_i) = j)$ are defined for every random atom
(i.e. atom formed by a random attribute) $roll(d_i)=j$ in $W$ as follows:

\st $P(W,roll(d_1) = i)$ = $PA(W, roll(d_1) = i) = 3/20, \mbox{ for each } i \mbox{ such that }1 \leq i \leq 5.$\\ $P(W,
roll(d_1) = 6)$ = $PA(W, roll(d_1) = 6) = 1/4.$\\ $P(W, roll(d_2) = i)$ = $PA(W, roll(d_2) = i) = 1/6, \mbox{ for each }
i \mbox{ such that }1 \leq i \leq 6.$ } \hfill $\Box$
\end{example}
\begin{example}\label{dp1}[Dice example continued: P-log program $T_{1.1}$] \\
{\rm In the previous example all random atoms of $W$ were assigned probabilities. Let us now consider what will happen
if explicit probabilistic information is omitted. Let $D_{3.1}$ be obtained from $D_3$ by removing all probability atoms
except

\st $pr(roll(D)=6 \ |_c \ owner(D) = mike) = 1/4$.

\st Let $T_{1.1}$ be the P-log program consisting of $D_1$, $D_2$ and $D_{3.1}$ and let $W$ be as in the previous
example. Only the atom $roll(d_1)=6$ will be given an assigned probability:

\st $P(W, roll(d_1) = 6)$ = $PA(W, roll(d_1) = 6) = 1/4$.

\noindent The remaining atoms receive the expected default probabilities:

\st $P(W, roll(d_1) = i)$ = $PD(W, roll(d_1) = i) = 3/20, \mbox{ for each } i \mbox{ such that }1 \leq i \leq 5.$

\noindent $P(W, roll(d_2) = i)$ = $PD(W, roll(d_2) = i) = 1/6, \mbox{ for each } i \mbox{ such that }1 \leq i \leq 6.$
}\hfill $\Box$
\end{example}

Now we are ready to define the measure, $\mu_\Pi$, induced by the P-log program $\Pi$.
\begin{definition}\label{meas1}{[Measure]}
{\rm
\begin{enumerate}
\item Let $W$ be a possible world of $\Pi$. The {\em unnormalized probability}, $\hat{\mu}_\Pi(W)$, of a possible
    world $W$ {\em induced by} $\Pi$ is
$$\hat{\mu}_\Pi(W) = \prod_{a(\overline{t},y) \in \ W}
P(W, a(\overline{t}) = y )
$$ where the product is taken over
atoms for which $P(W, a(\overline{t}) = y )$ is defined.

\item Suppose $\Pi$ is a P-log program having at least one possible world with nonzero unnormalized probability. The
    {\em measure}, $\mu_\Pi(W)$, of a possible world $W$ {\em induced by} $\Pi$ is the unnormalized probability of
    $W$ divided by the sum of the unnormalized probabilities of all possible worlds of $\Pi$, i.e.,
$$\mu_\Pi(W) = \frac{\hat{\mu}_\Pi(W)}
{\sum_{W_i \in \Omega}\hat{\mu}_\Pi(W_i)}$$
\end{enumerate}
\noindent When the program $\Pi$ is clear from the context we may simply write $\hat{\mu}$ and $\mu$ instead of
$\hat{\mu}_\Pi$ and $\mu_\Pi$ respectively. \hfill $\Box$ }
\end{definition}

\st The unnormalized measure of a possible world $W$ corresponds, from the standpoint of classical probability, to the
unconditional probability of $W$. Each random atom $a(\overline{t}) = y$ in $W$ is thought of as the outcome of a random experiment that takes place in the construction of $W$, and $P(W, a(\overline{t}) = y )$ is the probability of that
experiment having the result $a(\overline{t}) = y$ in $W$. The multiplication in the definition of unnormalized measure
is justified by an assumption that all experiments performed in the construction of $W$ are independent. This is subtle
because the experiments themselves do not show up in $W$ --- only their results do, and the results may {\em not} be
independent.\footnote{For instance, in the upcoming Example \ref{rat}, random attributes $arsenic$ and $death$
respectively reflect whether or not a given rat eats arsenic, and whether or not it dies. In that example, $death$ and
$arsenic$ are clearly dependent. However, we assume that the factors which determine whether a poisoning will lead to
death (such as the rat's constitution, and the strength of the poison) are independent of the factors which determine
whether poisoning occurred in the first place.}

\begin{example}\label{dp2}[Dice example continued: $T_1$ and $T_{1.1}$]\\
{\rm The measures of the possible worlds of Example \ref{dp1} are given by

\st $\mu(\{ roll(d_1, 6), roll(d_2, y), \dots \}) =1/24$, for $1 \leq y \leq 6$, and

\st $\mu(\{ roll(d_1, u), roll(d_2, y), \dots \}) =1/40$, for $1 \leq u \leq 5$ and $1 \leq y \leq 6$.

\noindent where only random atoms of each possible world are shown. } \hfill $\Box$
\end{example}

\st Now we are ready for our main definition.

\begin{definition}\label{prob}{[Probability]}\\
{\rm Suppose $\Pi$ is a P-log program having at least one possible world with nonzero unnormalized probability. The {\em
probability}, $P_{\Pi}(E)$, of a set $E$ of possible worlds of program $\Pi$ is the sum of the measures of the possible
worlds from $E$, i.e.
$$P_\Pi(E) = \sum_{W \in E}\mu_\Pi(W). $$
} \hfill $\Box$
\end{definition}
\noindent When $\Pi$ is clear from the context we may simply write $P$ instead of $P_\Pi$.

\st The function $P_\Pi$ is not always defined, since not every syntactically correct P-log program satisfies the
condition of having at least one possible world with nonzero unnormalized measure. Consider for instance a program $\Pi$
consisting of facts\\ $p(a).$\\ $\neg p(a).$\\ The program has no answer sets at all, and hence here $P_\Pi$ is not
defined. The following proposition, however, says that when $P_\Pi$ {\em is} defined, it satisfies the Kolmogorov axioms
of probability. This justifies our use of the term ``probability'' for the function $P_\Pi$. The proposition follows
straightforwardly from the definition.
\begin{proposition}\label{axioms}[Kolmogorov Axioms]\\
{\rm For a P-log program $\Pi$ for which the function $P_{\Pi}$ is defined we have
\begin{enumerate}
\item For any set $E$ of possible worlds of $\Pi$, $P_\Pi(E) \geq 0$.

\item If $\Omega$ is the set of all possible worlds of $\Pi$ then $P_\Pi(\Omega) =1$.

\item For any disjoint subsets $E_1$ and $E_2$ of possible worlds of $\Pi$, $P_\Pi(E_1 \cup E_2) = P_\Pi(E_1) +
    P_\Pi(E_2)$.
\hfill $\Box$
\end{enumerate}
}
\end{proposition}

\st In logic-based probability theory a set $E$ of possible worlds is often represented by a propositional formula $F$
such that $W \in E$ iff $W$ is a model of $F$. In this case the probability function may be defined on propositions as

\st $P(F) =_{def} P(\{W : W \mbox{ is a model of } F\})$.

\st The value of $P(F)$ is interpreted as the degree of reasoner's belief in $F$. A similar idea can be used in our
framework. But since the connectives of Answer Set Prolog are different from those of Propositional Logic the notion of
propositional formula will be replaced by that of formula of Answer Set Prolog (ASP formula). In this paper we limit our
discussion to relatively simple class of ASP formulas which is sufficient for our purpose.

\begin{definition}{ [ASP Formulas (syntax)]}\\
{\rm For any signature $\Sigma$
\begin{itemize}
\item An extended literal of $\Sigma$ is an \emph{ASP formula}.

\item if $A$ and $B$ are \emph{ASP formulas} then $(A \wedge B)$ and $(A \mbox{ \em{or} } B)$ are ASP formulas. \hfill $\Box$
\end{itemize}
}
\end{definition}

\noindent For example, $((p \wedge \no q \wedge \neg r) \mbox{ \em{or} } (\no r))$ is an ASP formula but $(\no (\no p))$
is not. More general definition of ASP formulas which allows the use of negations $\neg$ and $\no$ in front of arbitrary
formulas can be found in \cite{LiPeVa}.

\st Now we define the truth ($W \vdash A$) and falsity ($W \dashv A$) of an ASP formula $A$ with respect to a possible
world $W$:

\begin{definition}{[ASP Formulas (semantics)]}\\
{\rm
\begin{enumerate}
\item For any $\Sigma$-literal $l$, $W \vdash l$ if $l \in W$; $W \dashv l$ if $\overline{l} \in W$.

\item For any extended $\Sigma$-literal $\no l$, $W \vdash \no l$ if $l \not\in W$; $W \dashv \no l$ if $l \in W$.

\item $W \vdash (A_1 \wedge A_2)$ if $W \vdash A_1$ and $W \vdash A_2$; $W \dashv (A_1 \wedge A_2)$ if $W \dashv
    A_1$ or $W \dashv A_2$.

\item $W \vdash (A_1 \mbox{ \em{or} } A_2)$ if $W \vdash A_1$ or $W \vdash A_2$; $W \dashv (A_1 \mbox{ \em{or} }
    A_2)$ if $W \dashv A_1$ and $W \dashv A_2$.  \hfill $\Box$
\end{enumerate}
}
\end{definition}
\noindent An ASP formula $A$ which is neither true nor false in $W$ is {\em undefined} in $W$. This introduces some
subtlety. The axioms of modern mathematical probability are viewed as axioms about measures on sets of possible worlds,
and as such are satisfied by P-log probability measures. However, since we are using a three-valued logic, some
classical consequences of the axioms for the probabilities of {\em formulae} fail to hold. Thus, all theorems of
classical probability theory can be applied in the context of P-log; but we must be careful how we interpret set operations
in terms of formulae. For example, note that formula ($l \mbox{ \em{or} } \no l$) is {\em true} in every possible world
$W$. However formula ($p \mbox{ \em{or} } \neg p$) is undefined in any possible world containing neither $p$ nor $\neg
p$. Thus if $P$ is a P-log probability measure, we will always have $P(\no l) = 1 - P(l)$, but not necessarily $P(\neg
l) = 1 - P(l)$.

%%%%%%%%%%%%%%%% \st INSERTED

\st Consider for instance an ASP program $P_1$ from the introduction. If we expand $P_1$ by the appropriate declarations we obtain a program $\Pi_1$ of P-log. It's only possible world is $W_0 = \{p(a),\neg p(b), q(c)\}$. Since neither $p$ nor $q$ are random, its measure, $\mu(W_0)$ is $1$ (since the empty product is $1$). However, since the truth value of $p(c) \oor \neg p(c)$  in $W_0$ is undefined, $P_{\Pi_1}(p(c) \oor \neg p(c)) = 0$. This is not surprising since $W_0$ represents a possible set of beliefs of the agent associated with $\Pi_1$ in which $p(c)$ is simply ignored. (Note that the probability of formula $q(c)$ which expresses this fact is properly equal to $1$).

\st Let us now look at program $\Pi_2$ obtained from $\Pi_1$ by declaring $p$ to be a random attribute. This time $p(c)$ is not ignored. Instead the agent considers two possibilities and constructs two complete\footnote{A possible world $W$
of program $\Pi$ is called \emph{complete} if for any ground atom $a$ from the signature of $\Pi$, $a \in W$ or $\neg a
\in W$.} possible worlds:\\ $W_1 = \{p(a),\neg p(b), p(c),\neg q(c)\}$ and \\ $W_2 = \{p(a),\neg p(b), \neg
p(c),\neg q(c)\}$.\\ Obviously $P_{\Pi_2}(p(c) \oor \neg p(c)) = 1$.

\st It is easy to check that if all possible worlds of a P-log program $\Pi$ are complete then $P_{\Pi}(l \oor \neg l) = 1$. This is the case for instance when $\Pi$ contains no regular part, or when the regular part of $\Pi$ consists of
definitions of relations $p_1,\dots,p_n$ (where a {\emph definition} of a relation $p$ is a collection of rules which
determines the truth value of atoms built from $p$ to be true or false in all possible worlds).

%%%%%%%%%%%%%%%%

\st Now the definition of probability can be expanded to  ASP formulas.
\begin{definition}\label{peob-of-form}{[Probability of Formulas]}\\
{\rm The {\em probability} with respect to program $\Pi$ of a formula $A$, $P_{\Pi}(A)$, is the sum of the measures of
the possible worlds of $\Pi$ in which $A$ is true, i.e.
$$P_\Pi(A) = \sum_{W \vdash A}\mu_\Pi(W).$$
} \hfill $\Box$
\end{definition}
\noindent As usual when convenient we omit $\Pi$ and simply write $P$ instead of $P_\Pi$.

\begin{example}\label{dp2a}[Dice example continued]\\
{\rm Let $T_1$ be the program from Example~\ref{dice-a-prolog}. Then, using the measures computed in Example~\ref{dp2}
and the definition of probability we have, say

\st $P_{T_1}(roll(d_1)=6) = 6*(1/24) = 1/4$.\\ $P_{T_1}(roll(d_1)=6 \wedge even(d_2))= 3*(1/24) = 1/8$. } \hfill $\Box$
\end{example}

\begin{example}\label{e11} [Causal probability equal to $1$]\\
{\rm Consider the P-log program $\Pi_0$ consisting of:

\st $a \ : \ boolean$.\\ $random \ a$.\\ $pr(a)=1.$

\st The translation of its logical part, $\tau(\Pi_0)$, will consist of the following:

\st $intervene(a) \leftarrow do(a)$.

\st $intervene(a) \leftarrow do(\neg a)$.

\st $a \mbox{ or } \neg a \leftarrow \no intervene(a)$.

\st $ \leftarrow obs(a), \no a$.

\st $ \leftarrow obs(\neg a), \no \neg a$.

\st $ a \leftarrow do(a)$.

\st $\neg a \leftarrow do(\neg a)$.

\st $\tau(\Pi_{0})$ has two answer sets $W_1 = \{a,\dots \}$ and $W_2 = \{\neg a,\dots\}$. The probabilistic part of
$\Pi_0$ will lead to the following probability assignments.

\st $P(W_1, a ) = 1$.\\ $P(W_1, \neg a ) = 0$.\\ $P(W_2, a ) = 1$.\\ $P(W_2, \neg a ) = 0$.

\st $\hat{\mu}_{\Pi_0}(W_1) = 1$.\\ $\hat{\mu}_{\Pi_0}(W_2) = 0$.\\ $\mu_{\Pi_0}(W_1) = 1$.\\ $\mu_{\Pi_0}(W_2) = 0$.

\noindent This gives us $P_{\Pi_0}(a) =1$. } \hfill $\Box$
\end{example}

\begin{example}{ [Guns example continued]}\\
{\rm Let $\Pi_g$ be the P-log program from Example~\ref{guns}. It is not difficult to check that the program has four
possible worlds. All four contain $\{gun(1),gun(2),pull\_trigger(1),pull\_trigger(2)\}$. Suppose now that $W_1$ contains
$\{fatal(1), \neg fatal(2)\}$, $W_2$ contains $\{\neg fatal(1), fatal(2)\}$, $W_3$ contains $\{fatal(1), fatal(2)\}$,
and $W_4$ contains $\{\neg fatal(1), \neg fatal(2)\}$. The first three worlds contain $is\_dead$, the last one contains
$\neg is\_dead$. Then

\st $\mu_{\Pi_g}(W_1) = 1/6 * 5/6 = 5/36$.\\ $\mu_{\Pi_g}(W_2) = 5/6 * 1/6 = 5/36$.\\ $\mu_{\Pi_g}(W_3) = 1/6 * 1/6 =
1/36$.\\ $\mu_{\Pi_g}(W_4) = 5/6 * 5/6 = 25/36$.

\st and hence

\st $P_{\Pi_g}(is\_dead) = 11/36$. } \hfill $\Box$
\end{example}
As expected, this is exactly the intuitive answer from Example \ref{guns}. A similar argument can be used to compute
probability of $rigged$ from Example \ref{roulette}.

\st Even if $P_\Pi$ satisfies the Kolmogorov axioms it may still contain questionable probabilistic information. For
instance a program containing statements $pr(p) = 1$ and $pr(\neg p) = 1$ does not seem to have a clear intuitive
meaning. The next definition is meant to capture the class of programs which are logically and probabilistically
coherent.

\begin{definition}\label{consistency-defn}{[Program Coherency]}\\
{\rm Let $\Pi$ be a P-log program and $\Pi^\prime$ be obtained from $\Pi$ by removing all observations and actions.
$\Pi$ is said to be {\em consistent} if $\Pi$ has at least one possible world.

\noindent We will say that a consistent program $\Pi$ is {\em coherent} if
\begin{itemize}
\item $P_{\Pi}$ is defined.

\item For every selection rule $r$ with the premise $K$ and every probability atom $pr_r(a(t)=y \ |_c \ B) = v$ of
    $\Pi$, if $P_{\Pi^\prime}(B \cup K)$ is not equal to $0$ then $P_{\Pi^{\prime} \cup obs(B) \cup obs(K)}(a(t)=y)= v$.  \hfill $\Box$
\end{itemize}
}
\end{definition}

\st Coherency intuitively says that causal probabilities entail corresponding conditional probabilities. We now give two
examples of programs whose probability functions are defined, but which are not coherent.

\begin{example}\label{e13}
{\rm Consider the programs $\Pi_5$:

\st $a \ : \ boolean$.\\ $random \ a$.\\ $a.$\\ $pr(a)=1/2.$

\st and $\Pi_6$:

\st $a \ : \ \{0, 1, 2\}$.\\ $random \ a$.\\ $pr(a=0) = pr(a=1) = pr(a=2) = 1/2.$

\st Neither program is coherent. $\Pi_5$ has one possible world $W = \{a\}$. We have $\hat{\mu}_{\Pi_5}(W) = 1/2$,
$\mu_{\Pi_5}(W) = 1$, and $P_{\Pi_5}(a) = 1$. Since $pr(a)=1/2$, $\Pi_5$ violates condition (2) of coherency.

\st $\Pi_6$ has three possible worlds, $\{a=0\}$, $\{a=1\}$, and $\{a=2\}$ each with unnormalized probability $1/2$.
Hence $P_{\Pi_6}(a=0) = 1/3$, which is different from $pr(a=0)$ which is $1/2$; thus making $\Pi_6$ incoherent. }
\hfill $\Box$
\end{example}

\noindent The following two propositions give conditions on the probability atoms of a P-log program which are necessary
for its coherency.

\begin{proposition}\label{nc1}
{\rm Let $\Pi$ be a coherent P-log program without any observations or actions, and $a(\overline{t})$ be an attribute
term from the signature of $\Pi$. Suppose that $\Pi$ contains a selection rule
$$[r]\ random(a(\overline{t}): \{X\ :\ p(X)\}) \leftarrow B_1.$$
and there is a subset $c = \{y_1,\dots,y_{n}\}$ of the range of $a(\overline{t})$ such that for every possible world $W$
of $\Pi$ satisfying $B_1$, we have $\{Y : W \vdash p(Y)\}=\{y_1,\dots,y_n\}$. Suppose also that for some fixed $B_2$,
$\Pi$ contains probability atoms of the form
$$ pr_r(a(\overline{t})=y_i
\ |_c \ B_2) = p_i.$$ for all $1 \leq i \leq n$. Then
$$P_\Pi(B_1 \wedge B_2) = 0 \ \ \ \ \ \mbox{ or }\ \ \ \ \ \ \ \sum_{i=1}^{n} p_i = 1$$
} \hfill $\Box$
\end{proposition}

\st {\bf Proof}: Let $\hat{\Pi} = \Pi \cup obs(B_1) \cup obs(B_2)$ and let $P_\Pi(B_1 \wedge B_2) \neq 0$. From this,
together with rule \ref{obs-eq} from the definition of the mapping $\tau$ from section \ref{def-pos-worlds}, we have that
$\hat{\Pi}$ has a possible world with non-zero probability. Hence by Proposition \ref{axioms}, $P_{\hat{\Pi}}$ satisfies
the Kolmogorov Axioms. By Condition 2 of coherency, we have $P_{\hat{\Pi}}( a(\overline{t}) = y_i) = p_i$, for all
$1\leq i \leq n$. By rule \ref{obs-eq} of the definition of $\tau$ we have that every possible world of $\hat{\Pi}$
satisfies $B_1$. This, together with rules \ref{r0}, \ref{r1.1}, and \ref{r1.1a} from the same definition implies that
every possible world of $\hat{\Pi}$ contains exactly one literal of the form $a(\overline{t})=y$ where $y \in c$. Since
$P_{\hat{\Pi}}$ satisfies the Kolmogorov axioms we have that if $\{F_1,\dots, F_n\}$ is a set of literals exactly one of
which is true in every possible world of $\hat{\Pi}$ then
$$\sum_{i=1}^{n} P_{\hat{\Pi}}(F_i) = 1$$
This implies that
$$ \sum_{i=1}^{n} p_i = \sum_{i=1}^{n} P_{\hat{\Pi}}(a(\overline{t}) = y_i)
= 1$$

The proof of the following is similar:

\begin{proposition}
{\rm Let $\Pi$ be a coherent P-log program without any observations or actions, and $a(\overline{t})$ be an attribute
term from the signature of $\Pi$. Suppose that $\Pi$ contains a selection rule
$$[r]\ random(a(\overline{t}):p) \leftarrow B_1.$$
and there is a subset $c = \{y_1,\dots,y_{n}\}$ of the range of $a(\overline{t})$ such that for every possible world $W$
of $\Pi$ satisfying $B_1$, we have $\{Y : W \vdash p(Y)\}=\{y_1,\dots,y_n\}$. Suppose also that for some fixed $B_2$,
$\Pi$ contains probability atoms of the form
$$ pr_r(a(\overline{t})=y_i
\ |_c \ B_2) = p_i.$$ for some $1 \leq i \leq n$. Then
$$P_\Pi(B_1 \wedge B_2) = 0 \ \ \ \ \ \mbox{ or }\ \ \ \ \ \ \ \sum_{i=1}^{n} p_i \leq 1$$
} \hfill $\Box$
\end{proposition}

\section{Belief Update in P-log}\label{sec4}
In this section we address the problem of belief updating --- the ability of an agent to change degrees of belief
defined by his current knowledge base. If $T$ is a P-log program and $U$ is a collection of statements such that $T \cup
U$ is coherent we call $U$ an {\em update} of $T$. Intuitively $U$ is viewed as new information which can be added to an
existent knowledge base, $T$. Explicit representation of the agent's beliefs allows for a natural treatment of belief
updates in P-log. The reasoner should simply add the new knowledge $U$ to $T$ and check that the result is coherent. If
it is then the new degrees of the reasoner's beliefs are given by the function $P_{T \cup U}$. As mentioned before we
plan to expand our work on P-log with allowing its regular part be a program in CR-Prolog \cite{cr} which has a much
more liberal notion of consistency than Answer Set Prolog. The resulting language will allow a substantially larger set
of possible updates.

\st In what follows we compare and contrast different types of updates and investigate their relationship with the
updating mechanisms of more traditional Bayesian approaches.

\subsection{ P-log Updates and Conditional Probability}

In Bayesian probability theory the notion of conditional probability is used as the primary mechanism for updating
beliefs in light of new information. If $P$ is a probability measure (induced by a P-log program or otherwise), then the
conditional probability $P(A | B)$ is defined as $P(A \wedge B) / P(B)$, provided $P(B)$ is not $0$. Intuitively, $P(A | B)$
is understood as the probability of a formula $A$ with respect to a background theory and a set $B$ of all of the
agent's additional observations of the world. The new evidence $B$ simply eliminates the possible worlds which do not
satisfy $B$. To emulate this type of reasoning in P-log we first assume that the only formulas observable by the agent
are literals. (The restriction is needed to stay in the syntactic boundaries of our language. As mentioned in Section
\ref{sec2} this restriction is not essential and can be eliminated by using a syntactically richer version of Answer Set
Prolog.) The next theorem gives a relationship between classical conditional probability and updates in P-log. Recall
that if $B$ is a set of literals, adding the observation $obs(B)$ to a program $\Pi$ has the effect of removing all
possible worlds of $\Pi$ which fail to satisfy $B$.

\begin{proposition}\label{class_def}[Conditional Probability in P-log]\\
{\rm For any coherent P-log program $T$, formula $A$, and a set of $\Sigma$-literals $B$ such that $P_T(B) \neq 0$,
$$P_{T \cup obs(B)}(A) = P_T(A \wedge B) / P_T(B)$$
In other words,
$$P_T(A | B) = P_{T \cup obs(B)}(A)$$
}\hfill $\Box$
\end{proposition}

\noindent {\bf Proof:}\\

\noindent \st Let us order all possible worlds of $T$ in such a way that\\ \noindent $\{w_1 ... w_j\}$ is the set
of all possible worlds of $T$ that contain both $A$ and $B$, \\ $\{w_1 ... w_l\}$ is the set of all possible worlds of
$T$ that contain $B$, and \\ $\{w_1 ... w_n\}$ is the set of all possible worlds of $T$.

\noindent Programs of Answer Set Prolog are monotonic with respect to constraints, i.e. for any program $\Pi$ and a set
of constraints $C$, $X$ is an answer set of $\Pi \cup C$ iff it is an answer set of $P$ satisfying $C$. Hence the
possible worlds of $T \cup obs(B)$ will be all and only those of $T$ that satisfy $B$. In what follows, we will write
$\mu$ and $\hat{\mu}$ for $\mu_T$ and $\hat{\mu}_T$, respectively. Now, by the definition of probability in P-log, if
$P_T(B) \not=0$, then
$$ P_{T \cup obs(B)}(A) =
\frac{\sum_{i = 1}^j \hat{\mu}(w_i)} {\sum_{i = 1}^l \hat{\mu}(w_i) }
$$
\noindent Now if we divide both the numerator and denominator by the normalizing factor for $T$, we have
$$ \frac{\sum_{i = 1}^j \hat{\mu}(w_i)} {\sum_{i = 1}^l \hat{\mu}(w_i)
} = \frac { \sum_{i = 1}^j \hat{\mu}(w_i) / \sum_{i = 1}^n \hat{\mu}(w_i) } { \sum_{i = 1}^l \hat{\mu}(w_i) / \sum_{i =
1}^n \hat{\mu}(w_i) } = \frac { \sum_{i = 1}^j \mu(w_i) } { \sum_{i = 1}^l \mu(w_i) } = \frac{P_T(A \wedge B)}{
P_T(B)} $$ This completes the proof. \qed

\begin{example}\label{dp3}[Dice example: upgrading the degree of
belief] \\ {\rm Let us consider program $T_1$ from Example \ref{dice-a-prolog-b} and a new observation $even(d_2)$. To
see the influence of this new evidence on the probability of $d_2$ showing a $4$ we can compute $P_{T_2}(roll(d_2) = 4)$
where $T_2 = T_1 \cup \{ obs(even(d_2)) \}$. Addition of the new observations eliminates those possible worlds of $T_1$
in which the score of $d_2$ is not even. $T_2$ has $18$ possible worlds. Three of them, containing $roll(d_1)=6$, have
the unnormalized probabilities $1/24$ each. The unnormalized probability of every other possible world is $1/40$. Their
measures are respectively $1/12$ and $1/20$, and hence $P_{T_2}(roll(d_2) = 4) = 1/3$. By Proposition \ref{class_def}
the same result can be obtained by computing standard conditional probability $P_{T_1}(roll(d_2) = 4 | even(d_2))$. }
\hfill $\Box$
\end{example}

\noindent Now we consider a number of other types of P-log updates which will take us beyond the updating abilities of
the classical Bayesian approach. Let us start with an update of $T$ by
\begin{equation}\label{update2}
B = \{l_1,\dots,l_n\}.
\end{equation}
where $l$'s are literals.

\st To understand a substantial difference between updating $\Pi$ by $obs(l)$ and by a fact $l$ one should consider the
ASP counterpart $\tau(\Pi)$ of $\Pi$. The first update correspond to expanding $\tau(\Pi)$ by the denial $\leftarrow \no l$ while the second expands $\tau(\Pi)$ by the fact $l$. As discussed in Appendix III constraints and facts play
different roles in the process of forming agent's beliefs about the world and hence one can expect that $\Pi \cup
\{obs(l)\}$ and $\Pi \cup \{l\}$ may have different possible worlds.

\st The following examples show that it is indeed the case.
\begin{example}\label{obs-l-1}[Conditioning on $obs(l)$ versus
conditioning on $l$] \\ {\rm Consider a P-log program $T$

\st $p : \{y_1,y_2\}$.\\ $q : boolean$.\\ $random(p)$.\\ $ \neg q \leftarrow \no q, p = y_1$.\\ $ \neg q \leftarrow p =
y_2$.

\st It is easy to see that no possible world of $T$ contains $q$ and hence $P_{T}(q) = 0$. Now consider the set $B =
\{q,p=y_1\}$ of literals. The program $T \cup obs(B)$ has no possible worlds, and hence the $P_{T \cup obs(B)}(q)$ is
undefined. In contrast, $T \cup B$ has one possible world, $\{q, p = y_1,\dots\}$ and hence $P_{T \cup B}(q) = 1$. The
update $B$ allowed the reasoner to change its degree of belief in $q$ from $0$ to $1$,
a thing impossible in the classical Bayesian framework. } \hfill $\Box$
\end{example}
Note that since for $T$ and $B$ from Example \ref{obs-l-1} we have that $P_T(B)=0$, the classical conditional
probability of $A$ given $B$ is undefined. Hence from the standpoint of classical probability Example \ref{obs-l-1} may
not look very surprising. Perhaps somewhat more surprisingly, $P_{T \cup obs(B)}(A)$ and $P_{T \cup B}(A)$ may be
different even when the classical conditional probability of $A$ given $B$ is defined.
\begin{example}\label{obs-l}[Conditioning on $obs(l)$ versus
conditioning on $l$] \\ {\rm Consider a P-log program $T$

\st $p : \{y_1,y_2\}$.\\ $q : boolean$.\\ $random(p)$.\\ $ q \leftarrow p = y_1$.\\ $ \neg q \leftarrow \no q$.

\st It is not difficult to check that program $T$ has two possible worlds, $W_1$, containing $\{p=y_1, q\}$ and $W_2$,
containing $\{p=y_2, \neg q\}$. Now consider an update $T \cup obs(q)$. It has one possible world, $W_1$. Program $T
\cup \{q\}$ is however different. It has two possible worlds, $W_1$ and $W_3$ where $W_3$ contains $\{p=y_2, q\}$;
$\mu_{T \cup \{q\}}(W_1) = \mu_{T \cup \{q\}}(W_3) = 1/2$. This implies that $P_{T \cup obs(q)}(p=y_1) =1 $ while $P_{T
\cup \{q\}}(p=y_1) = 1/2$. } \hfill $\Box$
\end{example}

\noindent Note that in the above cases the new evidence contained a literal formed by an attribute, $q$, not explicitly
defined as random. Adding a fact $a(t)=y$ to a program for which $a(t)$ is random in some possible world will
usually cause the resulting program to be incoherent.

%The following proposition \ref{eq-cond-1} shows that this is essential, and the difference between
%updating with $obs(l)$ and $l$ is not manifested when $l$ is defined as unconditionally random attribute.

%\begin{proposition}\label{eq-cond-1}[Equivalence Condition 1]\\
%{\rm Let $T$ be a P-log program over signature $\Sigma$ not containing $pr$-atoms, and $B$ a collection of
%$\Sigma$-literals. If
%\begin{enumerate}
%\item all random selection rules of $T$ are of the form \st $random(a(\overline{t}))$, and \item $T \cup obs(B)$ is
%    coherent, and \item for every term $a(\overline{t})$ appearing in literals from $B$ program $T$ contains a
%    random selection rule $random(a(\overline{t}))$,
%\end{enumerate}
%then for every formula $A$
%$$ P_{T \cup B}(A) = P_{T \cup obs(B)}(A)$$
%}
%\end{proposition}

\subsection{ Updates Involving Actions}
Now we discuss updating the agent's knowledge by the effects of deliberate intervening actions, i.e. by a collection of
statements of the form
\begin{equation}\label{update3}
do(B) = \{ do(a(\overline{t}) = y) \ :\ (a(\overline{t}) = y) \in B\}
\end{equation}
As before the update is simply added to the background theory. The results however are substantially different from the
previous updates. The next example illustrates the difference.

\begin{example}\label{rat}[Rat Example]\\
{\rm Consider the following program, $T$, representing knowledge about whether a certain rat will eat arsenic today, and
whether it will die today.

\st $arsenic, death: boolean.$\\ $[\ 1\ ]\ random(arsenic).$\\ $[\ 2 \ ]\ random(death).$\\ $pr(arsenic) = 0.4.$\\
$pr(death \ |_c \ arsenic) = 0.8.$\\ $pr(death \ |_c \ \neg arsenic) = 0.01.$

\st The above program tells us that the rat is more likely to die today if it eats arsenic. Not only that, the intuitive semantics of the {\em pr} atoms expresses that the rat's consumption of arsenic carries information about the cause of
his death (as opposed to, say, the rat's death being informative about the causes of his eating arsenic).

\st An intuitive consequence of this reading is that seeing the rat die raises our suspicion that it has eaten arsenic,
while killing the rat (say, with a pistol) does not affect our degree of belief that arsenic has been consumed. The
following computations show that the principle is reflected in the probabilities computed under our semantics.

\st The possible worlds of the above program, with their unnormalized probabilities, are as follows (we show only {\em
arsenic} and {\em death} literals):

$
\begin{array}{lll}
w_1 : & \{arsenic, death \} . & \hat\mu(w_1) = 0.4 * 0.8 = 0.32\\ w_2: & \{arsenic, \neg death \}. & \hat\mu(w_2)= 0.4
*0.2 = 0.08\\ w_3: & \{\neg arsenic, death \}.& \hat\mu(w_3)= 0.6 * 0.01 = 0.06\\ w_4: & \{\neg arsenic, \neg death \}.
& \hat\mu(w_4) = 0.6 * 0.99 = 0.54
\end{array}
$

\st Since the unnormalized probabilities add up to 1, the respective measures are the same as the unnormalized
probabilities. Hence,

\st $P_T(arsenic ) = \mu(w1) + \mu(w3) = 0.32 + 0.08 = 0.4$

\st To compute probability of $arsenic$ after the observation of $death$ we consider the program $T_1 = T \cup
\{obs(death)\}$

\st The resulting program has two possible worlds, $w_1$ and $w_3$, with unnormalized probabilities as above.
Normalization yields

\st $P_{T_1}(arsenic) = 0.32/(0.32 + 0.06) = 0.8421$

\st Notice that the observation of death raised our degree of belief that the rat had eaten arsenic.

\st To compute the effect of $do(death)$ on the agent's belief in $arsenic$ we augment the original program with the
literal {\em do(death)}. The resulting program, $T_2$, has two answer sets, $w_1$ and $w_3$. However, the action defeats the randomness of death so that $w_1$ has unnormalized probability $0.4$ and $w_3$ has unnormalized probability $0.6$.
These sum to one so the measures are also $0.4$ and $0.6$ respectively, and we get

\st $P_{T_2}(arsenic) = 0.4$

\st Note this is identical to the initial probability $P_T(arsenic)$ computed above. In contrast to the case when the
effect (that is, death) was passively observed, deliberately bringing about the effect did not change our degree of
belief about the propositions relevant to the cause.

\st Propositions relevant to a cause, on the other hand, give equal evidence for the attendant effects whether they are
forced to happen or passively observed. For example, if we feed the rat arsenic, this increases its chance of death,
just as if we had observed the rat eating the arsenic on its own. The conditional probabilities computed under our
semantics bear this out. Similarly to the above, we can compute

\st $P_T(death) = 0.38$\\ $P_{T \cup \{do(arsenic)\}}(death) = 0.8$\\ $P_{T \cup \{obs(arsenic)\}}(death) = 0.8$ }\hfill
$\Box$
\end{example}

\st Note that even though the idea of action based updates comes from Pearl, our treatment of actions is technically
different from his. In Pearl's approach, the semantics of the $do$ operator are given in terms of operations on
graphs (specifically, removing from the graph all directed links leading into the acted-upon variable). In our approach
the semantics of $do$ are given by non-monotonic axioms (\ref{intervene1}) and (\ref{r1.1}) which are introduced by our
semantics as part of the translation of P-log programs into ASP. These axioms are triggered by the addition of
$do(a(\overline{t})=y)$ to the program.

\subsection{More Complex Updates}
Now we illustrate updating the agent's knowledge by more complex regular rules and by probabilistic information.

\begin{example}\label{new-notion}[Adding defined attributes]\\
{\rm In this example we show how updates can be used to expand the vocabulary of the original program. Consider for
instance a program $T_1$ from the die example \ref{dice-a-prolog}. An update, consisting of the rules

\st $max\_score : boolean.$\\ $max\_score \leftarrow score(d_1)=6,score(d_2)=6$.

\st introduces a new boolean attribute, $max\_score$, which holds iff both dice roll the max score. The probability of
$max\_score$ is equal to the product of probabilities of $score(d_1)=6$ and $score(d_2)=6$. } \hfill $\Box$
\end{example}

\begin{example}\label{con_on_rule}[Adding new rules]\\
{\rm Consider a P-log program $T$

\st $d = \{1,2\}$.\\ $p : d \rightarrow boolean$.\\ $random(p(X))$.

\st The program has four possible worlds: $W_1 = \{p(1),p(2)\}$, $W_2 = \{\neg p(1),p(2)\}$, $W_3 = \{p(1),\neg p(2)\}$,
$W_4 = \{\neg p(1), \neg p(2)\}$. It is easy to see that $P_T(p(1)) = 1/2$. What would be the probability of $p(1)$ if
$p(1)$ and $p(2)$ were mutually exclusive? To answer this question we can compute $P_{T \cup B}(p(1))$ where

\st $B = \{ \neg p(1) \leftarrow p(2); \ \ \neg p(2) \leftarrow p(1)\}$.

\st Since $T \cup B$ has three possible worlds, $W_2,W_3,W_4$, we have that $P_{T \cup B}(p(1)) = 1/3$. The new evidence
forced the reasoner to change the probability from $1/2$ to $1/3$. } \hfill $\Box$
\end{example}

\st The next example shows how a new update can force the reasoner to view a previously non-random attribute as random.

\begin{example}\label{rand1}[Adding Randomness]\\
{\rm Consider $T$ consisting of the rules:

\st $a_1, a_2, a_3 \ : \ boolean$.\\ $a_1 \leftarrow a_2.$\\ $a_2 \leftarrow \no \neg a_2.$

\st The program has one possible world, $W = \{a_1,a_2\}$.

\st Now let us update $T$ by $B$ of the form:

\st $\neg a_2.$\\ $random(a_1) \leftarrow \neg a_2.$

\st The new program, $T \cup B$, has two possible worlds\\ $W_1 = \{a_1,\neg a_2\}$ and\\ $W_2 = \{\neg a_1,\neg a_2\}$

\st The degree of belief in $a_1$ changed from $1$ to $1/2$. } \hfill $\Box$
\end{example}
\begin{example}{[Adding Causal Probability]}\\
{\rm Consider programs $T_1$ consisting of the rules:

\st $a : boolean$.\\ $random(a)$.

\st and $T_2$ consisting of the rules:

\st $a : boolean$.\\ $random(a)$.\\ $pr(a) = 1/2$.

\st The programs have the same possible worlds, $W_1 = \{p\}$ and $W_2 = \{\neg p\}$, and the same probability functions
assigning $1/2$ to $W_1$ and $W_2$. The programs however behave differently under simple update $U = \{pr(a) = 1/3\}$.
The updated $T_1$ simply assigns probability $1/3$ and $2/3$ to $W_1$ and $W_2$ respectively. In contrast the attempt to
apply the same update to $T_2$ fails, since the resulting program violates Condition \ref{l2} from \ref{assign-prob}.
This behavior may shed some light on the principle of indifference. According to \cite{kyburg01} ``One of the oddities
of the principle of indifference is that it yields the same sharp probabilities for a pair of alternatives about which
we know nothing at all as it does for the alternative outcomes of a toss of a thoroughly balanced and tested coin''. The
former situation is reflected in $T_1$ where principle of indifference is used to assign default probabilities. The
latter case is captured by $T_2$, where $pr(a) = 1/2$ is the result of some investigation. Correspondingly the update
$U$ of $T_1$ is viewed as simple additional knowledge - the result of study and testing. The same update to $T_2$
contradicts the established knowledge and requires revision of the program. } \hfill $\Box$
\end{example}

\noindent It is important to notice that an update in P-log cannot contradict original background information. An
attempt to add $\neg a$ to a program containing $a$ or to add $pr(a) = 1/2$ to a program containing $pr(a) = 1/3$ would
result in an incoherent program. It is possible to expand P-log to allow such new information (referred to as
``revision'' in the literature) but the exact revision strategy seems to depend on particular situations. If the later
information is more trustworthy then one strategy is justified. If old and new information are ``equally valid'', or the old one is preferable then other strategies are needed. The classification of such revisions and development of the
theory of their effects is however beyond the scope of this paper.

\section{Representing knowledge in P-log}\label{rk} \label{kr-sec}
This section describes several examples of the use of P-log for formalization of logical and probabilistic reasoning. We
do not claim that the problems are impossible to solve without P-log; indeed, with some intelligence and effort, each of
the examples could be treated using a number of different formal languages, or using no formal language at all. The
distinction claimed for the P-log solutions is that they arise directly from transcribing our knowledge of the
problem, in a form which bears a straightforward resemblance to a natural language description of the same knowledge.
The ``straightforwardness'' includes the fact that as additional knowledge is gained about a problem, it can be
represented by adding to the program, rather than by modifying existing code. All of the examples of this section have
been run on our P-log interpreter.

\subsection{Monty Hall problem}\label{monty}

\noindent We start by solving the Monty Hall Problem, which gets its name from the TV game show hosted by Monty Hall (we
follow the description from http://www.io.com/$\sim$kmellis/monty.html). A player is given the opportunity to select one
of three closed doors, behind one of which there is a prize. Behind the other two doors are empty rooms. Once the player
has made a selection, Monty is obligated to open one of the remaining closed doors which does not contain the prize,
showing that the room behind it is empty. He then asks the player if he would like to switch his selection to the other
unopened door, or stay with his original choice. Here is the problem: does it matter if he switches?

\st The answer is YES. In fact switching doubles the player's chance to win. This problem is quite interesting, because
the answer is felt by most people --- often including mathematicians --- to be counter-intuitive. Most people almost
immediately come up with a (wrong) negative answer and are not easily persuaded that they made a mistake. We believe
that part of the reason for the difficulty is some disconnect between modeling probabilistic and non-probabilistic
knowledge about the problem. In P-log this disconnect disappears which leads to a natural correct solution. In other
words, the standard probability formalisms lack the ability to explicitly represent certain non-probabilistic knowledge
that is needed in solving this problem. In the absence of this knowledge, wrong conclusions are made. This example is
meant to show how P-log can be used to avoid this problem by allowing us to specify relevant knowledge explicitly.
Technically this is done by using a random attribute $open$ with the dynamic range defined by regular logic programming
rules.

\st The domain contains the set of three doors and three 0-arity attributes, $selected$, $open$ and $prize$. This will
be represented by the following P-log declarations (the numbers are not part of the declaration; we number statements so
that we can refer back to them):

\st $1.\ \ doors = \{1,2,3\}.$\\ $2.\ \ open, selected, prize : doors.$

\st The regular part contains rules that state that Monty can open any door to a room which is not selected and which
does not contain the prize.

\st $3.\ \ \neg can\_open(D) \leftarrow \ selected = D.$\\ $4.\ \ \neg can\_open(D) \leftarrow \ prize = D.$\\ $5.\ \
can\_open(D) \leftarrow \ not \ \neg can\_open(D).$

\st The first two rules are self-explanatory. The last rule, which uses both classical and default negations, is a
typical ASP representation of the closed world assumption \cite{rei78} --- Monty can open any door except those which
are explicitly prohibited.

\st Assuming the player selects a door at random, the probabilistic information about the three attributes of doors can
be now expressed as follows:

\st $6.\ \ random(prize).$ \\ $7. \ \ random(selected).$\\ $8.\ \ random(open: \{X: can\_open(X) \}).$\\

\st Notice that rule (8) guarantees that Monty selects only those doors which can be opened according to rules (3)--(5).
The knowledge expressed by these rules (which can be extracted from the specification of the problem) is often not
explicitly represented in probabilistic formalisms leading to reasoners (who usually do not realize this) to insist that
their wrong answer is actually correct.

\st The P-Log program $\Pi_{monty0}$ consisting of the logical rules (1)-(8) represents our knowledge of the problem
domain. It has the following 12 possible worlds:

\st $W_1 = \{ selected =1, prize =1, open =2, ... \}$.\\ $W_2 = \{ selected =1, prize =1, open =3, ... \}$.\\ $W_3 = \{
selected =1, prize =2, open =3, ... \}$.\\ $W_4 = \{ selected =1, prize =3, open =2, ... \}$.\\ $W_5 = \{ selected =2,
prize =1, open =3, ... \}$.\\ $W_6 = \{ selected =2, prize =2, open =1, ... \}$.\\ $W_7 = \{ selected =2, prize =2, open
=3, ... \}$.\\ $W_8 = \{ selected =2, prize =3, open =1, ... \}$.\\ $W_9 = \{ selected =3, prize =1, open =2, ... \}$.\\
$W_{10} = \{ selected =3, prize =2, open =1, ... \}$.\\ $W_{11} = \{ selected =3, prize =3, open =1, ... \}$.\\ $W_{12}
= \{ selected =3, prize =3, open =2, ... \}$.

\st According to our definitions they will be assigned various probability measures. For instance, $selected$ has three
possible values in each $W_i$, none of which has assigned probabilities. Hence, according to the definition of the
probability of an atom in a possible world from Section~\ref{assign-prob},

\st $P(W_i, selected = j) = 1/3$

\st for each $i$ and $j$. Similarly for $prize$

\st $P(W_i, prize = j) = 1/3$

\st Consider $W_1$. Since $can\_open(1) \not\in W_1$ the atom $open = 1$ is not possible in $W_1$ and the corresponding
probability $P(W_1, open = 1)$ is undefined. The only possible values of $open$ in $W_1$ are $2$ and $3$. Since they
have no assigned probabilities

\st $P(W_1, open = 2) = PD(W_1, open = 2) = 1/2$

\st $P(W_1, open = 3) = PD(W_1, open = 3) = 1/2$

\st Now consider $W_4$. $W_4$ contains $can\_open(2)$ and no other $can\_open$ atoms. Hence the only possible value of
$open$ in $W_4$ is $2$, and therefore

\st $P(W_4, open = 2) = PD(W_4, open = 2) = 1$

\st The computations of other values of $P(W_i,open=j)$ are similar.

\st Now to proceed with the story, first let us eliminate an orthogonal problem of modeling time by assuming that we
observed that the player has already selected door $1$, and Monty opened door $2$ revealing that it did not contain the
prize. This is expressed as:

\st $obs(selected =1).\ \ obs(open = 2). \ \ obs(prize \not= 2).$

\st Let us refer to the above P-log program as $\Pi_{monty1}$. Because of the observations $\Pi_{monty1}$ has two
possible worlds $W_1$, and $W_4$: the first containing $prize=1$ and the second containing $prize=3$. It follows that

\st $\hat{\mu}(W_1) = P(W_1,selected=1) \times P(W_1,prize=1) \times P(W_1,open=2) = 1/18$

\st $\hat{\mu}(W_4) = P(W_1,selected=1) \times P(W_1,prize=3) \times P(W_1,open=2)=1/9$

\st $\mu(W_1) = \frac{1/18}{1/18+1/9} = 1/3$

\st $\mu(W_4) = \frac{1/9}{1/18+1/9} = 2/3$

\st $P_{\Pi_{monty1}}(prize=1)= \mu(W_1)= 1/3$

\st $P_{\Pi_{monty1}}(prize=3)= \mu(W_4)= 2/3$

\st Changing doors doubles the player's chance to win.

\st Now consider a situation when the player assumes (either consciously or without consciously realizing it) that Monty
could have opened any one of the unopened doors (including one which contains the prize). Then the corresponding program
will have a new definition of $can\_open$. The rules (3--5) will be replaced by

\st $\neg can\_open(D) \leftarrow \ selected = D.$\\ $can\_open(D) \leftarrow \ not \ \neg can\_open(D).$

\st The resulting program $\Pi_{monty2}$ will also have two possible worlds containing $prize=1$ and $prize=3$
respectively, each with unnormalized probability of 1/18, and therefore $P_{\Pi_{monty2}}(prize=1) = 1/2$ and
$P_{\Pi_{monty2}}(prize=3) = 1/2$. In that case changing the door will not increase the probability of getting the
prize.

\st Program $\Pi_{monty1}$ has no explicit probabilistic information and so the possible results of each random
selection are assumed to be equally likely. If we learn, for example, that given a choice between opening doors $2$ and
$3$, Monty opens door $2$ four times out of five, we can incorporate this information by the following statement:

\st $9. \ \ pr(open = 2 \ |_c \ can\_open(2), can\_open(3)) = 4/5$

\st A computation similar to the one above shows that changing doors still increases the players chances to win. Of
course none of the above computations need be carried out by hand. The interpreter will do them automatically.

\st In fact changing doors is advisable as long as each of the available doors can be opened with some positive
probability. Note that our interpreter cannot prove this general result even though it will give proper advice for any
fixed values of the probabilities.

\st The problem can of course be generalized to an arbitrary number $n$ of doors simply by replacing rule (1) with
$doors = \{1,\dots,n\}$.

\subsection{Simpson's paradox}\label{sp}

Let us consider the following story from \cite{pearl99b}: A patient is thinking about trying an experimental drug and
decides to consult a doctor. The doctor has tables of the recovery rates that have been observed among males and
females, taking and not taking the drug.

\begin{tabular}{cccc}
  \hline
  Males: & & &  \\
  \hline
   & & fraction\_of\_population & recovery\_rate \\
  & drug & 3/8 & 60\% \\
  & $\neg$ drug & 1/8 & 70\% \\
  \hline
  Females: & & &  \\
  \hline
  & & fraction\_of\_population & recovery\_rate \\
  & drug & 1/8 & 20\% \\
  & $\neg$ drug & 3/8 & 30\% \\
  \hline
\end{tabular}

\noindent
What should the doctor's advice be? Assuming that the patient is a male, the doctor may attempt to reduce the problem to checking the following inequality involving classical conditional probabilities:

\begin{equation}\label{simp1}
P(recover | male, \neg drug) < \ P(recover | male,drug) \
\end{equation}

\st The corresponding probabilities, if directly calculated from the tables\footnote{If the tables are treated as giving probabilistic information, then we get the following: $P(male) = P(\neg male) = 0.5$. $P(drug) = P(\neg drug) = 0.5$.
$P(recover \mid male, drug) = 0.6$. $P(recover \mid male, \neg drug) = 0.7$.
$P(recover \mid \neg male, drug) = 0.2$. $P(recover \mid \neg male, \neg drug) = 0.3$.
$P(drug \mid male) = 0.75$. $P(drug \mid \neg male) = 0.25$.}, are $0.7$ and $0.6$. The inequality fails, and hence the advice is not to take the drug. A similar argument shows that a female patient should not take the drug.

\st But what should the doctor do if he has forgotten to ask the patient's sex? Following the same reasoning, the doctor might check whether the following inequality is satisfied:
\begin{equation}\label{simp2}
P(recover | \neg drug) < \ P(recover | drug)
\end{equation}

\st This will lead to an unexpected result. $P(recovery | drug) = 0.5$ while $P(recovery | \neg drug) = 0.4$. The drug
seems to be beneficial to patients of unknown sex --- though similar reasoning has shown that the drug is harmful to the patients of known sex, whether they are male or female!

\st This phenomenon is known as Simpson's Paradox: conditioning on $A$ may increase the probability of $B$ among the
general population, while decreasing the probability of $B$ in every subpopulation (or vice-versa). In the current
context, the important and perhaps surprising lesson is that classical conditional probabilities do not faithfully
formalize what we really want to know: {\em what will happen if we do X?} In \cite{pearl99b} Pearl suggests a solution
to this problem in which the effect of deliberate action $A$ on condition $C$ is represented by $P(C | do(A))$ --- a
quantity defined in terms of graphs describing causal relations between variables. Correct reasoning therefore should be
based on evaluating the inequality
\begin{equation}\label{simp3}
P(recover | do(\neg drug)) < \ P(recover | do(drug))
\end{equation}
instead of (\ref{simp2}); this is also what should have been done for (\ref{simp1}).

\st To calculate (\ref{simp3}) using Pearl's approach one needs a causal model and it should be noted that multiple causal models may be consistent with the same statistical data. P-log allows us to express causality and we can determine the probability $P_\Pi$ of a formula $C$ given that action $A$ is performed by computing $P_{\Pi \cup \{do(A)\}}(C)$.

%In Pearl's calculus the result is obtained by deriving a new model of the problem with an altered causal graph. The %first value can be shown by this method to be $0.4$, the second, $0.5$. Hence the drug is most likely to be harmful for %patients of unknown sex as well.

\st Using the tables and added assumption about the direction of causality\footnote{A different assumption about the direction of causality may lead to a different conclusion.} between the variables, we have the values
of the following causal probabilities:

\st $pr(male) = 0.5$.\\ $pr(recover \ |_c \ male,drug) = 0.6$.\\ $pr(recover \ |_c \ male,\neg drug) = 0.7$.\\
$pr(recover \ |_c \ \neg male,drug) = 0.2$.\\ $pr(recover \ |_c \ \neg male,\neg drug) = 0.3$.\\ $pr(drug \ |_c \ male)
= 0.75$.\\ $pr(drug \ |_c \ \neg male) = .25$.

\st These statements, together with declarations:

\st $male, recover, drug \ : \ boolean$\\ $[1]\ random(male)$.\\ $[2]\ random(recover)$.\\ $[3]\ random(drug)$.

\st constitute a P-log program, $\Pi$, that formalizes the story.

\st The program describes eight possible worlds containing various values of the attributes. Each of these worlds and
their unnormalized and normalized probabilities is calculated below.

\st $W_1 = \{ male, recover, drug \}$. $\hat{\mu}(W_1) = 0.5 \times 0.6 \times 0.75 = 0.225$. $\mu(W_1) = 0.225$. \\
$W_2 = \{ male, recover,\neg drug \}$. $\hat{\mu}(W_2) = 0.5 \times 0.7 \times 0.75 = 0.2625$. $\mu(W_2) = 0.2625$. \\
$W_3 = \{ male, \neg recover, drug \}$. $\hat{\mu}(W_3) = 0.5 \times 0.4 \times 0.75 = 0.15$. $\mu(W_3) = 0.15$. \\ $W_4
= \{ male, \neg recover, \neg drug \}$. $\hat{\mu}(W_4) = 0.5 \times 0.3 \times 0.75 = 0.1125$. $\mu(W_4) = 0.1125$. \\
$W_5 = \{ \neg male, recover, drug \}$. $\hat{\mu}(W_5) = 0.5 \times 0.2 \times 0.25 = 0.025$. $\mu(W_5) = 0.025$. \\
$W_6 = \{ \neg male, recover, \neg drug \}$. $\hat{\mu}(W_6) = 0.5 \times 0.3 \times 0.35 = 0.0375$. $\mu(W_6) =
0.0375$. \\ $W_7 = \{ \neg male, \neg recover, drug \}$. $\hat{\mu}(W_7) = 0.5 \times 0.8 \times 0.25 = 0.1$. $\mu(W_7)
= 0.1$. \\ $W_8 = \{ \neg male, \neg recover, \neg drug \}$. $\hat{\mu}(W_8) = 0.5 \times 0.7 \times 0.25 = 0.0875$.
$\mu(W_8) = 0.0875$.

\st Now let us compute $P_{\Pi_1}(recover)$ and $P_{\Pi_2}(recover)$ respectively, where $\Pi_1 = \Pi \cup \{do(drug)\}$
and $\Pi_2 = \Pi \cup \{do(\neg drug)\}$.

\st The four possible worlds of $\Pi_1$ and their unnormalized and normalized probabilities are as follows:

\st $W_1' = \{ male, recover, drug \}$. $\hat{\mu}(W_1') = 0.5 \times 0.6 \times 1 = 0.3$. $\mu(W_1') = 0.3$. \\ $W_3' =
\{ male, \neg recover, drug \}$. $\hat{\mu}(W_3') = 0.5 \times 0.4 \times 1 = 0.2$. $\mu(W_3') = 0.2$. \\ $W_5' = \{
\neg male, recover, drug \}$. $\hat{\mu}(W_5') = 0.5 \times 0.2 \times 1 = 0.1$. $\mu(W_5') = 0.1$. \\ $W_7' = \{ \neg
male, \neg recover, drug \}$. $\hat{\mu}(W_7') = 0.5 \times 0.8 \times 0.1 = 0.4$. $\mu(W_7') = 0.4$.

\st From the above we obtain $P_{\Pi_1}(recover) = .4$.

\st The four possible worlds of $\Pi_2$ and their unnormalized and normalized probabilities are as follows:

\st $W_2' = \{ male, recover,\neg drug \}$. $\hat{\mu}(W_2') = 0.5 \times 0.7 \times 1 = 0.35$. $\mu(W_2') = 0.35$. \\
$W_4' = \{ male, \neg recover, \neg drug \}$. $\hat{\mu}(W_4') = 0.5 \times 0.3 \times 1 = 0.15$. $\mu(W_4') = 0.15$. \\
$W_6' = \{ \neg male, recover, \neg drug \}$. $\hat{\mu}(W_6') = 0.5 \times 0.3 \times 1 = 0.15$. $\mu(W_6') = 0.15$. \\
$W_8' = \{ \neg male, \neg recover, \neg drug \}$. $\hat{\mu}(W_8') = 0.5 \times 0.7 \times 1 = 0.35$. $\mu(W_8') =
0.35$.

\st From the above we obtain $P_{\Pi_2}(recover) = .5$. Hence, if one assumes the direction of causality that we assumed, it is better not to take the drug than to take the drug.

\st Similar calculations also show the following:

\st $P_{\Pi \cup \{obs(male), do(drug)\} }(recover) = 0.6$\\
    $P_{\Pi \cup \{obs(male), do(\neg drug)\} }(recover) = 0.7$

\st
    $P_{\Pi \cup \{obs(\neg male), do(drug)\} }(recover) = 0.2$\\
    $P_{\Pi \cup \{obs(\neg male), do(\neg drug)\} }(recover) = 0.3$

\st I.e., if we know the person is male then it is better not to take the drug than to take the drug, the same if we know the person is female, and both agree with the case when we do not know if the person is male or female.

\st The example shows that queries of the form ``What will happen if we {\emph do} $X$?'' can be easily stated
and answered in P-log. The necessary P-log reasoning is nonmonotonic and is based on rules (\ref{intervene1}) and
(\ref{r1.1}) from the definition of $\tau(\Pi)$.

\subsection{A Moving Robot }\label{robot}
Now we consider a formalization of a problem whose original version, not containing probabilistic reasoning, first
appeared in \cite{iwan02}.

\st There are rooms, say $r_0,r_1,r_2$ reachable from the current position of a robot. The rooms can be open or closed.
The robot cannot open the doors. It is known that the robot navigation is usually successful. However, a malfunction can
cause the robot to go off course and enter any one of the open rooms.

\st We want to be able to use our formalization for correctly answering simple questions about the robot's behavior
including the following scenario: the robot moved toward open room $r_1$ but found itself in some other room. What room
can this be?

\st As usual we start with formalizing this knowledge. We need the initial and final moments of time, the rooms, and the
actions.

\st $time = \{0,1\}$ \ \ \ \ $rooms = \{r_0,r_1,r_2\}.$

\st We will need actions:

\st $go\_in \ :\ rooms \rightarrow boolean.$

\st $break \ : \ boolean$.

\st $ab \ : \ boolean$.

\st The first action consists of the robot {\em attempting} to enter the room $R$ at time step $0$. The second is an
exogenous breaking action which may occur at moment $0$ and alter the outcome of this attempt. In what follows,
(possibly indexed) variables $R$ will be used for rooms.

\st A state of the domain will be modeled by a time-dependent attribute, $in$, and a time independent attribute $open$.
(Time dependent attributes and relations are often referred to as \emph{fluents}).

\st $open : rooms \rightarrow boolean.$

\st $ in : time \rightarrow rooms.$

\st The description of dynamic behavior of the system will be given by the rules below:

\st First two rules state that the robot navigation is usually successful, and a malfunctioning robot constitutes an
exception to this default.

\st 1. $in(1)=R \leftarrow go\_in(R), \no ab.$ \\ 2. $ab \leftarrow break.$

\st The random selection rule (3) below plays a role of a (non-deterministic) causal law. It says that a malfunctioning
robot can end up in any one of the open rooms.

\st 3. $[r] \mbox{ random}(in(1) : \{R : open(R)\}) \leftarrow go\_in(R), break.$

\st We also need inertia axioms for the fluent $in$.

\st 4a. $in(1)=R \leftarrow in(0)=R, \no \neg in(1)=R.$\\ 4b. $in(1)\not=R \leftarrow in(0)\not=R, \no in(1)=R.$

\st Finally, we assume that only closed doors will be specified in the initial situation. Otherwise doors are assumed to
be open.

\st 5. $ open(R) \leftarrow \no \neg open(R).$

\st The resulting program, $\Pi_0$, completes the first stage of our formalization. The program will be used in
conjunction with a collection $X$ of atoms of the form $in(0) = R$, $\neg open(R)$, $go\_in(R)$, $break$ which satisfies
the following conditions: $X$ contains at most one atom of the form $in(0) = R$ (robot cannot be in two rooms at the
same time); $X$ has at most one atom of the form $go\_in(R)$ (robot cannot move to more than one room); $X$ does not
contain a pair of atoms of the form $\neg open(R)$, $go\_in(R)$ (robot does not attempt to enter a closed room); and $X$
does not contain a pair of atoms of the form $\neg open(R)$, $in(0) = R$ (robot cannot start in a closed room). A set
$X$ satisfying these properties will be normally referred to as a {\em valid input} of $\Pi_0$.

\st Given an input $X_1 = \{go\_in(r_0)\}$ the program $\Pi_0 \cup X_1$ will correctly conclude $in(1)=r_0$. The input
$X_2 = \{go\_in(r_0),break\}$ will result in three possible worlds containing $in(1)=r_0, in(1)=r_1$ and $in(1)=r_2$
respectively. If, in addition, we are given $\neg open(r_2)$ the third possible world will disappear, etc.

\st Now let us expand $\Pi_0$ by some useful probabilistic information. We can for instance consider $\Pi_1$ obtained
from $\Pi_0$ by adding:

\st 8. $pr_{r}(in(1)=R \ |_c \ \ go\_in(R), break) = 1/2.$

\st (Note that for any valid input $X$, Condition 3 of Section \ref{assign-prob} is satisfied for $\Pi_1 \cup X$ , since
rooms are assumed to be open by default and no valid input may contain $\neg open(R)$ and $go\_in(R)$ for any $R$.)
Program $T_1=\Pi_1 \cup X_1$ has the unique possible world which contains $in(1)=r_0$. Hence, $P_{T_1}(in(1)=r_0) = 1$.

\st Now consider $T_2 = \Pi_1 \cup X_2$. It has three possible worlds: $W_0$ containing $in(1)=r_0$, and $W_1, W_2$
containing $in(1)=r_1$ and $in(1)=r_2$ respectively. $P_{T_2}(W_0)$ is assigned a probability of $1/2$, while
$P_{T_2}(W_1) = P_{T_2}(W_2) = 1/4$ by default. Therefore $P_{T_2}(in(1)=r_0) = 1/2$. Here the addition of $break$ to
the knowledge base changed the degree of reasoner's belief in $in(1) = r_0$ from $1$ to $1/2$. This is not possible in
classical Bayesian updating, for two reasons. First, the prior probability of $break$ is 0 and hence it cannot be
conditioned upon. Second, the prior probability of $in(1)=r_0$ is 1 and hence cannot be diminished by classical
conditioning. To account for this change in the classical framework requires the creation of a new probabilistic model.
However, each model is a function of the underlying background knowledge; and so P-log allows us to represent the change
in the form of an update.

%(P.S. THIS SHOWS THAT WE REALLY DEFINED A NEW NOTION OF CONDITIONAL %PROBABILITY WHICH EXTENDS THE PREVIOUS ONE).

\subsection{Bayesian squirrel} \label{sec-squirrel}
In this section we consider an example from \cite{hilborn97} used to illustrate the notion of Bayesian learning. One
common type of learning problem consists of selecting from a set of models for a random phenomenon by observing repeated
occurrences of the phenomenon. The Bayesian approach to this problem is to begin with a ``prior density'' on the set of
candidate models and update it in light of our observations.

\st As an example, Hilborn and Mangel describe the Bayesian squirrel. The squirrel has hidden its acorns in one of two
patches, say Patch 1 and Patch 2, but can't remember which. The squirrel is 80\% certain the food is hidden in Patch 1.
Also, it knows there is a 20\% chance of finding food per day when it looking in the right patch (and, of course, a 0\%
probability if it's looking in the wrong patch).

\st To represent this knowledge in P-log's program $\Pi$ we introduce sorts

\st $patch = \{p1,p2\}$.

\st $day = \{1\dots n\}$.

\st (where $n$ is some constant, say, $5$)

\st and attributes

\st $hidden\_in : patch$.

\st $found : patch * day \rightarrow boolean$.

\st $look : day \rightarrow patch$.

\st Attribute $hidden\_in$ is always random. Hence we include

\st $[r_1]\mbox{ random } (hidden\_in)$.

\st $found$ is random only if the squirrel is looking for food in the right patch, i.e. we have

\st $[r_2]\mbox{ random } (found(P,D)) \leftarrow hidden\_in=P,look(D)=P$.

\st The regular part of the program consists of the closed world assumption for $found$:

\st $\neg found(P,D) \leftarrow \no found(P,D)$.

\st Probabilistic information of the story is given by statements:

\st $pr_{r_1}(hidden\_in = p1) = 0.8$.

\st $pr_{r_2}(found(P,D)) = 0.2$.

\st This knowledge, in conjunction with description of the squirrel's activity, can be used to compute probabilities of
possible outcomes of the next search for food.

\st Consider for instance program $\Pi_1 = \Pi \cup \{do(look(1) = p_1)\}$. The program has three possible worlds

\st $W^{1}_1 = \{look(1)=p_1,hidden\_in = p_1, found(p_1,1),\dots\}$,

\st $W^{1}_2 = \{look(1)=p_1,hidden\_in = p_1, \neg found(p_1,1),\dots \}$,

\st $W^{1}_3 = \{look(1)=p_1,hidden\_in = p_2, \neg found(p_1,1),\dots \}$,

\st with probability measures $\mu(W_1) = 0.16$, $\mu(W_2) = 0.64$, $ \mu(W_3) = 0.2$.

\st As expected

\st $P_{\Pi_1} (hidden\_in = p_1) = 0.8$, and

\st $P_{\Pi_1}(found(p_1,1)) = 0.16$.

\st Suppose now that the squirrel failed to find its food during the first day, and decided to continue her search in
the first patch next morning.

\st The failure to find food in the first day should decrease the squirrel's degree of belief that the food is hidden in
patch one, and consequently decreases her degree of belief that she will find food by looking in the first patch again.
This is reflected in the following computation:

\st Let \st $\Pi_2 = \Pi_1 \cup \{obs(\neg found(p_1,1)), do(look(2) = p_1)\}$.

\st The possible worlds of $\Pi_2$ are:

\st $W^{2}_1= W \cup \{hidden\_in = p_1,look(2)=p_1, found(p_1,2)\dots \}$,

\st $W^{2}_2=W \cup \{hidden\_in = p_1,look(2)=p_1, \neg found(p_1,2) \dots\}$,

\st $W^{2}_3=W \cup \{hidden\_in = p_2,look(2)=p_1, \neg found(p_1,2) \dots\}$.

\st where $W = \{look(1)=p_1,\neg found(p_1,1)\}.$

\st Their probability measures are

\st $\mu(W^{2}_1) = .128 / .84 = .152$, $\mu(W^{2}_2) = .512 / .84 = . 61$, $\mu(W^{2}_3) = .2 / .84 = .238$.

\st Consequently,

\st $P_{\Pi_2}(hidden\_in = p_1) = 0.762$, and $P_{\Pi_2}(found (p_1,2)) = 0.152$, and so on.

\st After a number of unsuccessful attempts to find food in the first patch the squirrel can come to the conclusion that
food is probably hidden in the second patch and change her search strategy accordingly.

\st Notice that each new experiment changes the squirrel's probabilistic model in a non-monotonic way. That is, the set
of possible worlds resulting from each successive experiment is not merely a subset of the possible worlds of the
previous model. The program however is changed only by the addition of new actions and observations. Distinctive
features of P-log such as the ability to represent observations and actions, as well as conditional randomness, play an
important role in allowing the squirrel to learn new probabilistic models from experience.

\st For comparison, let's look at a classical Bayesian solution. If the squirrel has looked in patch 1 on day 1 and not
found food, the probability that the food is hidden in patch 1 can be computed as follows. First, by Bayes Theorem,

$$P(hidden=1 | \neg found(p_1,1)) =
\frac{P(\neg find(1) | \ hidden\_in = p_1) * P(hidden\_in =p_1)} {P(\neg found(p_1,1))}$$ The denominator can then be
rewritten as follows:

\st $P(\neg find(1))$

\st $= P(\neg found(p_1,1) \cup hidden\_in = 1) + P(\neg found(p_1,1) \cup hidden\_in = p_2)$

\st $= P(\neg found(p_1,1) |\ hidden\_in = p_1) * P(hidden\_in = p_1) + P(hidden\_in = p_2)$

\st $= 0.8 * 0.8 + 0. 2$

\st $= 0.84$

\st Substitution yields
$$ P(hidden\_in = p_1 |\ \neg found(p_1,1)) = (0.8 * 0.8)/0.84 = 0.762 $$

\st {\bf Discussion}

\noindent Note that the classical solution of this problem does not contain any formal mention of the action $look(2) =
p_1$. We must keep this informal background knowledge in mind when constructing and using the model, but it does not
appear explicitly. To consider and compare distinct action sequences, for example, would require the use of several
intuitively related but formally unconnected models. In Causal Bayesian nets (or P-log), by contrast, the corresponding
programs may be written in terms of one another using the do-operator.

\st In this example we see that the use of the do-operator is not strictly necessary. Even if we were choosing between
sequences of actions, the job could be done by Bayes theorem, combined with our ability to juggle several intuitively
related but formally distinct models. In fact, if we are very clever, Bayes Theorem itself is not necessary --- for we
could use our intuition of the problem to construct a new probability space, implicitly based on the knowledge we want
to condition upon.

\st However, though not necessary, Bayes theorem is very useful --- because it allows us to formalize subtle reasoning
{\em within} the model which would otherwise have to be performed in the informal process of {\em creating} the
model(s). Causal Bayesian nets carry this a step further by allowing us to formalize interventions in addition to
observations, and P-log yet another step by allowing the formalization of logical knowledge about a problem or family of
problems. At each step in this hierarchy, part of the informal process of creating a model is replaced by a formal
computation.

\st As in this case, probabilistic models are often most easily described in terms of the conditional probabilities of
effects given their causes. From the standpoint of traditional probability theory, these conditional probabilities are
viewed as constraints on the underlying probability space. In a learning problem like the one above, Bayes Theorem can
then be used to relate the probabilities we are given to those we want to know: namely, the probabilities of
evidence-given-models with the probabilities of models-given-evidence. This is typically done without describing or even
thinking about the underlying probability space, because the given conditional probabilities, together with Bayes
Theorem, tell us all we need to know. The use of Bayes Theorem in this manner is particular to problems with a certain
look and feel, which are loosely classified as ``Bayesian learning problems''.

\st From the standpoint of P-log things are somewhat different. Here, all probabilities are defined with respect to
bodies of knowledge, which include models and evidence in the single vehicle of a P-log program. Within this framework,
Bayesian learning problems do not have such a distinctive quality. They are solved by writing down what we know and
issuing a query, just like any other problem. Since P-log probabilities satisfy the axioms of probability, Bayes Theorem
still applies and could be useful in calculating the P-log probabilities by hand. On the other hand, it is possible and
even natural to approach these problems in P-log without mentioning Bayes Theorem. This would be awkward in ordinary
mathematical probability, where the derivation of models from knowledge is considerably less systematic.

\subsection{Maneuvering the Space Shuttle} \label{shuttleRCS}

So far we have presented a number of small examples to illustrate various features of P-log. In this section we outline
our use of P-log for an industrial size application: diagnosing faults in the reactive control system (RCS) of the Space
Shuttle.

\st To put this work in the proper perspective we need to briefly describe the history of the project. The RCS actuates
the maneuvering of the shuttle.  It consists of fuel and oxidizer tanks, valves, and other plumbing needed to provide
propellant to the shuttle's maneuvering jets. It also includes electronic circuitry, both to control the valves in the
fuel lines, and to prepare the jets to receive firing commands. To perform a maneuver, Shuttle controllers (i.e.,
astronauts and/or mission controllers) must find a sequence of commands which delivers propellant from tanks to a proper combination of jets.

\st Answer Set Programming (without probabilities) was successfully used to design and implement the decision support
system USA-Adviser \cite{bal01,bgnw02}, which, given information about the desired maneuver and the current state of the system (including its known faults), finds a plan allowing the controllers to achieve this task. In addition the USA-Advisor is capable of diagnosing an unexpected behavior of the system. The success of the project hinged on Answer Set Prolog's ability to describe controllers' knowledge about the system, the corresponding operational procedures, and a fair amount of commonsense knowledge. It also depended on the existence of efficient ASP solvers.

\st The USA-Advisor is build on a detailed but straightforward model of the RCS. For instance, the hydraulic part of the RCS can be viewed as a graph whose nodes are labeled by tanks containing propellant, jets, junctions of pipes, etc. Arcs of the graph are labeled by valves which can be opened or closed by a collection of switches. The graph is described by a collection of ASP atoms of the form $connected(n_1,v,n_2)$ (valve $v$ labels the arc from $n_1$ to $n_2$) and
$controls(s,v)$ (switch $s$ controls valve $v$). The description of the system may also contain a collection of faults,
e.g. a valve can be \emph{stuck}, it can be \emph{leaking}, or have a \emph{bad\_circuitry}. Similar models exists for
electrical part of the RCS and for the connection between electrical and hydraulic parts. Overall, the system is rather
complex, in that it includes $12$ tanks, $44$ jets, $66$ valves, $33$ switches, and around $160$ computer commands
(computer-generated signals).

\st In addition to simple description of the RCS,  USA-Advisor contains knowledge of the system's dynamic behavior. For
instance the axiom
$$
\begin{array}{lll}
\neg faulty(C) & \leftarrow & \no may\_be\_faulty(C).
\end{array}
$$
says that in the absence of evidence to the contrary, components of the RCS are assumed to be working properly (Note
that concise representation of this knowledge depends critically on the ability of ASP to represent defaults.) the axioms
$$\begin{array}{lll}
h(state(S,open),T+1) & \leftarrow & occurs(flip(S),T),\\
                         &            & h(state(S,closed),T),\\
                         &            & \neg faulty(S).\\
h(state(S,closed),T+1) & \leftarrow & occurs(flip(S),T),\\
                         &            & h(state(S,open),T),\\
                         &            & \neg faulty(S).\\
\end{array}
$$
express the direct effect of an action of flipping switch $S$. Here $state$ is a function symbol with the first
parameter ranging over switches and valves and the second ranging over their possible states; $flip$ is a function
symbol whose parameter is of type switch. Predicate symbol $h$ (holds) has the first parameters ranging over fluents and the second one ranging over time-steps; two parameters of $occur$ are of type $action$ and $time$-$step$ respectively.
Note that despite the presence of function symbols our typing guarantees finiteness of the Herbrand universe of the
program. The next axiom describes the connections between positions of switches and valves.
$$
\begin{array}{lll}
h(state(V,P),T) & \leftarrow & controls(S,V),\\
                &            & h(state(S,P),T),\\
                &            & \neg fault(V,stuck).
\end{array}
$$
A recursive rule
$$
\begin{array}{lll}
h(pressurized(N_2),T) & \leftarrow & connected(N_1,V,N_2),\\
                      &            & h(pressurized(N_1),T),\\
                      &            & h(state(V,open),T),\\
                      &            & \neg fault(V,leaking).
\end{array}
$$
describes the relationship between the values of relation $pressurized(N)$ for neighboring nodes. (Node $N$ is
$pressurized$ if it is reached by a sufficient quantity of the propellant). These and other axioms, which are rooted in
a substantial body of research on actions and change, describe a comparatively complex effect of a simple $flip$
operation which propagates the pressure through the system.

\st The plan to execute a desired maneuver can be extracted by a simple procedural program from answer sets of a program
$\Pi_s \cup PM$, where $\Pi_s$ consists of the description of the RCS and its dynamic behavior, and $PM$ is a ``planning
module,'' containing a statement of the goal (i.e., maneuver), and rules needed for ASP-based planning. Similarly, the
diagnosis can be  extracted from answer sets of $\Pi_s \cup DM$, where the diagnostic module $DM$ contains unexpected
observations, together with axioms needed for the ASP diagnostics.

\st After the development of the original USA-Advisor, we learned that, as could be expected, some faults of the RCS
components are more likely than others, and, moreover, reasonable estimates of the probabilities of these faults can be
obtained and utilized for finding the most probable diagnosis of unexpected observations. Usually this is done under the
assumption that the number of multiple faults of the system is limited by some fixed bound.

\st P-log allowed us to write software for finding such diagnoses. First we needed to expand $\Pi_s$ by the
corresponding declarations including the statement
$$
\begin{array}{lll}
[r(C,F)]\ random(fault(C,F)) & \leftarrow & may\_be\_faulty(C).
\end{array}
$$
where $may\_be\_fault(C,F)$ is a boolean attribute which is true if component $C$ may (or may not) have a fault of type
$F$. The probabilistic information about faults is given by the $pr$-atoms, e.g.
$$pr_{r(V,stack)}(fault(V,stuck) |_c \ may\_be\_faulty(V)) = 0.0002.$$
etc. To create a probabilistic model of our system, the ASP diagnostic module finds components relevant to the agent's
unexpected observations, and adds them to $DM$ as a collection of atoms of the form $may\_be\_faulty(c)$. Each possible
world of the resulting program (viz., $P = \Pi_s \cup DM$)  uniquely corresponds to a possible explanation of the
unexpected observation. The system finds possible worlds with maximum probability measure and returns diagnoses defined
by these worlds, where an ``explanation'' consists of all atoms of the form $fault(c,f)$ in a given possible world. \st
This system works very efficiently if we assume that maximum number, $n$, of faults in the explanation does not exceed
two (a practically realistic assumption for our task). If $n$ equals $3$ the computation is substantially slower. There
are two obvious ways to improve efficiency of the system:  improve our prototype implementation of P-log or reduce
the number of possibly faulty components returned by the original diagnostic program or both. We are currently working in both of these directions. It is of course important to realize that the largest part of all these computations is not probabilistic and is performed by the ASP solvers, which are themselves quite mature. However the conceptual blending of ASP with probabilities achieved by P-log allowed us to successfully express our probabilistic knowledge, and to define
the corresponding probabilistic model, which was essential for the success of the project.

\section{Proving Coherency of P-log Programs} \label{property-sec}

In this section we state theorems which can be used to show the coherency of P-log programs. The proofs of the theorems
are given in an Appendix I. We begin by introducing terminology which makes it easier to state the theorems.

\subsection{Causally ordered programs}
Let $\Pi$ be a (ground) P-log program with signature $\Sigma$.

\begin{definition}{[Dependency relations]}\\
{\rm Let $l_1$ and $l_2$ be literals of $\Sigma$. We say that
\begin{enumerate}
\item $l_1$ is {\em immediately dependent} on $l_2$, written as $l_1 \leq_i l_2$, if there is a rule $r$ of $\Pi$ such
    that $l_1$ occurs in the head of $r$ and $l_2$ occurs in the $r$'s body; \item $l_1$ {\em depends} on $l_2$,
    written as $l_1 \leq l_2$, if the pair $\langle l_1, l_2 \rangle$ belongs to the reflexive transitive closure of relation
    $l_1 \leq_i l_2$; \item An attribute term $a_1(\overline{t}_1)$ {\em depends} on an attribute term
    $a_2(\overline{t}_2)$ if there are literals $l_1$ and $l_2$ formed by $a_1(\overline{t}_1)$ and
    $a_2(\overline{t}_2)$ respectively such that $l_1$ depends on $l_2$.  \hfill $\Box$
\end{enumerate}
}
\end{definition}
\begin{example}\label{dependency}[Dependency]\\
{\rm Let us consider a version of the Monty Hall program consisting of rules (1) -- (9) from Subsection \ref{monty}. Let
us denote it by $\Pi_{monty3}$. From rules (3) and (4) of this program we conclude that $\neg can\_open(d)$ is
immediately dependent on $prize = d$ and $selected = d$ for every door $d$. By rule (5) we have that for every $d \in
doors$, $can\_open(d)$ is immediately dependent on $\neg can\_open(d)$. By rule (8), $open = d_1$ is immediately
dependent on $can\_open(d_2)$ for any $d_1,d_2 \in doors$. Finally, according to (9), $open = 2$ is immediately
dependent on $can\_open(2)$ and $can\_open(3)$. Now it is easy to see that an attribute term $open$ depends on itself
and on attribute terms $prize$ and $selected$, while each of the latter two terms depends only on itself. } \hfill
$\Box$
\end{example}
\begin{definition} {[Leveling function]}\\
{\rm A {\em leveling function}, $|\ |$, of $\Pi$ maps attribute terms of $\Sigma$ onto a set $[0,n]$ of natural numbers.
It is extended to other syntactic entities over $\Sigma$ as follows:
$$|a(\overline{t}) = y| = |a(\overline{t}) \not= y| =
|\no a(\overline{t}) = y| = |\no a(\overline{t}) \not= y| = |a(\overline{t})|$$ We'll often refer to $|e|$ as the {\em
rank} of $e$. Finally, if $B$ is a set of expressions then $|B| = max(\{|e| : e \in B\})$. } \hfill $\Box$
\end{definition}
\begin{definition} {[Strict probabilistic leveling and reasonable programs]} \\
{\rm A leveling function $|\ |$ of $\Pi$ is called {\em strict probabilistic} if
\begin{enumerate}
\item no two random attribute terms of $\Sigma$ have the same level under $|\ |$ ; \item for every random selection
    rule $\ \ [ r ] \ random(a(\overline{t}) : \{y : p(y)\}) \leftarrow B \ \ $ of $\Pi$ we have\\
    $|a(\overline{t})=y| \leq |\{p(y) : y \in range(a)\} \cup B |$; \item for every probability atom $\ \
    pr_r(a(\overline{t}) = y \ |_c \ B ) \ \ $ of $\Pi$ we have $\ |a(\overline{t})| \leq | B |$;

\item if $a_1(\overline{t}_1)$ is a random attribute term, $a_2(\overline{t}_2)$ is a non-random attribute term, and
    $a_2(\overline{t}_2)$ depends on $a_1(\overline{t}_1)$ then $\ |a_2(\overline{t}_2)| \geq
    |a_1(\overline{t}_1)|$.
\end{enumerate}
A P-log program $\Pi$ which has a strict probabilistic leveling function is called {\em reasonable}. } \hfill $\Box$
\end{definition}
\begin{example}\label{reasonable}[Strict probabilistic leveling for Monty Hall]\\
{\rm Let us consider the program $\Pi_{monty3}$ from Example \ref{dependency} and a leveling function

\st $|prize| = 0$\\ $|selected| = 1$\\ $|can\_open(D)| = 1$\\ $|open| = 2$

\st We claim that this leveling is a strict probabilistic levelling. Conditions (1)--(3) of the definition can be
checked directly. To check the last condition it is sufficient to notice that for every $D$ the only random attribute
terms on which non-random attribute term $can\_open(D)$ depends are $selected$ and $prize$. } \hfill $\Box$
\end{example}

\st Let $\Pi$ be a reasonable program with signature $\Sigma$ and leveling $|\ |$, and let $a_1(t_1),\dots,a_n(t_n)$ be
an ordering of its random attribute terms induced by $|\ |$. By $L_i$ we denote the set of literals of $\Sigma$ which do not depend on literals formed by $a_j(t_j)$ where $i \leq j$. $\Pi_i$ for $1 \leq i \leq n+1$ consists of all
declarations of $\Pi$, along with the regular rules, random selection rules, actions, and observations of $\Pi$ such
that every literal occurring in them belongs to $L_i$. We'll often refer to $\Pi_1,\dots,\Pi_{n+1}$ as a $|\ |$-induced
structure of $\Pi$.

\begin{example}\label{induced}[Induced structure for Monty Hall]\\
{\rm To better understand this construction let us consider a leveling function $|\ |$ from Example \ref{reasonable}. It induces the following ordering of random attributes of the corresponding program.

\st $a_1 = prize$.\\ $a_2 = selected$.\\ $a_3 = open$.

\st The corresponding languages are

\st $L_1 = \emptyset$\\ $L_2 = \{prize = d : d \in doors\}$\\ $L_3 = L_2 \cup \{selected = d : d \in doors\} \cup
\{can\_open(d) : d \in doors\} \cup \{\neg can\_open(d) : d \in doors\}$\\ $L_4 = L_3 \cup \{open = d : d \in doors\}$

\st Finally, the induced structure of the program is as follows (numbers refer to the numbered statements of Subsection
\ref{monty}.

\st $\Pi_1 = \{1,2\}$\\ $\Pi_2 = \{1,2,6\}$\\ $\Pi_3 = \{1, \dots, 7\}$\\ $\Pi_4 = \{1,\dots, 8\}$ }\hfill $\Box$
\end{example}
Before proceeding we introduce some terminology.

\begin{definition}\label{act-atr}{[Active attribute term]} \\
{\rm If there is $y$ such that $a(\overline{t}) = y$ is possible in $W$ with respect to $\Pi$, we say that
$a(\overline{t})$ is {\em active} in $W$ with respect to $\Pi$. } \hfill $\Box$
\end{definition}
\begin{definition}\label{c-order}{[Causally ordered programs]} \\
{\rm Let $\Pi$ be a P-log program with a strict probabilistic leveling $|\ |$ and let $a_i$ be the $i^{th}$ random
attribute of $\Pi$ with respect to $|\ |$. We say that $\Pi$ is {\em causally ordered} if

\begin{enumerate}
\item $\Pi_1$ has exactly one possible world; \item if $W$ is a possible world of $\Pi_i$ and atom
    $a_{i}(\overline{t}_{i}) = y_0$ is possible in $W$ with respect to $\Pi_{i+1}$ then the program $W \cup
    \Pi_{i+1} \cup obs(a_{i}(\overline{t}_{i}) = y_0)$ has exactly one possible world; and \item if $W$ is a
    possible world of $\Pi_i$ and $a_{i}(\overline{t}_{i})$ is not active in $W$ with respect to $\Pi_{i+1}$ then
    the program $W \cup \Pi_{i+1}$ has exactly one possible world. \hfill $\Box$
\end{enumerate}
}
\end{definition}

\st Intuitively, a program is causally ordered if (1) all nondeterminism in the program results from random selections,
and (2) whenever a random selection is active in a given possible world, the possible outcomes of that selection are not constrained in that possible world by logical rules or other random selections. The following is a simple example of a
program which is not causally ordered, because it violates the second condition. By comparison with Example~\ref{e11}, it also illustrates the difference between the statements $a$ and $pr(a)=1$.

\begin{example}\label{e11a}[A non-causally ordered programs]\\
{\rm Consider the P-log program $\Pi$ consisting of:

\st $1.\ a \ : \ boolean$.\\ $2. \ random \ a$.\\ $3. \ a.$

\st The only leveling function for this program is $|a| = 0$, hence $L_1 = \emptyset$ while $L_2 = \{a,\neg a\}$; and
$\Pi_1 = \{1\}$ while $\Pi_2 = \{1,2,3\}$. Obviously, $\Pi_1$ has exactly one possible world, namely $W_1 = \emptyset$.
Both literals, $a$ and $\neg a$ are possible in $W_1$ with respect to $\Pi_2$. However, $W_1 \cup \Pi_2 \cup obs(\neg
a)$ has no possible worlds, and hence the program does not satisfy Condition 2 of the definition of {\em causally
ordered.}

\st Now let us consider program $\Pi^\prime$ consisting of rules (1) and (2) of $\Pi$ and the rules

\st $b \leftarrow \no \neg b, a$.\\ $\neg b \leftarrow \no b, a$.

\st The only strict probabilistic leveling function for this program maps $a$ to $0$ and $b$ to $1$. The resulting
languages are $L_1 = \emptyset$ and $L_2 = \{a,\neg a, b, \neg b\}$. Hence $\Pi^{\prime}_1 = \{1\}$ and $\Pi^{\prime}_2
= \Pi^\prime$. As before, $W_1$ is empty and $a$ and $\neg a$ are both possible in $W_1$ with respect to
$\Pi^{\prime}_2$. It is easy to see that program $W_1 \cup \Pi^{\prime}_2 \cup obs(a)$ has two possible worlds, one
containing $b$ and another containing $\neg b$. Hence Condition 2 of the definition of causally ordered is again
violated.

\st Finally, consider program $\Pi^{\prime\prime}$ consisting of rules:

\st $1.\ \ a,b \ : \ boolean$.\\ $2. \ \ random(a)$. \\ $3. \ \ random(b) \leftarrow a$. \\ $4. \ \ \neg b \leftarrow
\neg a$.\\ $5. \ \ c \leftarrow \neg b$.\\ $6. \ \ \neg c$.

\st It is easy to check that $c$ immediately depends on $\neg b$, which in turn immediately depends on $a$ and $\neg a$.
$b$ immediately depends on $a$. It follows that any strict probabilistic leveling function for this program will lead to
the ordering $a,b$ of random attribute terms. Hence $L_1 = \{\neg c\}$, $L_2 = \{ \neg c, a, \neg a\}$, and $L_3 = L_2
\cup \{b, \neg b, c\}$. This implies that $\Pi^{\prime\prime}_1 = \{1, 6\}$, $\Pi^{\prime\prime}_2 = \{1,2, 6\}$, and
$\Pi^{\prime\prime}_3 = \{1,\dots,6\}$. Now consider a possible world $W = \{ \neg c, \neg a\}$ of
$\Pi^{\prime\prime}_2$. It is easy to see that the second random attribute, $b$, is not active in $W$ with respect to
$\Pi^{\prime\prime}_3$, but $W \cup \Pi^{\prime\prime}_3$ has no possible world. This violates Condition 3 of causally
ordered.

\st Note that all the above programs are consistent. A program whose regular part consists of the rule $p \leftarrow \no p$ is neither causally ordered nor consistent. Similarly, the program obtained from $\Pi$ above by adding the atom
$pr(a) = 1/2$ is neither causally ordered nor consistent. } \hfill $\Box$
\end{example}

\begin{example}\label{monty-is-causally ordered}[Monty Hall program is causally ordered]\\
{\rm We now show that the Monty Hall program $\Pi_{monty3}$ is causally ordered. We use the strict probabilistic
leveling and induced structure from the Examples~\ref{reasonable} and ~\ref{induced}. Obviously, $\Pi_1$ has one possible world $W_1 = \emptyset$.
The atoms possible in $W_1$ with respect to $\Pi_2$ are $prize = 1$, $prize = 2$, $prize = 3$. So we must check
Condition 2 from the definition of causally ordered for every atom $prize = d$ from this set. It is not difficult to
show that the translation $\tau(W_1 \cup \Pi_2 \cup obs(prize = d))$ is equivalent to logic program consisting of the
translation of declarations into Answer Set Prolog along with the following rules:

\st $prize(1) \mbox{ \bf or } prize(2) \mbox{ \bf or } prize(3)$.\\ $\neg prize(D_1) \leftarrow prize(D_2), D_1 \not=
D_2$.\\ $\leftarrow obs(prize(1)), \no prize(d)$.\\ $obs(prize(d))$.

\st where $D_1$ and $D_2$ range over the doors. Except for the possible occurrences of observations this program is
equivalent to

\st $\neg prize(D_1) \leftarrow prize(D_2), D_1 \not= D_2$.\\ $prize(d)$.

\st which has a unique answer set of the form
\begin{equation}\label{anset}
\{prize(d),\neg prize(d_1),\neg prize(d_2)\}
\end{equation}
(where $d_1$ and $d_2$ are the other two doors besides $d$). \st Now let $W_2$ be an arbitrary possible world of
$\Pi_2$, and $l$ be an atom possible in $W_2$ with respect to $\Pi_3$. To verify Condition 2 of the definition of
causally ordered for $i=2$, we must show that $W_2 \cup \Pi_2 \cup obs(l)$ has exactly one answer set. It is easy to see
that $W_2$ must be of the form (\ref{anset}), and $l$ must be of the form $selected = d^{\prime}$ for some door
$d^{\prime}$.

\st Similarly to above, the translation of $W_2 \cup \Pi_3 \cup obs(selected(d^{\prime}))$ has the same answer sets (except
for possible occurrences of observations) as the program consisting of $W_2$ along with the following rules:

\st $selected(d^{\prime})$.\\ $\neg selected(D_1) \leftarrow selected(D_2), D_1 \not= D_2$.\\ $\neg can\_open(D)
\leftarrow selected(D)$.\\ $\neg can\_open(D) \leftarrow prize(D)$.\\ $can\_open \leftarrow \no \neg can\_open(D)$.

\st If negated literals are treated as new predicate symbols we can view this program as stratified. Hence the program
obtained in this way has a unique answer set. This means that the above program has at {\em most} one answer set; but it is easy to see it is consistent and so it has exactly one. It now follows that Condition 2 is satisfied for $i=2$.

\st Checking Condition 2 for $i=3$ is similar, and completes the proof. } \hfill $\Box$
\end{example}

\st ``Causal ordering'' is one of two conditions which together guarantee the coherency of a P-log program. Causal
ordering is a condition on the logical part of the program. The other condition --- that the program must be ``unitary'' --- is a condition on the $pr$-atoms. It says that, basically, assigned probabilities, if any, must be given in a way
that permits the appropriate assigned and default probabilities to sum to 1. In order to define this notion precisely,
and state the main theorem of this section, we will need some terminology.

\st Let $\Pi$ be a ground P-log program containing the random selection rule
$$[r] \ \ random(a(t) : \{Y :p(Y)\}) \leftarrow K.$$

\st We will refer to a ground pr-atom
$$pr_r(a(t) = y \ |_c \ B) = v.$$
as a \emph{pr-atom indexing $r$}. We will refer to $B$ as the \emph{body} of the $pr$-atom. We will refer to $v$ as the
\emph{probability assigned by the $pr$-atom}.

\st Let $W_1$ and $W_2$ be possible worlds of $\Pi$ satisfying $K$. We say that $W_1$ and $W_2$ are \emph{probabilistically
equivalent with respect to $r$} if
\begin{enumerate}
\item for all $y$, $p(y) \in W_1$ if and only if $p(y) \in W_2$, and \item For every $pr$-atom $q$ indexing $r$,
    $W_1$ satisfies the body of $q$ if and only if $W_2$ satisfies the body of $q$.
\end{enumerate}

\st A \emph{scenario} for $r$ is an equivalence class of possible worlds of $\Pi$ satisfying $K$, under probabilistic
equivalence with respect to $r$.

\begin{example}\label{scenarios}[Rat Example Revisited]\\
{\rm Consider the program from Example \ref{rat} involving the rat, and its possible worlds $W_1, W_2, W_3, W_4$. All
four possible worlds are probabilistically equivalent with respect to Rule [1]. With respect to Rule [2] $W_1$ is
equivalent to $W_2$, and $W_3$ is equivalent to $W_4$. Hence Rule [2] has two scenarios, $\{W_1, W_2\}$ and $\{W_3,
W_4\}$. } \hfill $\Box$
\end{example}

\st $range(a(t), r, s)$ will denote the set of possible values of $a(t)$ in the possible worlds belonging to scenario
$s$ of rule $r$. This is well defined by (1) of the definition of probabilistic equivalence w.r.t. $r$. For example, in
the rat program, $range(death, 2, \{W_1,W_2\}) = \{true,false\}$.

\st Let $s$ be a scenario of rule $r$. A $pr$-atom $q$ indexing $r$ is said to be {\em active in s} if every possible
world of $s$ satisfies the body of $q$.

\st For a random selection rule $r$ and scenario $s$ of $r$, let $at_r(s)$ denote the set of probability atoms which are
active in $s$. For example, $at_2(\{W_1,W_2\})$ is the singleton set $\{pr(death \ |_c \ arsenic) = 0.8\}$.

\st
\begin{definition}{[Unitary Rule]} \label{unitary-rule}\\
{\rm Rule $r$ is {\em unitary in $\Pi$}, or simply $unitary$, if for every scenario $s$ of $r$, one of the following
conditions holds:
\begin{enumerate}
\item For every $y$ in $range(a(t), r, s)$, $at_r(s)$ contains a $pr$-atom of the form $pr_r(a(t) = y \ |_c \ B) =
    v$, and moreover the sum of the values of the probabilities assigned by members of $at_r(s)$ is 1; or \item
    There is a $y$ in $range(a(t), r, s)$ such that $at_r(s)$ contains no $pr$-atom of the form $pr_r(a(t) = y \ |_c
    \ B) = v$, and the sum of the probabilities assigned by the members of $at_r(s)$ is less than or equal to 1. \hfill $\Box$
\end{enumerate}
}
\end{definition}

\st
\begin{definition}{[Unitary Program]}\\
{\rm A P-log program is {\em unitary} if each of its random selection rules is unitary. } \hfill $\Box$
\end{definition}

\begin{example}\label{rat-is-unitary}[Rat Example Revisited]\\
{\rm Consider again Example \ref{rat} involving the rat. There is clearly only one scenario, $s_1$, for the Rule $[\ 1 \ ]\ random(arsenic)$, which consists of all possible worlds of the program. $at_1(s_1)$ consists of the single $pr$-atom $pr(arsenic)=0.4$. Hence the scenario satisfies Condition 2 of the definition of unitary.

\st We next consider the selection rule $[\ 2\ ] random(death).$ There are two scenarios for this rule: $s_{arsenic}$,
consisting of possible worlds satisfying $arsenic$, and its complement $s_{noarsenic}$. Condition 2 of the definition of
unitary is satisfied for each element of the partition. } \hfill $\Box$
\end{example}
We are now ready to state the main theorem of this section, the proof of which will be given in Appendix I.

\st
\begin{theorem}\label{consisconds}[Sufficient Conditions for Coherency]\\
{\rm Every causally ordered, unitary P-log program is coherent. } \hfill $\Box$
\end{theorem}
Using the above examples one can easily check that the rat, Monty Hall, and Simpson's examples are causally ordered and
unitary, and therefore coherent.

\st For the final result of this section, we give a result that P-log can represent the probability distribution of any
finite set of random variables each taking finitely many values in a classical probability space.

\begin{theorem}\label{th2}[Embedding Probability Distributions in P-log]\\
{\rm Let $x_1, \dots, x_n$ be a nonempty vector of random variables, under a classical probability $P$, taking
finitely many values each. Let $R_i$ be the set of possible values of each $x_i$, and assume $R_i$ is nonempty for each
$i$. Then there exists a coherent P-log program $\Pi$ with random attributes $x_1, \dots, x_n$ such that for
every vector $r_1,\dots,r_n$ from $R_1 \times .. \times R_n,$ we have
\begin{equation}\label{in-th2-1}
P(x_1=r_1,\dots,x_n = r_n) = P_{\Pi}(x_1=r_1,\dots, x_n = r_n)
\end{equation}
} \hfill $\Box$
\end{theorem}

\st The proof of this theorem appears in Appendix I. It is a corollary of this theorem that if $B$ is a finite Bayesian
network, each of whose nodes is associated with a random variable taking finitely many possible values, then there is a
P-log program which represents the same probability distribution as $B$. This by itself is not surprising, and could be
shown trivially by considering a single random attribute whose values range over possible states of a given Bayes net.
Our proof, however, shows something more -- namely, that the construction of the P-log program corresponds
straightforwardly to the graphical structure of the network, along with the conditional densities of its variables given their parents in the network. Hence any Bayes net can be represented by a P-log program which is ``syntactically
isomorphic'' to the network, and preserves the intuitions present in the network representation.

\section{Relation with other work}\label{other_work}

As we mention in the first sentence of this paper, the motivation behind developing P-log is to have a knowledge
representation language that allows natural and elaboration tolerant representation of common-sense knowledge involving
logic and probabilities. While some of the other probabilistic logic programming languages such as
\cite{poole93,poole2000} and \cite{vvb04,joost07z} have similar goals, many other probabilistic logic programming
languages have ``statistical relational learning (SRL)'' \cite{getoor07} as one of their main goals and as a result they perhaps consciously sacrifice on the knowledge representation dimensions. In this section we describe the approaches in
\cite{poole93,poole2000} and \cite{vvb04,joost07z} and compare them with P-log. We also survey many other works on
probabilistic logic programming, including the ones that have SRL as one of their main goals, and relate them to P-log
from the perspective of representation and reasoning.

\subsection{Relation with Poole's work}

Our approach in this paper has a lot of similarity (and many differences) with the works of Poole
\cite{poole93,poole2000}. To give a somewhat detailed comparison, we start with some of the definitions from
\cite{poole93}.

\subsubsection{Overview of Poole's probabilistic Horn abduction}

In Poole's probabilistic Horn abduction (PHA), disjoint declarations are an important component. We start with their
definition. (In our adaptation of the original definitions we consider the grounding of the theory, so as to make it
simpler.)

\begin{definition}
{\rm Disjoint declarations are of the form $disjoint([h_1:p_1 \ ; \ \ldots \ ; \ h_n:p_n])$, where $h_i$s are different
ground atoms -- referred to as hypotheses or assumables, $p_i$s are real numbers and $p_1 + \ldots + p_n = 1$.} \hfill $\Box$
\end{definition}

\noindent We now define a PHA theory.

\begin{definition}
{\rm A probabilistic Horn abduction (PHA) theory is a collection of definite clauses and disjoint declarations such that no atom occurs in two disjoint declarations.} \hfill $\Box$
\end{definition}

\st Given a PHA theory $T$, the facts of $T$, denoted by $F_T$ consists of

\begin{itemize}

\item the collection of definite clauses in $T$, and

\item for every disjoint declarations $D$ in $T$, and for every $h_i$ and $h_j$, $i \neq j$ in $D$, integrity
    constraints of the form:

\st $\leftarrow h_i, h_j$.

\end{itemize}

\st The hypotheses of $T$, denoted by $H_T$, is the set of $h_i$ occurring in disjoint declarations of $T$.

\st The prior probability of $T$ is denoted by $P_T$ and is a function $H_T \rightarrow [0,1]$ defined such that
$P_T(h_i) = p_i$ whenever $h_i : p_i$ is in a disjoint declaration of $T$. Based on this prior probability and the
assumption, denoted by {\bf (Hyp-independent)}, that hypotheses that are consistent with $F_T$ are (probabilistically)
independent of each other, we have the following definition of the joint probability of a set of hypotheses.

\begin{definition}
{\rm Let $\{h_1, \ldots, h_k\}$ be a set of hypotheses where each $h_i$ is from a disjoint declaration. Then, their
joint probability is given by $P_T(h_1) \times \ldots \times P_T(h_k)$. } \hfill $\Box$
\end{definition}

\st Poole \cite{poole93} makes the following additional assumptions about $F_T$ and $H_T$:

\begin{enumerate}

\item {\bf (Hyp-not-head)} There are no rules in $F_T$ whose head is a member of $H_T$. (i.e., hypotheses do not
    appear in the head of rules.)

\item {\bf (Acyclic-definite)} $F_T$ is acyclic.

\item {\bf (Completion-cond)} The semantics of $F_T$ is given via its Clark's completion.

\item {\bf (Body-not-overlap)} The bodies of the rules in $F_T$ for an atom are mutually exclusive. (i.e., if we
    have $a \leftarrow B_i$ and $a \leftarrow B_j$ in $F_T$, where $i\neq j$, then $B_i$ and $B_j$ can not be true
    at the same time.)

\end{enumerate}

\noindent Poole presents his rationale behind the above assumptions, which he says makes the language weak. His
rationale is based on his goal to develop a simple extension of Pure Prolog (definite logic programs) with Clark's
completion based semantics, that allows interpreting the number in the hypotheses as probabilities. Thus he restricts
the syntax to disallow any case that might make the above mentioned interpretation difficult.

\st We now define the notions of explanations and minimal explanations and use it to define the probability distribution
and conditional probabilities embedded in a PHA theory.

\begin{definition}
{\rm If $g$ is a formula, an explanation of $g$ from $\langle F_T, H_T \rangle$ is a subset $D$ of $H_T$ such that $F_T
\cup D \models g$ and $F_T \cup D$ has a model.

\st A minimal explanation of $g$ is an explanation of $g$ such that no strict subset is an explanation of $g$ } \hfill
$\Box$
\end{definition}
Poole proves that under the above mentioned assumptions, if $min\_expl(g,T)$ is the set of all minimal explanations of
$g$ from $\langle F_T, H_T \rangle$ and $Comp(T)$ is the Clark's completion of $F_T$ then
$$Comp(T) \ \ \models \ \ \ (g \equiv \bigvee_{e_i \ \in \ min\_expl(g,T)} e_i)$$
\begin{definition}
{\rm For a formula $g$, its probability $P$ with respect to a PHA theory $T$ is defined as:

$$P(g) = \sum_{e_i \ \in \ min\_expl(g,T)} P_T(e_i)$$
} \hfill $\Box$
\end{definition}

\noindent Conditional probabilities are defined using the standard definition:

$$P (\alpha|\beta) = \frac{P (\alpha \wedge \beta)}{P(\beta)}$$

\st We now relate his work with ours.

\subsubsection{Poole's PHA compared with P-log}

\begin{itemize}

\item The disjoint declarations in PHA have some similarity with our random declarations. Following are some of the
    main differences:

\begin{itemize}

\item {\bf (Disj1)} The disjoint declarations assign probabilities to the hypothesis in that declaration. We use
    probability atoms to specify probabilities, and our random declarations do not mention probabilities.

\item {\bf (Disj2)} Our random declarations have conditions. We also specify a range for the attributes. Both
    the conditions and attributes use predicates that are defined using rules. The usefulness of this is evident
    from the formulation of the Monty Hall problem where we use the random declaration

\st $random(open: \{X: can\_open(X) \} )$.

\st The disjoint declarations of PHA theories do not have conditions and they do not specify ranges.

\item {\bf (Disj3)} While the hypotheses in disjoint declarations are arbitrary atoms, our random declarations
    are about attributes.

\end{itemize}

\item {\bf (Pr-atom-gen)} Our specification of the probabilities using pr-atoms is more general than the probability
    specified using disjoint declarations. For example, in specifying the probabilities of the dices we say:

\st $pr(roll(D) =Y \ |_c \ owner(D) = john) = 1/6$.

\item {\bf (CBN)} We directly specify the conditional probabilities in causal Bayes nets, while in PHA only prior
    probabilities are specified. Thus expressing a Bayes network is straightforward in P-log while in PHA it would
    necessitate a transformation.

\item {\bf (Body-not-overlap2)} Since Poole's PHA assumes that the definite rules with the same hypothesis in the
    head have bodies that can not be true at the same time, many rules that can be directly written in our formalism
    need to be transformed so as to satisfy the above mentioned condition on their bodies.

\item {\bf (Gen)} While Poole makes many a-priori restrictions on his rules, we follow the opposite approach and
    initially do not make any restrictions on our logical part. Thus we have an unrestricted logical knowledge
    representation language (such as ASP or CR-Prolog) at our disposal. We define a semantic notion of consistent
    P-log programs and give sufficiency conditions, more general than Poole's restrictions, that guarantee
    consistency.

\item {\bf (Obs-do)} Unlike us, Poole does not distinguish between doing and observing.

\item {\bf (Gen-upd)} We consider very general updates, beyond an observation of a propositional fact or an action
    that makes a propositional fact true.

\item {\bf (Prob-def)} Not all probability numbers need be explicitly given in P-log. It has a default mechanism to
    implicitly assume certain probabilities that are not explicitly given. This often makes the representation
    simpler.

\item Our probability calculation is based on possible worlds, which is not the case in PHA, although Poole's later
    formulation of Independent Choice Logic \cite{poo97,poole2000} (ICL) uses possible worlds.

\end{itemize}

\subsubsection{Poole's ICL compared with P-log}

Poole's Independent Choice Logic \cite{poo97,poole2000} refines his PHA by replacing the set of disjoint declarations by
a choice space (where individual disjoint declarations are replaced by alternatives, and a hypothesis in an individual
disjoint declaration is replaced by an atomic choice), by replacing definite programs and their Clark's completion
semantics by acyclic normal logic programs and their stable model semantics, by enumerating the atomic choices across
alternatives and defining possible worlds\footnote{Poole's possible worlds are very similar to ours except that he
explicitly assumes that the possible worlds whose core would be obtained by the enumeration, can not be eliminated by
the acyclic programs through constraints. We do not make such an assumption, allow elimination of such cores, and if
elimination of one or more (but not all) possible worlds happen then we use normalization to redistribute the
probabilities.} rather than using minimal explanation based abduction, and in the process making fewer assumptions. In
particular, the assumption {\bf Completion-cond} is no longer there, the assumption {\bf Body-not-overlap} is only made
in the context of being able to obtain the probability of a formula $g$ by adding the probabilities of its explanations,
and the assumption {\bf Acyclic-definite} is relaxed to allow acyclic normal programs; while the assumptions {\bf
Hyp-not-head} and {\bf Hyp-independent} remain in slightly modified form by referring to atomic choices across
alternatives rather than hypothesis across disjoint statements. Nevertheless, most of the differences between PHA and
P-log carry over to the differences between ICL and P-log. In particular, all the differences mentioned in the previous
section -- with the exception of {\bf Body-not-overlap2} -- remain, modulo the change between the notion of hypothesis
in PHA to the notion of atomic choices in ICL. \comment{Michael's comment: (a) Have a look at the ``loosing the keys''
example in his JLP 2000 paper.\\ (b) There is 1-1 correspondence between explanation of formula g and models Pi + obs(g)
in our formalism, where Pi is obtained naturally.}

\subsection{LPAD : Logic programming with annotated disjunctions}\label{sec-LPAD}

In recent work \cite{vvb04} Vennekens et al. have proposed the LPAD formalism. An LPAD program consists of rules of the
form:

\st $(h_1 : \alpha_1) \vee \ldots \vee (h_n : \alpha_n) \leftarrow b_1, \ldots, b_m$

\st where $h_i$'s are atoms, $b_i$s are atoms or atoms preceded by {\em not}, and $\alpha_i$s are real numbers in the
interval $[0,1]$, such that $\sum_{i=1}^{n} \alpha_i = 1$.

\st An LPAD rule instance is of the form:

\st $h_i \leftarrow b_1, \ldots, b_m$.

\st The associated probability of the above rule instance is then said to be $\alpha_i$.

\st An instance of an LPAD program $P$ is a (normal logic program) $P'$ obtained as follows: for each rule in $P$
exactly one of its instance is included in $P'$, and nothing else is in $P'$. The associated probability of an instance
$P'$, denoted by $\pi(P')$, of an LPAD program is the product of the associated probability of each of its rules.

\st An LPAD program is said to be sound if each of its instances has a 2-valued well-founded model. Given an LPAD program
$P$, and a collection of atoms $I$, the probability assigned to $I$ by $P$ is given as follows:

$$\pi_P(I) = \sum_{P' \mbox{ is an instance of } P \mbox{ and } I \mbox{ is the well-founded model of } P'} \ \pi(P')$$

\noindent The probability of a formula $\phi$ assigned by an LPAD program $P$ is then defined as:

$$\pi_P(\phi) = \sum_{\phi \mbox{ is satisfied by } I } \ \pi_P(I)$$

\subsubsection{Relating LPAD with P-log}

LPAD is richer in syntax than PHA or ICL in that its rules (corresponding to disjoint declarations in PHA and a choice
space in ICL) may have conditions. In that sense it is closer to the random declarations in P-log. Thus, unlike PHA and
ICLP, and similar to P-log, Bayes networks can be expressed in LPAD fairly directly. Nevertheless LPAD has some
significant differences with P-log, including the following:

\begin{itemize}

\item The goal of LPAD is to provide succinct representations for probability distributions. Our goals are broader,
    {\em viz}, to combine probabilistic and logical reasoning. Consequently P-log is logically more expressive, for
    example containing classical negation and the ability to represent defaults.

\item The ranges of random selections in LPAD are taken directly from the heads of rules, and are therefore static.
    The ranges of of selections in P-log are dynamic in the sense that they may be different in different possible
    worlds. For example, consider the representation

\st $random(open : \{X : can\_open(X)\})$.

\st of the Monty Hall problem. It is not clear how the above can be succinctly expressed in LPAD.

\end{itemize}

\subsection{Bayesian logic programming:
}

A Bayesian logic program (BLP) \cite{kersting07} has two parts, a logical part and a set of conditional probability
tables. The logical part of the BLP consists of clauses (referred to as BLP clauses) of the form:

\st $H \ | \ A_1, \ldots, A_n$

\st where $H, A_1, \ldots, A_n$ are (Bayesian) atoms which can take a value from a given domain associated with the
atom. Following is an example of a BLP clause from \cite{kersting07}:

\st $burglary(X) \ | \ neighborhood(X)$.

\st Its corresponding domain could be, for example, $D_{burglary} = \{yes, no\}$, and $D_{neighbourhood} = \{bad,
average, good \}$.

\st Each BLP clause has an associated conditional probability table (CPT). For example, the above clause may have the
following table:

\begin{tabular}{|c|c|c|c|}
\hline neighborhood(X) & & burglary(X) & burglary(X) \\ & & yes & no \\ \hline bad & & 0.6 & 0.4 \\ average & & 0.4 &
0.6 \\ good & & 0.3 & 0.7 \\ \hline
\end{tabular}

\st A ground BLP clause is similar to a ground logic programming rule. It is obtained by substituting variables with
ground terms from the Herbrand universe. If the ground version of a BLP program is acyclic, then a BLP can be considered
as representing a Bayes network with possibly infinite number of nodes. To deal with the situation when the ground
version of a BLP has multiple rules with the same atom in the head, the formalisms allows for specification of {\em
combining rules} that specify how a set of ground BLP rules (with the same ground atom in the head) and their CPT can be
combined to a single BLP rule and a single associated CPT.

\st The semantics of an acyclic BLP is thus given by the characterization of the corresponding Bayes net obtained as
described above.

\subsubsection{Relating BLPs with P-log}

The aim of BLPs is to enhance Bayes nets so as to overcome some of the limitations of Bayes nets such as difficulties
with representing relations. On the other hand like Bayes nets, BLPs are also concerned about statistical relational
learning. Hence the BLP research is less concerned with general knowledge representation than P-log is, and this is the
source of most of the differences in the two approaches. Among the resulting differences between BLP and P-log are:

\begin{itemize}

\item In BLP every ground atoms represents a random variable. This is not the case in P-log.

\item In BLP the values the atoms can take are fixed by their domain. This is not the case in P-log where through
    the random declarations an attribute can have different domains under different conditions.

\item Although the logical part of a BLP looks like a logic program (when one replaces $|$ by the connective
    $\leftarrow$), its meaning is different from the meaning of the corresponding logic program. Each BLP clause is
    a compact representation of multiple logical relationships with associated probabilities that are given using a
    conditional probability table.

\item In BLP one can specify a combining rule. We do not allow such specification.

\end{itemize}

\noindent The ALTERID language of \cite{breese90,wellman92} is similar to BLPs and has similar differences with P-log.

\subsubsection{Probabilistic knowledge bases}

Bayesian logic programs mentioned in the previous subsections was inspired by the probabilistic knowledge bases (PKBs)
of \cite{ngo97}. We now give a brief description of this formalism.

\st In this formalism each predicate represents a set of similar random variables. It is assumed that each predicate has
at least one attribute representing the value of random attributes made up of that predicate. For example, the random
variable $Colour$ of a car $C$ can be represented by a 2-ary predicate $color(C,Col)$, where the first position takes
the id of particular car, and the second indicates the color (say, blue, red, etc.) of the car $C$.

\st A probabilistic knowledge base consists of three parts:

\begin{itemize}

\item A set of probabilistic sentences of the form:

\st $pr(A_0 \ | \ A_1, \ldots, A_n) = \alpha$, where $A_i$s are atoms.

\item A set of value integrity constraints of the form:

\st $EXCLUSIVE(p, a_1, \ldots, a_n)$, where $p$ is a predicate, and $a_i$s are values that can be taken by random
variables made up of that predicate.

\item A set of combining rules.

\end{itemize}

\noindent The combining rules serve similar purpose as in Bayesian logic programs. Note that unlike Bayesian logic
programs that have CPTs for each BLP clause, the probabilistic sentences in PKBs only have a single probability
associated with it. Thus the semantic characterization is much more complicated. Nevertheless the differences between
P-log and Bayesian logic programs also carry over to PKBs.

\subsection{Stochastic logic programs}

A Stochastic logic program (SLP) \cite{muggleton95} $P$ is a collection of clauses of the form

\st $p \ : \ A \leftarrow B_1, \ldots, B_n$

\st where $p$ (referred to as the probability label) belongs to $[0,1]$, and $A, B_1, \ldots B_n$ are atoms, with
the requirements that (a) $A \leftarrow B_1, \ldots, B_n$ is range restricted and (b) for each predicate symbol $q$ in
$P$, the probability labels for all clauses with $q$ in the head sum to 1.

\st The probability of an atom $g$ with respect to an SLP $P$ is obtained by summing the probability of the various
SLD-refutation of $\leftarrow g$ with respect to $P$, where the probability of a refutation is computed by multiplying
the probability of various choices; and doing appropriate normalization. For example, if the first atom of a subgoal
$\leftarrow g'$ unifies with the head of stochastic clauses $p_1 \ : \ C_1$, $\ldots$, $p_m \ : \ C_m$, and the
stochastic clause $p_i \ : \ C_i$ is chosen for the refutation, then the probability of this choice is $\frac{p_i}{p_1 +
\cdots + p_m}$.

\subsubsection{Relating SLPs with P-log}

SLPs, both as defined in the previous section and as in \cite{cussens99}, are very different from P-log both in its
syntax and semantics.

\begin{itemize}

\item To start with, SLPs do not allow the `not' operator, thus limiting the expressiveness of the logical part.

\item In SLPs all ground atoms represent random variables. This is not the case in P-log.

\item In SLPs probability computation is through computing probabilities of refutations, a top down approach. In
    P-log it is based on the possible worlds, a bottom up approach.

\end{itemize}

\noindent The above differences also carry over to probabilistic constraint logic programs \cite{riezler98,santos03}
that generalize SLPs to Constraint logic programs (CLPs).

\subsection{Probabilistic logic programming}

The probabilistic logic programming formalisms in \cite{ng92,ng94,dekdek04} and \cite{luka98} take the representation of
uncertainty to another level. In these two approaches they are interested in classes of probability distributions and
define inference methods for checking if certain probability statements are true with respect to all the probability
distributions under consideration. To express classes of probability distributions, they use intervals where the
intuitive meaning of $p : [\alpha, \beta]$ is that the probability of $p$ is in between $\alpha$ and $\beta$. We now
discuss the two formalisms in \cite{ng92,ng94,dekdek04} and \cite{luka98} in further detail. We refer to the first one
as NS-PLP (short for Ng-Subrahmanian probabilistic logic programming) and the second one as L-PLP (short for Lukasiewicz
probabilistic logic programming).

\subsubsection{NS-PLP}

A simple NS-PLP program \cite{ng92,ng94,dekdek04} is a finite collection of p-clauses of the form

\st $A_0:[\alpha_0, \beta_0] \leftarrow A_1:[\alpha_1, \beta_1], \ldots, A_n:[\alpha_n, \beta_n]$.

\st where $A_0, A_1, \ldots, A_n$ are atoms, and $[\alpha_i, \beta_i] \subseteq [0,1]$. Intuitively, the meaning of the
above rule is that if the probability of $A_1$ is in the interval $[\alpha_1, \beta_1]$, ..., and the probability of
$A_n$ is in the interval $[\alpha_n, \beta_n]$ then the probability of $A_0$ is in the interval $[\alpha_0, \beta_0]$.

\st The goal behind the semantic characterization of an NS-PLP program $P$ is to obtain and express the set of
(probabilistic) p-interpretations (each of which maps possible worlds, which are subsets of the Herbrand Base, to a
number in [0,1]), $Mod(P)$, that satisfy all the p-clauses in the program. Although initially it was thought that
$Mod(P)$ could be computed through the iteration of a fixpoint operator, recently \cite{dekdek04} shows that this is not
the case and gives a more complicated way to compute $Mod(P)$. In particular, \cite{dekdek04} shows that for many NS-PLP
programs, although its fixpoint, a mapping from the Herbrand base to an interval in $[0,1]$, is defined, it does not
represent the set of satisfying p-interpretations.

\st Ng and Subrahmanian \cite{ng94} consider more general NS-PLP programs where $A_i$s are `basic formulas' (which are
conjunction or disjunction of atoms) and some of $A_1, \ldots, A_n$ are preceded by the $not$ operator. In presence of
$not$ they give a semantics inspired by the stable model semantics. But in this case an NS-PLP program may have multiple
stable formula functions, each of which map formulas to intervals in $[0,1]$. While a single stable formula function can
be considered as a representation of a set of p-interpretations, it is not clear what a set of stable formula functions
correspond to. Thus NS-PLP programs and their characterization is very different from P-log and it is not clear if one
is more expressive than the other.

\subsubsection{L-PLP}

An L-PLP program \cite{luka98} is a finite set of L-PLP clauses of the form

\st $(H \ | \ B) [c_1, c_2]$

\st where $H$ and $B$ are conjunctive formulas and $c_1 \leq c_2$.

\st Given a probability distribution $Pr$, an L-PLP clause of the above form is said to be in $Pr$ if $c_1 \leq Pr(H|B)
\leq c_2$. $Pr$ is said to be a {\em model} of an L-PLP program $\pi$ if each clause in $\pi$ is true in $Pr$. $(H \ | \
B) [c_1, c_2]$ is said to be a logical consequence of an L-PLP program $\pi$ denoted by $\pi \models (H \ | \ B) [c_1,
c_2]$ if for all models $Pr$ of $\pi$, $(H \ | \ B) [c_1, c_2]$ is in $Pr$. A notion of tight entailment, and correct
answer to ground and non-ground queries of the form $\exists (H \ | \ B) [c_1, c_2]$ is then defined in \cite{luka98}.
In recent papers Lukasiewicz and his colleagues generalize L-PLPs in several ways and define many other notions of
entailment.

\st In relation to NS-PLP programs, L-PLP programs have a single interval associated with an L-PLP clause and an L-PLP
clause can be thought of as a constraint on the corresponding conditional probability. Thus, although `logic' is used in
L-PLP programs and their characterization, it is not clear whether any of the `logical knowledge representation'
benefits are present in L-PLP programs. For example, it does not seem that one can define the values that a random
variable can take, in a particular possible world, using an L-PLP program.

\subsection{PRISM: Logic programs with distribution semantics}
Sato in \cite{sato95} proposes the notion of ``logic programs with distribution\ semantics,'' which he refers to as
PRISM as a short form for\ ``PRogramming In Statistical Modeling.'' Sato starts with a possibly infinite collection of
ground atoms, $F$, the set $\Omega_F$ of all interpretations of $F$\footnote{By interpretation $I_F$ of $F$ we mean an
arbitrary subset of $F$. Atom $A \in F$ is \emph{true} in $I_F$ iff $A \in I_F$.}, and a completely additive probability measure $P_F$ which quantifies the likelihood of interpretations. $P_F$ is defined on some fixed $\sigma$ algebra of
subsets of $\Omega_F$.

\st
In Sato's framework interpretations of $F$ can be used in conjunction with a Horn logic program $R$, which contains no
rules whose heads unify with atoms from $F$. Sato's logic program is a triple, $\Pi = \langle F, P_F,R\rangle$. The
semantics of $\Pi$ are given by a collection $\Omega_{\Pi}$ of possible worlds and the probability measure $P_{\Pi}$. A
set $M$ of ground atoms in the language of $\Pi$ belongs to $\Omega_{\Pi}$ iff $M$ is a minimal Herbrand model of a
logic program $I_F \cup R$ for some interpretation $I_F$ of $F$. The completely additive probability measure of
$P_{\Pi}$ is defined as an extension of $P_F$.

\st Given a specification of $P_F$, the formalism provides a powerful tool for defining complex probability measures,
including those which can be described by Bayesian nets and Hidden Markov models. The emphasis of the original work by
Sato and other PRISM related research seems to be on the use of the formalism for design and investigation of efficient
algorithms for statistical learning. The goal is to use the pair $DB = \langle F,R \rangle$ together with observations
of atoms from the language of $DB$ to learn a suitable probability measure $P_F$.

\st P-log and PRISM share a substantial number of common features. Both are declarative languages capable of
representing and reasoning with logical and probabilistic knowledge. In both cases logical part of the language is
rooted in logic programming. There are also substantial differences. PRISM seems to be primarily intended as ``a
powerful tool for building complex statistical models'' with emphasis of using these models for statistical learning. As a result PRISM allows infinite possible worlds, and has the ability of learning statistical parameters embedded in its
inference mechanism. The goal of P-log designers was to develop a knowledge representation language allowing natural,
elaboration tolerant representation of commonsense knowledge involving logic and probabilities. Infinite possible worlds and algorithms for statistical learning were not a priority. Instead the emphasis was on greater logical power provided
by Answer Set Prolog, on causal interpretation of probability, and on the ability to perform and differentiate between
various types of updates. In the near future we plan to use the PRISM ideas to expand the semantics of P-log to allow
infinite possible worlds. Our more distant plans include investigation of possible adaptation of PRISM statistical
learning algorithms to P-log.

\subsection{Other approaches}
So far we have discussed logic programming approaches to integrate logical and probabilistic reasoning.
Besides them, the paper \cite{vv00} proposes a notion where the theory has two parts, a logic
programming part that can express preferences and a joint probability distribution.
The probabilities are then used in determining the priorities of the alternatives.

\st Besides the logic programming based approaches, there have been other approaches to combine logical and
probabilistic reasoning, such as probabilistic relational models \cite{koller99,getoor01}, various probabilistic
first-order logics such as \cite{nilsson86,bch90,bacchus96c,halpern90,halpern03-book,pasula01,poole93}, approaches that
assign a weight to first-order formulas \cite{paskin02,richardson05} and first-order MDPs \cite{boutilier01}. In all
these approaches the logic parts are not quite rich from the `knowledge representation' angle. To start with they use
classical logic, which is monotonic and hence has many drawbacks with respect to knowledge representation.  A difference between first-order MDPs and our approach is that actions, rewards and utilities are inherent part of the former; one may encode them in P-log though. In the next subsection we summarize specific differences between these approaches (and all the other approaches that we mentioned so far) and P-log.

\subsection{Summary}

In summary, our focus in P-log has many broad differences with most of the earlier formalisms that have tried to
integrate logical and probabilistic knowledge. We now list some of the main issues.

\begin{itemize}

\item To the best of our knowledge P-log is the only probabilistic logic programming language which differentiates
    between doing and observing, which is useful for reasoning about causal relations.

\item P-log allows a relatively wide variety of updates compared with other approaches we surveyed.

\item Only P-log allows logical reasoning to dynamically decide on the range of values that a random variable can
    take.

\item P-log is the only language surveyed which allows a programmer to write a program which represent the logical
    aspects of a problem and its possible worlds, and add causal probabilistic information to this program as it
    becomes relevant and available.

\item Our formalism allows the explicit specification of background knowledge and thus eliminates the difference
    between implicit and explicit background knowledge that is pointed out in \cite{wang04} while discussing the
    limitation of Bayesianism.

\item As our formalization of the Monty Hall example shows, P-log can deal with non-trivial conditioning and is able
    to encode the notion of protocols mentioned in Chapter 6 of \cite{halpern03-book}.

\end{itemize}

\section{Conclusion and Future Work}\label{summary-sec}
In this paper we presented a non-monotonic probabilistic logic programming language, P-log, suitable for representing
logical and probabilistic knowledge. P-log is based on logic programming under answer set semantics, and on Causal
Bayesian networks. We showed that it generalizes both languages.

\st P-log comes with a natural mechanism for belief updating --- the ability of the agent to change degrees of belief
defined by his current knowledge base. We showed that conditioning of classical probability is a special case of this
mechanism. In addition, P-log programs can be updated by actions, defaults and other logic programming rules, and by
some forms of probabilistic information. The non-monotonicity of P-log allows us to model situations when new
information forces the reasoner to change its collection of possible worlds, i.e. to move to a new probabilistic model
of the domain. (This happens for instance when the agent's knowledge is updated by observation of an event deemed to be
impossible under the current assumptions.)

\st The expressive power of P-log and its ability to combine various forms of reasoning was demonstrated on a number of
examples from the literature. The presentation of the examples is aimed to give a reader some feeling for the
methodology of representing knowledge in P-log. Finally the paper gives sufficiency conditions for coherency of P-log
programs and discusses the relationship of P-log with a number of other probabilistic logic programming formalisms.

\st We plan to expand our work in several directions. First we need to improve the efficiency of the P-log inference
engine. The current, naive, implementation relies on computation of all answer sets of the logical part of P-log
program. Even though it can efficiently reason with a surprising variety of interesting examples and puzzles, a more
efficient approach is needed to attack some other kinds of problems. We also would like to investigate the impact of replacing
Answer Set Prolog --- the current logical foundation of P-log --- by a more powerful logic programming language,
CR-prolog. The new extension of P-log will be able to deal with updates which are currently viewed as inconsistent. We
plan to use P-log as a tool for the investigation of various forms of reasoning, including reasoning with
counterfactuals and probabilistic abductive reasoning capable of discovering most probable explanations of unexpected
observations. Finally, we plan to explore how statistical relational learning (SRL) can be done with respect to P-log
and how P-log can be used to accommodate different kinds of uncertainties tackled by existing SRL approaches.

\st {\bf Acknowledgments}\\ We would like to thank Weijun Zhu and Cameron Buckner for their work in implementing a P-log
inference engine, for useful discussions and for helping correct errors in the original draft of this paper.

\section{Appendix I: Proofs of major theorems} \label{sufficiency-sec}
Our first goal in this section is to prove Theorem \ref{consisconds} from Section \ref{property-sec}. We'll begin by
proving a theorem which is more general but whose hypothesis is more difficult to verify. In order to state and prove
this general theorem, we need some terminology and lemmas.
\begin{definition} {\rm
Let $T$ be a tree in which every arc is labeled with a real number in [0,1]. We say $T$ is {\em unitary} if the labels
of the arcs leaving each node add up to 1. } \hfill $\Box$
\end{definition}

\begin{figure}[!htbp]
\centering
\includegraphics[width=9cm]{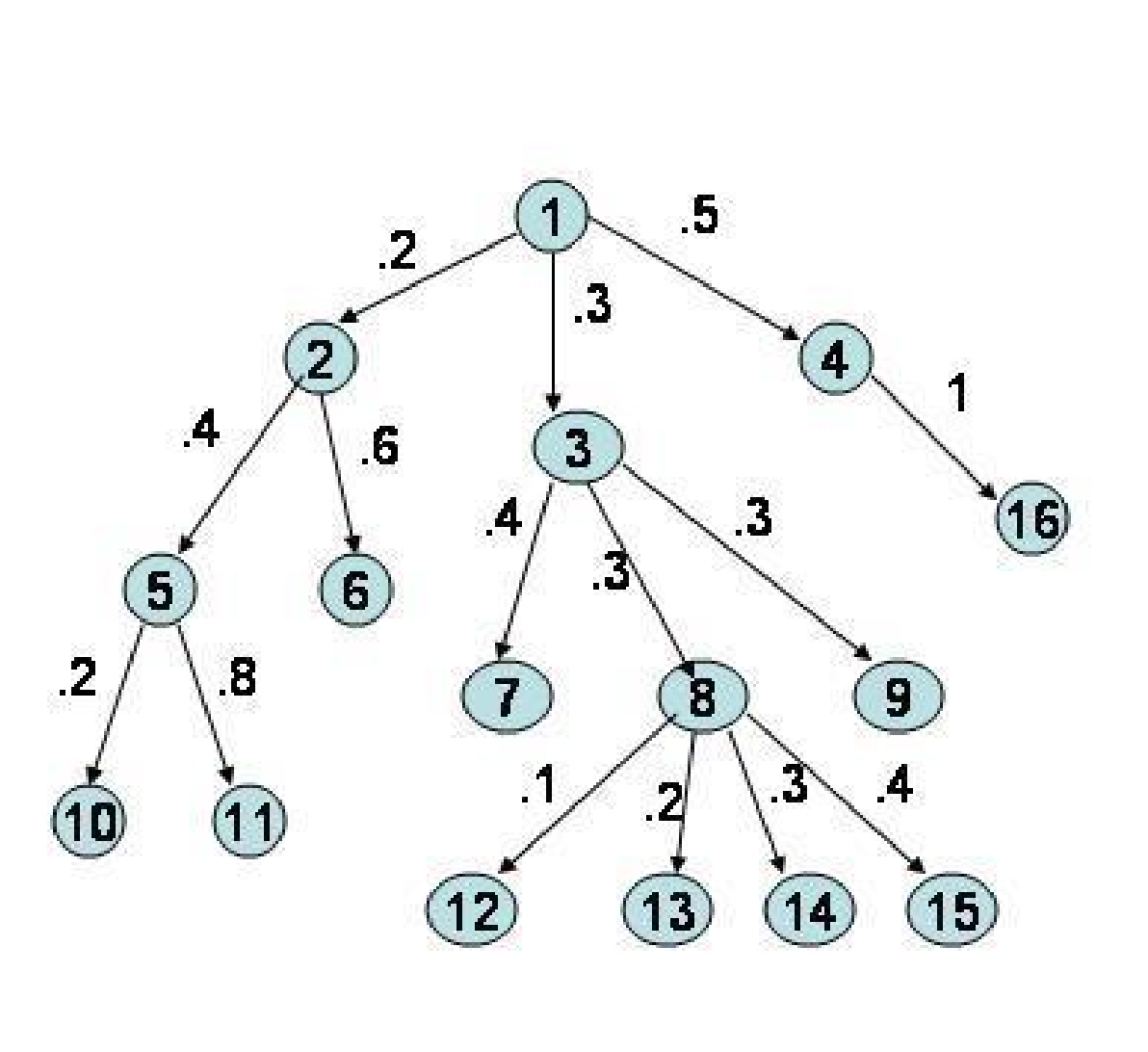}
\caption{Unitary tree T} \label{figure1}
\end{figure}

\noindent Figure~\ref{figure1} gives an example of a unitary tree.

\begin{definition} {\rm
Let $T$ be a tree with labeled nodes and $n$ be a node of $T$. By $p_T(n)$ we denote the set of labels of nodes lying on
the path from the root of $T$ to $n$, including the label of $n$ and the label of the root. } \hfill $\Box$
\end{definition}

\begin{example}
{\rm Consider the tree $T$ from Figure~\ref{figure1}. If $n$ is the node labeled (13), then $p_T(n) = \{1, 3, 8, 13\}$.}
\hfill $\Box$
\end{example}

\begin{definition}\label{path-value}{[Path Value]}\\
{\rm Let $T$ be a tree in which every arc is labeled with a number in [0,1]. The {\em path value} of a node $n$ of $T$,
denoted by $pv_T(n)$, is defined as the product of the labels of the arcs in the path to $n$ from the root. (Note that
the path value of the root of $T$ is $1$.) }\hfill $\Box$
\end{definition}
When the tree $T$ is obvious from the context we will simply right $pv(n)$.

\begin{example}
{\rm Consider the tree $T$ from Figure~\ref{figure1}. If $n$ is the node labeled (8), then $pv(n) = 0.3 \times 0.3 =
0.09$. } \hfill $\Box$
\end{example}

\begin{lemma}\label{lemma1}[Property of Unitary Trees]\\
{\rm Let $T$ be a unitary tree and $n$ be a node of $T$. Then the sum of the path values of all the leaf nodes descended
from $n$ (including $n$ if $n$ is a leaf) is the path value of $n$. } \hfill $\Box$
\end{lemma}

\noindent {\bf Proof:} We will prove that the conclusion holds for every unitary subtree of $T$ containing $n$, by
induction on the number of nodes descended from $n$. Since $T$ is a subtree of itself, the lemma will follow.

\st If $n$ has only one node descended from it (including $n$ itself if $n$ is a leaf) then $n$ is a leaf and then the
conclusion holds trivially.

\st Consider a subtree $S$ in which $n$ has $k$ nodes descended from it for some $k > 0$, and suppose the conclusion is
true for all subtrees where $n$ has less than $k$ descendents. Let $l$ be a leaf node descended from $n$ and let $p$ be
its parent. Let $S'$ be the subtree of $S$ consisting of all of $S$ except the children of $p$. By induction hypothesis,
the conclusion is true of $S'$. Let $c_1,\dots,c_n$ be the children of $p$. The sum of the path values of leaves
descended from $n$ in $S$ is the same as that in $S'$, except that $pv(p)$ is replaced by $pv(c_1)+ \dots +pv(c_n)$.
Hence, we will be done if we can show these are equal.

\st Let $l_1,...,l_n$ be the labels of the arcs leading to nodes $c_1,..,c_n$ respectively. Then $pv(c_1)+\dots+pv(c_n)
= l_1*pv(p) + \dots +l_n*pv(p)$ by definition of path value. Factoring out $pv(p)$ gives $pv(p)* (l_1+\dots+l_n)$. But
Since $S'$ is unitary, $l_1+\dots +l_n =1$ and so this is just $pv(p)$. \hfill $\Box$

\st Let $\Pi$ be a P-log program with signature $\Sigma$. Recall that $\tau (\Pi)$ denotes the translation of its
logical part into an Answer Set Prolog program. Similarly for a literal $l$ (in $\Sigma$) with respect to $\Pi$,
$\tau(l)$ will represent the corresponding literal in $\tau(\Pi)$. For example, $\tau(owner(d_1)= mike) = owner(d_1,
mike)$. For a set of literals $B$ (in $\Sigma$) with respect to $\Pi$, $\tau(B)$ will represent the set $\{ \tau(l) \ |
\ l \in B\}$.

\begin{definition} {\rm
A set $S$ of literals of $\Pi$ is $\Pi$-{\em compatible} with a literal $l$ of $\Sigma$ if there exists an answer set of
$\tau(\Pi)$ containing $\tau(S) \cup \{\tau(l)\}$. Otherwise $S$ is $\Pi$-{\em incompatible} with $l$. $S$ is $\Pi$-{\em
compatible} with a set $B$ of literals of $\Pi$ if there exists an answer set of $\tau(\Pi)$ containing $\tau(S) \cup
\tau(B)$; otherwise $S$ is $\Pi$-{\em incompatible} with $B$. } \hfill $\Box$
\end{definition}

\begin{definition} {\rm
A set $S$ of literals is said to $\Pi$-{\em guarantee} a literal $l$ if $S$ and $l$ are $\Pi$-compatible and every
answer set of $\tau(\Pi)$ containing $\tau(S)$ also contains $\tau(l)$; $S$ $\Pi$-{\em guarantees} a set $B$ of literals
if $S$ $\Pi$-guarantees every member of $B$. } \hfill $\Box$
\end{definition}

\begin{definition} {\rm
We say that $B$ is a {\em potential $\Pi$-cause} of $a(t)=y$ with respect to a rule $r$ if $\Pi$ contains rules of the
form
\begin{equation}\label{one}
[r]\ random(a(t) : \{X : p(X)\}) \leftarrow K.
\end{equation}
and
\begin{equation}
pr_r(a(t)=y \ |_c \ B) = v.
\end{equation}
} \hfill $\Box$
\end{definition}

\begin{definition}\label{def-ready}{[Ready to branch]}\\
{\rm Let $T$ be a tree whose nodes are labeled with literals and $r$ be a rule of $\Pi$ of the form
$$random(a(t): \{X : p(X)\}) \leftarrow K.$$
or
$$random(a(t)) \leftarrow K.$$
where $K$ can be empty. A node $n$ of $T$ is {\em ready to branch on $a(t)$ via $r$ relative to $\Pi$} if
\begin{enumerate}
\item $p_T(n)$ contains no literal of the form $a(t)=y$ for any $y$,

\item $p_T(n)$ $\Pi$-guarantees $K$,

\item for every rule of the form $pr_r(a(t)=y \ |_c \ B) = v$ in $\Pi$, either $p_T(n)$ $\Pi$-guarantees $B$ or is     $\Pi$-incompatible with $B$, and

\item if $r$ is of the first form then for every $y$ in the range of $a(t)$, $p_T(n)$ either $\Pi$-guarantees $p(y)$ or is $\Pi$-incompatible with $p(y)$ and moreover there is at least one
    $y$ such that $\ p_T(n)$ $\Pi$-guarantees $p(y)$.
\end{enumerate}
If $\Pi$ is obvious from context we may simply say that $n$ is ready to branch on $a(t)$ via $r$. \hfill $\Box$ }
\end{definition}

\begin{proposition}\label{v-defined}
{\rm Suppose $n$ is ready to branch on $a(t)$ via some rule $r$ of $\Pi$, and $a(t)=y$ is $\Pi$-compatible with
$p_T(n)$; and let $W_1$ and $W_2$ be possible worlds of $\Pi$ compatible with $p_T(n)$. Then $P(W_1, a(t)=y) = P(W_2,
a(t)=y)$. } \hfill $\Box$
\end{proposition}

\noindent {\bf Proof:} Suppose $n$ is ready to branch on $a(t)$ via some rule $r$ of $\Pi$, and $a(t)=y$ is
$\Pi$-compatible with $p_T(n)$; and let $W_1$ and $W_2$ be possible worlds of $\Pi$ compatible with $p_T(n)$.

Case 1: Suppose $a(t)=y$ has an assigned probability in $W_1$. Then there is a rule $pr_r(a(t)=y\ | \ B)=v$ of $\Pi$
such that $W_1$ satisfies $B$. Since $W_1$ also satisfies $p_T(n)$, $B$ is $\Pi$-compatible with $p_T(n)$. It follows
from the definition of ready-to-branch that $p_T(n)$ $\Pi$-guarantees $B$. Since $W_2$ satisfies $p_T(n)$ it must also
satisfy $B$ and so $P(W_2, a(t)=y) = v$.

Case 2: Suppose $a(t)=y$ does not have an assigned probability in $W_1$. Case 1 shows that the assigned probabilities
for values of $a(t)$ in $W_1$ and $W_2$ are precisely the same; so $a(t)=y$ has a default probability in both worlds. We
need only show that the possible values of $a(t)$ are the same in $W_1$ and $W_2$. Suppose then that for some $z$,
$a(t)=z$ is possible in $W_1$. Then $W_1$ satisfies $p(y)$. Hence since $W_1$ satisfies $p_T(n)$, we have that $p_T(n)$
is $\Pi$-compatible with $p(y)$. By definition of ready-to-branch, it follows that $p_T(n)$ $\Pi$-guarantees $p(y)$. Now
since $W_2$ satisfies $p_T(n)$ it must also satisfy $p(y)$ and hence $a(t)=y$ is possible in $W_2$. The other direction
is the same. $\Box$

Suppose $n$ is ready to branch on $a(t)$ via some rule $r$ of $\Pi$, and $a(t)=y$ is $\Pi$-compatible with $p_T(n)$, and
$W$ is a possible world of $\Pi$ compatible $p_T(n)$. We may refer to the $P(W,a(t)=y)$ as $v(n,a(t),y)$. Though the
latter notation does not mention $W$, it is well defined by proposition \ref{v-defined}.

\begin{figure}[!htbp]
\centering
\includegraphics[width=15cm]{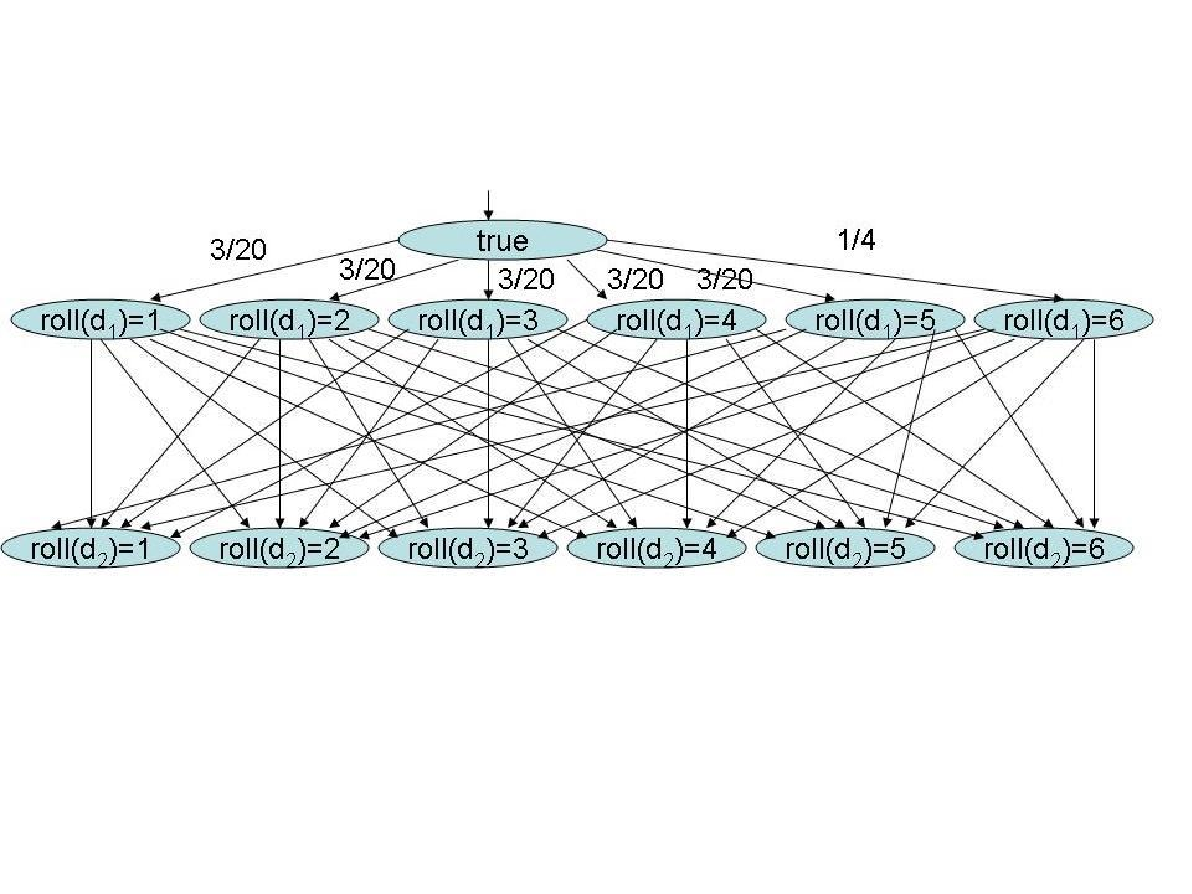}
\caption{$T_2$: The tree corresponding to the dice P-log program $\Pi_2$} \label{figure2}
\end{figure}

\begin{example}\label{ready-to}[Ready to branch]\\
{\rm Consider the following version of the dice example. Lets refer to it as $\Pi_2$

\st $dice = \{d_1,d_2\}.$\\ $score = \{1, 2, 3, 4, 5, 6\}.$\\ $person = \{mike,john\}.$\\ $roll : dice \rightarrow
score.$\\ $owner : dice \rightarrow person.$\\ $owner(d_1) = mike.$\\ $owner(d_2) = john.$\\ $even(D) \leftarrow roll(D)
= Y, Y \ mod \ 2 = 0.$\\ $\neg even(D) \leftarrow \no even(D).$\\ $[\ r(D) \ ]\ random(roll(D)).$\\ $pr(roll(D)=Y \ |_c
\ owner(D) = john) = 1/6.$\\ $pr(roll(D)=6 \ |_c \ owner(D) = mike) = 1/4$.\\ $pr(roll(D)=Y \ |_c \ Y \not= 6, owner(D)
= mike) = 3/20$.\\ where $D$ ranges over $\{d_1,d_2\}$.

\st Now consider a tree $T_2$ of Figure~\ref{figure2}. Let us refer to the root of this tree as $n_1$, the node
$roll(d_1)=1$ as $n_2$, and the node $roll(d_2)=2$ connected to $n_2$ as $n_3$. Then \st $p_{T_2}(n_1) = \{true\}$,
$p_{T_2}(n_2) = \{true,roll(d_1)=1\}$, and $p_{T_2}(n_3) = \{true, roll(d_1)=1, roll(d_2)=2\}$. The set $\{true\}$ of
literals $\Pi_2$-guarantees $\{owner(d_1) = mike, owner(d_2) = john\}$ and is $\Pi_2$-incompatible with $\{owner(d_1) =
john, owner(d_2) = mike\}$. Hence $n_1$ and the attribute $roll(d_1)$ satisfy condition 3 of definition \ref{def-ready}.
Similarly for $roll(d_2)$. Other conditions of the definition hold vacuously and therefore $n_1$ is ready to branch on
$roll(D)$ via $r(D)$ relative to $\Pi_2$ for $D \in \{d_1,d_2\}$. It is also easy to see that $n_2$ is ready to branch
on $roll(d_2)$ via $r(d_2)$, and that $n_3$ is not ready to branch on any attribute of $\Pi_2$. }\hfill $\Box$
\end{example}

\begin{definition}\label{expansion}{[Expanding a node]}\\
{\rm In case $n$ is ready to branch on $a(t)$ via some rule of $\Pi$, the $\Pi$-\emph{expansion} of $T$ at $n$ by $a(t)$
is a tree obtained from $T$ as follows: for each $y$ such that $p_T(n)$ is $\Pi$-compatible with $a(t)=y$, add an arc
leaving $n$, labeled with $v(n,a(t),y)$, and terminating in a node labeled with $a(t)=y$. We say that $n$ \emph{branches
on} $a(t)$. }\hfill $\Box$
\end{definition}

\begin{definition}\label{pi-exp}{[Expansions of a tree]}\\
{\rm A \emph{zero-step} $\Pi$-expansion of $T$ is $T$. A \emph{one-step $\Pi$-expansion of} $T$ is an expansion of $T$
at one of its leaves by some attribute term $a(t)$. For $n > 1$, an \emph{n-step $\Pi$-expansion} of $T$ is a one-step
$\Pi$-expansion of an $(n-1)$-step $\Pi$-expansion of $T$. A $\Pi$-\emph{expansion} of $T$ is an $n$-step
$\Pi$-expansion of $T$ for some non-negative integer $n$. } \hfill $\Box$
\end{definition}
For instance, the tree consisting of the top two layers of tree $T_2$ from Figure~\ref{figure2} is a $\Pi_2$-expansion
of one node tree $n_1$ by $roll(d_1)$.

\begin{definition} {\rm
A \emph{seed} is a tree with a single node labeled \emph{true}. }\hfill $\Box$ \end{definition}

\begin{definition}\label{tablo}{[Tableau]}\\
{\rm A \emph{tableau} of $\Pi$ is a $\Pi$-expansion of a seed which is maximal with respect to the subtree
relation.}\hfill $\Box$
\end{definition}
For instance, a tree $T_2$ of Figure~\ref{figure2} is a tableau of $\Pi_2$.
\begin{definition}\label{rep1}{[Node Representing a Possible World]}\\
{\rm Suppose $T$ is a tableau of $\Pi$. A possible world $W$ of $\Pi$ is \emph{represented} by a leaf node $n$ of $T$ if $W$ is the set of literals $\Pi$-guaranteed by $p_T(n)$. } \hfill $\Box$
\end{definition}

\noindent
For instance, a node $n_3$ of $T_2$
represents a possible world\\ $\{owner(d_1,mike), owner(d_2,john),roll(d_1,1),roll(d_2,2),\neg even(d_1), even(d_2)\}$.

\begin{definition}\label{tree-rep-prog}{[Tree Representing a Program]}\\
{\rm If every possible world of $\Pi$ is represented by exactly one leaf node of $T$, and every leaf node of $T$
represents exactly one possible world of $\Pi$, then we say $T$ \emph{represents} $\Pi$. } \hfill $\Box$
\end{definition}
It is easy to check that the tree $T_2$ represents $\Pi_2$.

\begin{definition}\label{prob-sound}{[Probabilistic Soundness]}\\
{\rm Suppose $\Pi$ is a P-log program and $T$ is a tableau representing $\Pi$, such that $R$ is a mapping from the possible worlds of $\Pi$ to the leaf nodes of $T$ which represent them. If for every possible world $W$ of $\Pi$ we have $$pv_T(R(W)) = \mu(W)$$ i.e. the path value in $T$ of $R(W)$ is equal to the probability of $W,$ then we say that the representation of $\Pi$ by $T$ is \emph{probabilistically sound}. }\hfill $\Box$
\end{definition}

\st The following theorem gives conditions sufficient for the coherency of P-log programs (Recall that we only consider
programs satisfying Conditions \ref{l1}, \ref{l2}, and \ref{l3} of Section \ref{assign-prob}). It will later be shown
that all unitary, ok programs satisfy the hypothesis of this theorem, establishing Theorem \ref{consisconds}.

\begin{theorem}\label{th1}[Coherency Condition]\\
{\rm Suppose $\Pi$ is a consistent P-log program such that $P_{\Pi}$ is defined. Let $\Pi^\prime$ be obtained from $\Pi$
by removing all observations and actions. If there exists a unitary tableau $T$ representing $\Pi^\prime$, and this
representation is probabilistically sound, then for every pair of rules
\begin{equation}\label{temp-r}
[r]\ random(a(t) : \{Y:p(Y)\}) \leftarrow K.
\end{equation}
and
\begin{equation}
pr_r(a(t)=y \ |_c \ B) = v.
\end{equation}
of $\Pi^\prime$ such that $P_{\Pi^\prime}(B \cup K) > 0$ we have
$$P_{\Pi^{\prime} \cup obs(B) \cup obs(K)}(a(t)=y)= v$$
Hence $\Pi$ is coherent.
} \hfill $\Box$
\end{theorem}

\noindent {\bf Proof:} For any set $S$ of literals, let $lgar(S)$ (pronounced ``L-gar'' for ``leaves guaranteeing'') be
the set of leaves $n$ of $T$ such that $p_T(n)$ $\Pi^\prime$-guarantees $S$.

\st Let $\mu$ denote the measure on possible worlds induced by $\Pi^\prime$. Let $\Omega$ be the set of possible worlds
of $\Pi^\prime \cup obs(B) \cup obs(K)$. Since $P_{\Pi^\prime}(B \cup K)>0$ we have
\begin{equation}\label{quotient}
P_{\Pi^{\prime} \cup obs(B) \cup obs(K)}(a(t)=y) =\frac{ \sum_{\{W \ :\ W \in \Omega \ \wedge \ a(t) = y \ \in \ W\} }
\mu(W)} { \sum_{\{W \ : \ W \in \Omega \}} \mu(W)}
\end{equation}
\noindent Now, let

$$ \alpha = \sum_{n \in lgar(B \cup K \cup \{a(t)=y)\}}pv(n) $$
$$ \beta = \sum_{n \in lgar(B \cup K)}pv(n)$$

\noindent Since $T$ is a probabilistically sound representation of $\Pi^\prime$, the right-hand side of (\ref{quotient})
can be written as $\alpha / \beta$. So we will be done if we can show that $\alpha / \beta = v$.

\st We first claim
\begin{equation}\label{anc}
\mbox{ Every } n \in lgar(B \cup K) \mbox{ has a unique ancestor } ga(n) \mbox{ which branches on } a(t) \mbox { via } r
\mbox{ (\ref{temp-r})}.
\end{equation}

\noindent If existence failed for some leaf $n$ then $n$ would be ready to branch on $a(t)$ which contradicts maximality of the tree. Uniqueness follows from Condition 1 of Definition \ref{def-ready}.

\st Next, we claim the following:
\begin{equation}\label{2}
\mbox{ For every } n \in lgar(B \cup K), \ p_T(ga(n)) \ \Pi\mbox{-guarantees } B \cup K.
\end{equation}

\noindent Let $n \in lgar(B \cup K)$. Since $ga(n)$ branches on $a(t)$, $ga(n)$ must be ready to $\Pi$-expand using
$a(t)$. So by (2) and (3) of the definition of ready-to-branch, $ga(n)$ either $\Pi^\prime$-guarantees $B$ or is
$\Pi^\prime$-incompatible with $B$. But $p_T(ga(n)) \subset p_T(n)$, and $p_T(n)$ $\Pi^\prime$-guarantees $B$, so
$p_T(ga(n))$ cannot be $\Pi^\prime$-incompatible with $B$. Hence $p_T(ga(n))$ $\Pi^\prime$-guarantees $B$. It is also
easy to see that $p_T(ga(n))$ $\Pi^\prime$-guarantees $K$.

\st From (\ref{2}), it follows easily that
\begin{equation}\label{3}
\mbox{ If } n \in lgar(B \cup K), \mbox{ every leaf descended from of } ga(n) \mbox{ belongs to } lgar(B \cup K).
\end{equation}

\noindent Let
$$ A =\{ga(n): n\in lgar(B \cup K)\}$$
In light of (\ref{anc}) and (\ref{3}), we have
\begin{equation}\label{4}
lgar(B \cup K) \mbox{ is precisely the set of leaves descended from nodes in } A.
\end{equation}

\noindent Therefore,
$$\beta= \sum_{n \mbox{ is a leaf descended from some } a \in A}
pv(n)$$ Moreover, by construction of $T$, no leaf may have more than one ancestor in $A$, and hence
$$ \beta= \sum_{a \in A} \ \ \sum_{n \mbox{ is a leaf descended
from } a} pv(n)$$ Now, by Lemma \ref{lemma1} on unitary trees, since $T$ is unitary,
$$\beta = \sum_{a \in A} pv(a)$$
This way of writing $\beta$ will help us complete the proof. Now for $\alpha$.

Recall the definition of $\alpha$:
$$\alpha = \sum_{n \in lgar(B \cup K \cup \{a(t)=y\})}pv(n) $$
Denote the index set of this sum by $lgar(B,K,y)$. Let
$$A_y = \{n : parent(n) \in A, \mbox{ the label of } n \mbox{
is } a(t)=y\}$$ Since $lgar(B,K,y)$ is a subset of $lgar(B) \cup K$, (\ref{4}) implies that $lgar(B,K,y)$ is precisely
the set of nodes descended from nodes in $A_y$. Hence
$$\alpha = \sum_{n^\prime \mbox{ is a leaf descended from some } n
\in A_y} pv(n^\prime) $$ Again, no leaf may descend from more than one node of $A_y$, and so by the lemma on unitary
trees,
\begin{equation} \label{5}
\alpha = \sum_{n \in A_y}\ \ \sum_{n^\prime \mbox{ is a leaf descended from } n} pv(n^\prime) = \sum_{n \in A_y} pv(n)
\end{equation}

\noindent Finally, we claim that every node $n$ in $A$ has a unique child in $A_y$, which we will label $ychild(n)$. The
existence and uniqueness follow from (\ref{2}), along with Condition \ref{l3} of Section \ref{assign-prob}, and the fact
that every node in $A$ branches on $a(t)$ via [$r$]. Thus from (\ref{5}) we obtain
$$\alpha = \sum_{n \in A} pv(ychild(n))$$

\noindent Note that if $n \in A,$ the arc from $n$ to $ychild(n)$ is labeled with $v$. Now we have:
$$ P_{\Pi^\prime \cup obs(B) \cup obs(K)}(a(t)=y) $$
$$ = \alpha/\beta$$
$$ = \sum_{n \in A} pv(ychild(n)) / \sum_{n \in A} pv(n) $$
$$ = \sum_{n \in A} pv(n)* v / \sum_{n \in A} pv(n) $$
$$ = v.$$
$\Box$

%%%%%%

\begin{proposition}\label{pp1}[Tableau for causally ordered programs]\\
Suppose $\Pi$ is a causally ordered P-log program; then there exists a tableau $T$ of $\Pi$ which represents $\Pi$. \hfill $\Box$
\end{proposition}

\st {\bf Proof:}\\ Let $|\ |$ be a causal order of $\Pi$, $a_1(t_1),\dots,a_m(t_m)$ be the ordering of its terms induced
by $|\ |$, and $\Pi_1,\dots,\Pi_{m+1}$ be the $|\ |$-induced structure of $\Pi$.

\st Consider a sequence $T_0,\dots,T_{m}$ of trees where $T_0$ is a tree with one node, $n_0$, labeled by {\em true},
and $T_i$ is obtained from $T_{i-1}$ by expanding every leaf of $T_{i-1}$ which is ready to branch on $a_i(t_i)$ via any
rule relative to $\Pi_{i}$ by this term. Let $T = T_{m}$. We will show that $T_m$ is a tableau of $\Pi$ which represents
$\Pi$.

\st Our proof will unfold as a sequence of lemmas:

\begin{lemma}\label{ll1}
{\rm For every $k \geq 0$ and every leaf node $n$ of $T_{k}$ program $\Pi_{k+1}$ has a unique possible world $W$
containing $p_{T_{k}}(n)$. } \hfill $\Box$
\end{lemma}

\st {\bf Proof:}\\ We use induction on $k$. The case where $k=0$ follows from Condition (1) of Definition \ref{c-order}
of causally ordered program. Assume that the lemma holds for $i = k-1$ and consider a leaf node $n$ of $T_k$. By
construction of $T$, there exists a leaf node $m$ of $T_{k-1}$ which is either the parent of $n$ or equal to $n$. By
inductive hypothesis there is a unique possible world $V$ of $\Pi_{k}$ containing $p_{T_{k-1}}(m)$.

\st (i) First we will show that every possible world $W$ of $\Pi_{k+1}$ containing $p_{T_{k-1}}(m)$ also contains $V$.
By the splitting set theorem \cite{lif94a}, set $V^\prime=W |_{L_k}$ is a possible world of $\Pi_k$. Obviously,
$p_{T_{k-1}}(m) \subseteq V^\prime$. By inductive hypothesis, $V^\prime = V$, and hence $V \subseteq W$.

\st Now let us consider two cases.

\st (ii) $a_k(\overline{t}_k)$ is not active in $V$ with respect to $\Pi_{k+1}$. In this case for every random selection
rule of $\Pi_{k+1}$ either Condition (2) or Condition (4) of definition \ref{def-ready} is not satisfied and hence there
is no rule $r$ such that $m$ is ready to branch on $a_k(\overline{t}_k)$ via $r$ relative to $\Pi_{k+1}$. From
construction of $T_k$ we have that $m=n$. By (3) of the definition of causally ordered, the program $V \cup \Pi_{k+1}$
has exactly one possible world, $W$. Since $L_k$ is a splitting set \cite{lif94a} of $\Pi_{k+1}$ we can use splitting
set theorem to conclude that $W$ is a possible world of $\Pi_{k+1}$. Obviously, $W$ contains $V$ and hence
$p_{T_{k-1}}(m)$. Since $n=m$ this implies that $W$ contains $p_{T_{k}}(n)$.

\st Uniqueness follows immediately from (i) and Condition (3) of Definition \ref{c-order}.

\st (iii) A term $a_k(\overline{t}_k)$ is active in $V$. This means that there is some random selection rule $r$
$$[r]\ random(a_k(\overline{t}_k) : \{Y:p(Y)\}) \leftarrow K.$$
such that $V$ satisfies $K$ and there is $y_0$ such that $p(y_0) \in V$. (If $r$ does not contain $p$ the latter
condition can be simply omitted). Recall that in this case $a_k(\overline{t}_k)=y_0$ is possible in $V$ with respect to
$\Pi_{k+1}$.

\st We will show that $m$ is ready to branch on $a_k(\overline{t}_k)$ via rule $r$ relative to $\Pi_{k+1}$.

\st Condition (1) of the definition of``ready to branch'' (Definition \ref{def-ready}) follows immediately from
construction of $T_{k-1}$.

\st To prove Condition (2) we need to show that $p_{T_{k-1}}(m)$ $\Pi_{k+1}$-guarantees $K$. To see that
$p_{T_{k-1}}(m)$ and $K$ are $\Pi_{k+1}$-compatible notice that, from Condition (2) of Definition \ref{c-order} and the
fact that $p(y_0) \in V$ we have that $V \cup \Pi_{k+1}$ has a possible world, say, $W_0$. Obviously it satisfies both,
$K$ and $p_{T_{k-1}}(m)$. Now consider a possible world $W$ of $\Pi_{k+1}$ which contains $p_{T_{k-1}}(m)$. By (i) we
have that $V \subseteq W$. Since $V$ satisfies $K$ so does $W$. Condition (2) of the definition of ready to branch is
satisfied.

\st To prove condition (3) consider $pr_r(a_k(\overline{t}_k)=y \ |_c\ B)=v$ from $\Pi_{k+1}$ such that $B$ is
$\Pi_{k+1}$-compatible with $p_{T_{k-1}}(m)$. $\Pi_k$-compatibility implies that there is a possible world $W_0$ of
$\Pi_{k+1}$ which contains both, $p_{T_{k-1}}(m)$ and $B$. By (i) we have that $V \subseteq W_0$ and hence $V$ satisfies
$B$. Since every possible world $W$ of $\Pi_{k+1}$ containing $p_{T_{k-1}}(m)$ also contains $V$ we have that $W$
satisfies $B$ which proves condition (3) of the definition.

\st To prove Condition (4) we consider $y_0$ such that $p(y_0) \in V$ (The existence of such $y_0$ is proven at the
beginning of (iii)). We show that $p_{T_{k-1}}(m)$ $\Pi_{k+1}$-guarantees $p(y_0)$. Since $a_k(\overline{t}_k)=y_0$ is
possible in $V$ with respect to $\Pi_{k+1}$ Condition (2) of Definition \ref{c-order} guarantees that $\Pi_{k+1}$ has
possible world, say $W,$ containing $V$. By construction, $p(y_0) \in V$ and hence $p(y_0)$ and $p_{T_{k-1}}(m)$ are
$\Pi_{k+1}$ compatible. From (i) we have that $p_{T_{k-1}}(m)$ $\Pi_{k+1}$-guarantees $p(y_0)$. Similar argument shows
that if $p_{T_{k-1}}(m)$ is $\Pi_{k+1}$-compatible with $p(y)$ then $p(y)$ is also $\Pi_{k+1}$-guaranteed by
$p_{T_{k-1}}(m)$.

\st We can now conclude that $m$ is ready to branch on $a_k(\overline{t}_k)$ via rule $r$ relative to $\Pi_{k+1}$. This
implies that a leaf node $n$ of $T_k$ is obtained from $m$ by expanding it by an atom $a_k(\overline{t}_k)=y$.

\st By Condition (2) of Definition \ref{c-order}, program $V \cup \Pi_{k+1} \cup obs(a_k(\overline{t}_k)=y)$ has exactly
one possible world, $W$. Since $L_k$ is a splitting set of $\Pi_{k+1}$ we have that $W$ is a possible world of
$\Pi_{k+1}$. Clearly $W$ contains $p_{T_{k}}(n)$. Uniqueness follows immediately from (i) and Condition (2) of
Definition \ref{c-order}.

\begin{lemma}\label{ll2}
{\rm For all $k \geq 0$, every possible world of $\Pi_{k+1}$ contains $p_{T_{k}}(n)$ for some unique leaf node $n$ of
$T_{k}$. } \hfill $\Box$
\end{lemma}

\st {\bf Proof:}\\ We use induction on $k$. The case where $k=0$ is immediate. Assume that the lemma holds for $i=k-1$,
and consider a possible world $W$ of $\Pi_{k+1}$. By the splitting set theorem $W$ is a possible world of $V \cup
\Pi_{k+1}$ where $V$ is a possible world of $\Pi_k$. By the inductive hypothesis there is a unique leaf node $m$ of
$T_{k-1}$ such that $V$ contains $p_{T_{k-1}}(m)$. Consider two cases.

\st (a) The attribute term $a_k(\overline{t}_k)$ is not active in $V$ and hence $m$ is not ready to branch on
$a_k(\overline{t}_k)$. This means that $m$ is a leaf of $T_k$ and $p_{T_{k-1}}(m) = p_{T_{k}}(m)$. Let $n = m$. Since $V
\subseteq W$ we have that $p_{T_{k}}(n) \subseteq W$. To show uniqueness suppose $n^\prime$ is a leaf node of $T_k$ such
that $p_{T_k}(n^\prime)\subseteq W$, and $n^\prime$ is not equal to $n$. By construction of $T_k$ there is some $j$ and
some $y_1 \not= y_2$ such that $a_j(\overline{t}_j) = y_1 \in p_{T_k}(n^\prime)$ and $a_j(\overline{t}_j) = y_2 \in
p_{T_k}(n)$. Since W is consistent and $a_j$ is a function we can conclude $n$ cannot differ from $n^\prime$.

\st (b) If $a_k(\overline{t}_k)$ is active in $V$ then there is a possible outcome $y$ of $a_k(\overline{t}_k)$ in $V$
with respect $\Pi_{k+1}$ via some random selection rule $r$ such that $a_k(\overline{t}_k) = y \in W$. By inductive
hypothesis $V$ contains $p_{T_{k-1}}(m)$ for some leaf $m$ of $T_{k-1}$. Repeating the argument from part (iii) of the
proof of Lemma \ref{ll1} we can show that $m$ is ready to branch on $a_k(\overline{t}_k)$ via $r$ relative to
$\Pi_{k+1}$. Since $a_k(\overline{t}_k) = y$ is possible in $V$ there is a son $n$ of $m$ in $T_k$ labeled by
$a_k(\overline{t}_k) = y$. It is easy to see that $W$ contains $p_{T_k}(n)$. The proof of uniqueness is similar to that
used in (a).

\begin{lemma}\label{ll3}
{\rm For every leaf node $n$ of $T_{i-1}$, every set $B$ of extended literals of $L_{i-1}$, and every $i \leq j \leq
m+1$ we have $p_{T_{i-1}}(n)$ is $\Pi_i$-compatible with $B$ iff $p_{T_{i-1}}(n)$ is $\Pi_j$-compatible with $B$. } \hfill $\Box$
\end{lemma}

{\bf Proof:}\\ $\rightarrow$\\ Suppose that $p_{T_{i-1}}(n)$ is $\Pi_i$-compatible with $B$. This means that there is a
possible world $V$ of $\Pi_i$ which satisfies $p_{T_{i-1}}(n)$ and $B$. To construct a possible world of $\Pi_j$ with
the same property consider a leaf node $m$ of $T_{j-1}$ belonging to a path containing node $n$ of $T_{i-1}$. By Lemma
\ref{ll1} $\Pi_j$ has a unique possible world $W$ containing $p_{T_{j-1}}(m)$. $L_i$ is a splitting set of $\Pi_j$ and
hence, by the splitting set theorem, we have that $W = V^\prime \cup U$ where $V^\prime$ is a possible world of $\Pi_i$
and $U \cap L_{i} = \emptyset$. This implies that $V^\prime$ contains $p_{T_{i-1}}(n)$, and hence, by Lemma \ref{ll1}
$V^\prime = V$. Since $V$ satisfies $B$ and $U \cap L_{i} = \emptyset$ we have that $W$ also satisfies $B$ and hence
$p_{T_{i-1}}(n)$ is $\Pi_j$-compatible with $B$.

\st $\leftarrow$\\ Let $W$ be a possible world of $\Pi_j$ satisfying $p_{T_{i-1}}(n)$ and $B$. By the splitting set
theorem we have that $W = V \cup U$ where $V$ is a possible world of $\Pi_i$ and $U \cap L_{i} = \emptyset$. Since $B$
and $p_{T_{i-1}}(n)$ belong to the language of $L_i$ we have that $B$ and $p_{T_{i-1}}(n)$ are satisfied by $V$ and
hence $p_{T_{i-1}}(n)$ is $\Pi_i$-compatible with $B$.

\begin{lemma}\label{ll4}
{\rm For every leaf node $n$ of $T_{i-1}$, every set $B$ of extended literals of $L_{i-1}$, and every $i \leq j \leq
m+1$ we have $p_{T_{i-1}}(n)$ $\Pi_i$-guarantees $B$ iff $p_{T_{i-1}}(n)$ $\Pi_j$-guarantees $B$. } \hfill $\Box$
\end{lemma}

$\rightarrow$\\ Let us assume that $p_{T_{i-1}}(n)$ $\Pi_i$-guarantees $B$. This implies that $p_{T_{i-1}}(n)$ is
$\Pi_i$-compatible with $B$, and hence, by Lemma \ref{ll3} $p_{T_{i-1}}(n)$ is $\Pi_j$-compatible with $B$. Now let $W$
be a possible world of $\Pi_j$ satisfying $p_{T_{i-1}}(n)$. By the splitting set theorem $W = V \cup U$ where $V$ is a
possible world of $\Pi_i$ and $U \cap L_{i} = \emptyset$. This implies that $V$ satisfies $p_{T_{i-1}}(n)$. Since
$p_{T_{i-1}}(n)$ $\Pi_i$-guarantees $B$ we also have that $V$ satisfies $B$. Finally, since $U \cap L_{i} = \emptyset$
we can conclude that $W$ satisfies $B$.

\st $\leftarrow$\\ Suppose now that $p_{T_{i-1}}(n)$ $\Pi_j$-guarantees $B$. This implies that $p_{T_{i-1}}(n)$ is
$\Pi_i$-compatible with $B$. Now let $V$ be a possible world of $\Pi_i$ containing $p_{T_{i-1}}(n)$. To show that $V$
satisfies $B$ let us consider a leaf node $m$ of a path of $T_{j-1}$ containing $n$. By Lemma \ref{ll1} $\Pi_j$ has a
unique possible world $W$ containing $p_{T_{j-1}}(m)$. By construction, $W$ also contains $p_{T_{i-1}}(n)$ and hence
satisfies $B$. By the splitting set theorem $W = V^\prime \cup U$ where $V^\prime$ is a possible world of $\Pi_i$ and $U \cap L_{i} = \emptyset$. Since $B$ belongs to the language of $L_i$ it is satisfied by $V^\prime$. By Lemma \ref{ll1}
$V^\prime = V$. Thus V satisfies B and we conclude $p_{T_{i-1}}(n)$ $\Pi_i$-guarantees $B$.

\begin{lemma}\label{ll5}
{\rm For every $i \leq j \leq m+1$ and every leaf node $n$ of $T_{i-1}$, $n$ is ready to branch on term
$a_i(\overline{t}_i)$ relative to $\Pi_{i}$ iff $n$ is ready to branch on $a_i(\overline{t}_i)$ relative to $\Pi_{j}$.
} \hfill $\Box$ \end{lemma}

\noindent
{\bf Proof:} \\ $\rightarrow$\\ Condition (1) of Definition \ref{def-ready} follows immediately from construction of
$T$'s. To prove condition (2) consider a leaf node $n$ of $T_{i-1}$ which is ready to branch on $a_i(\overline{t}_i)$
relative to $\Pi_{i}$. This means that $\Pi_{i}$ contains a random selection rule $r$ whose body is $\Pi_i$-guaranteed
by $p_{T_{i-1}}(n)$. By definition of $L_i$, the extended literals from $K$ belong to the language $L_{i}$ and hence, by
Lemma \ref{ll4}, $p_{T_{i-1}}(n)$ $\Pi_j$-guarantees $K$.

\st Now consider a set $B$ of extended literals from condition (3) of Definition \ref{def-ready} and assume that
$p_{T_{i-1}}(n)$ is $\Pi_j$-compatible with $B$. To show that $p_{T_{i-1}}(n)$ $\Pi_j$-guarantees $B$ note that, by
Lemma \ref{ll3}, $p_{T_{i-1}}(n)$ is $\Pi_i$-compatible with $B$. Since $n$ is ready to branch on $a_i(\overline{t}_i)$
relative to $\Pi_{i}$ we have that $p_{T_{i-1}}(n)$ $\Pi_i$-guarantees $B$. By Lemma \ref{ll4} we have that
$p_{T_{i-1}}(n)$ $\Pi_j$-guarantees $B$ and hence Condition (3) of Definition \ref{def-ready} is satisfied. Condition
(4) is similar to check.\\

\st $\leftarrow$\\ As before Condition (1) is immediate. To prove Condition (2) consider a leaf node $n$ of $T_{i-1}$
which is ready to branch on $a_i(\overline{t}_i)$ relative to $\Pi_{j}$. This means that $p_{T_{i-1}}(n)$
$\Pi_j$-guarantees $K$ for some rule $r$ from $\Pi_j$. Since $\Pi_j$ is causally ordered we have that $r$ belongs to
$\Pi_i$. By Lemma \ref{ll4} $p_{T_{i-1}}(n)$ $\Pi_i$-guarantees $K$. Similar proof can be used to establish Conditions
(3) and (4).

\begin{lemma}\label{ll6}
{\rm $T=T_m$ is a tableau for $\Pi=\Pi_{m+1}$. } \hfill $\Box$
\end{lemma}

\noindent
{\bf Proof:}\\ Follows immediately from the construction of the $T$'s and $\Pi$'s, the definition of a tableau, and
Lemmas \ref{ll5} and \ref{ll3}. $\Box$

\begin{lemma}\label{ll7}
{\rm $T=T_{m}$ represents $\Pi = \Pi_{m+1}$. } \hfill $\Box$
\end{lemma}

\noindent
{\bf Proof:}\\ Let $W$ be a possible world of $\Pi$. By Lemma \ref{ll2} $W$ contains $p_{T}(n)$ for some unique leaf
node $n$ of $T$. By Lemma 2, $W$ is the set of literals $\Pi$-guaranteed by $p_{T}(n)$, and hence $W$ is represented by
$n$. Suppose now that $n^\prime$ is a node of $T$ representing $W$. Then $p_{T}(n^\prime)$ $\Pi$-guarantees $W$ which
implies that $W$ contains $p_{T_{m}}(n^\prime)$. By Lemma \ref{ll2} this means that $n = n^\prime$, and hence we proved
that every answer set of $\Pi$ is represented by exactly one leaf node of $T$.

\st Now let $n$ be a leaf node of $T$. By Lemma \ref{ll1} $\Pi$ has a unique possible world $W$ containing $p_T(n)$. It
is easy to see that $W$ is the set of literals represented by $n$. $\Box$

\st
\begin{lemma}\label{A}{\ }
{\rm Suppose $T$ is a tableau representing $\Pi$. If $n$ is a node of $T$ which is ready to branch on $a(t)$ via $r$, then all possible worlds of $\Pi$ compatible with $p_T(n)$ are probabilistically equivalent with respect to $r$. } \hfill $\Box$
\end{lemma}

\noindent {\bf Proof:}\\ This is immediate from Conditions (3) and (4) of the definition of ready-to-branch.

\st Notation: If $n$ is a node of $T$ which is ready to branch on $a(t)$ via $r$, the Lemma \ref{A} guarantees that
there is a unique scenario for $r$ containing all possible worlds compatible with $p_T(n)$. We will refer to this
scenario as \emph{the scenario determined by $n$}.

\st We are now ready to prove the main theorem.

\st {\bf Theorem \ref{consisconds}}\\ Every causally ordered, unitary program is coherent.

\st {\bf Proof:}

\st Suppose $\Pi$ is causally ordered and unitary. Proposition~\ref{pp1} tells us that $\Pi$ is represented by some
tableau $T$. By Theorem~\ref{th1} we need only show that $\Pi$ is unitary --- i.e., that for every node $n$ of $\Pi$,
the sum of the labels of the arcs leaving $n$ is 1. Let $n$ be a node and let $s$ be the scenario determined by $n$. $s$
satisfies (1) or (2) of the Definition~\ref{unitary-rule}. In case (1) is satisfied, the definition of $v(n, a(t),
y)$, along with the construction of the labels of arcs of $T$, guarantee that the sum of the labels of the arcs leaving
$n$ is 1. In case (2) is satisfied, the conclusion follows from the same considerations, along with the definition of
$PD(W, a(t)=y)$.

\st We now restate and prove Theorem \ref{th2}.

\st {\bf Theorem \ref{th2}}\\ Let $x_1, \dots, x_n$ be a nonempty vector of random variables, under a classical
probability $P$, taking finitely many values each. Let $R_i$ be the set of possible values of each $x_i$, and assume
$R_i$ is nonempty for each $i$. Then there exists a coherent P-log program $\Pi$ with random attributes $x_1, \dots, x_n$ such that for every vector $r_1,\dots,r_n$ from $R_1 \times .. \times R_n,$ we have
\begin{equation}\label{in-th2-1-b}
P(x_1=r_1,\dots,x_n = r_n) = P_{\Pi}(x_1=r_1,\dots, x_n = r_n)
\end{equation}
\hfill $\Box$

\noindent {\bf Proof}:

\comment{Use different letters for the number of var and an arbitrary node}

\st
For each $i$ let $pars(x_i) = \{x_1,\dots,x_{i-1}\}$. Let
$\Pi$ be formed as follows: For each $x_i$, $\Pi$ contains
$$x_i : R_i.$$
$$random(x_i).$$
Also, for each $x_i$, every possible value $y$ of $x_i$, and
every vector of possible values $y_p$ of $pars(x_i)$, let
$\Pi$ contain
$$pr(x_i = y \ |_c \ pars(i) = y_p) = v(i,y, y_p)$$
where $v(i,y, y_p) = P(x_i = y | pars(i) = y_p)$.

\st
Construct a tableau $T$ for $\Pi$
as follows: Beginning with the root which has depth 0, for every
node $n$ at depth $i$ and every possible value $y$ of $x_{i+1},$
add an arc leaving $n$, terminating in a node labeled
$x_{i+1}=y$;
label the arc with $P(x_{i+1} = y | p_{T}(n))$.

\st
We first claim that $T$ is unitary. This follows from the
construction of $T$ and basic probability theory, since the
labels of the arcs leaving any node $n$ at depth $i$ are the
respective conditional probabilities, given $p_T(n),$ of all
possible values of $x_{i+1}$.

\st
We now claim that $T$ represents $\Pi$.
Each answer set of $\tau(\Pi)$, the translation of $\Pi$ into
Answer Set Prolog, satisfies $x_1=r_1, \dots, x_n=r_n$ for exactly one
vector $r_1, \dots, r_n$ in $R_1\times \dots \times R_n$,
and every such vector is satisfied
in exactly one answer set. For the answer set
$S$ satisfying $x_1=r_1, \dots, x_n=r_n$, let $M(S)$ be
the leaf node $n$ of $T$ such that
$p_T(n) = \{x_1=r_1, \dots, x_n=r_n \}$.  $M(S)$
represents $S$ by Definition \ref{rep1}, since $\Pi$
has no non-random attributes.
Since $M$ is a one-to-one correspondence, $T$ represents $\Pi$.
($\ref{in-th2-1-b}$) holds because

\st
$P(x_1=r_1,...,x_n=r_n)$

\st
$= P(x_1=r_1) \times P(x_2=r_2 | x_1 = r_1) \times \dots \times P(x_n = r_n |
x_1=r_1,...,x_{n-1}=r_{n-1})$

\st
$= v(1,r_1,(\ )) \times \dots \times v(n,r_n,(r_1,\dots,r_{n-1}))$

\st
$= P_{\Pi}(x_1=r_1,\dots,x_n=r_n)$

\st
To complete the proof we will use Theorem \ref{th1} to show that
$\Pi$ is coherent. $\Pi$ trivially satisfies the Unique
selection rule. The Unique probability assignment rule is
satisfied because $pars(x_i)$ cannot take on two different
values $y_p^1$ and $y_p^2$ in the same answer set. $\Pi$
is consistent because by assumption $1 \leq n $ and $R_1$ is
nonempty. For the same reason, $P_{\Pi}$ is defined.  $\Pi$
contains no {\em do}  or {\em obs} literals; so we can apply
Theorem 3 directly to $\Pi$ without removing anything. We
have shown that $T$ is unitary and represents $\Pi$. The
representation is probabilistically sound by the construction
of $T$. These are all the things
that need to be checked to apply Theorem \ref{th1} to show that
$\Pi$ is coherent. \hfill $\Box$

\COMMENT

\st For each $i$ let $pars(x_i)$ be a fixed vector of Markov parents of $x_i$ with respect to the ordering $x_1, \dots,
x_n$. Let $\Pi$ be formed as follows: For each $x_i$, $\Pi$ contains
$$x_i : R_i.$$
$$random(x_i).$$
Also, for each $x_i$, every possible value $y$ of $x_i$, and every vector of possible values $y_p$ of $pars(x_i)$, let
$\Pi$ contain
$$pr(x_i = y \ |_c \ pars(i) = y_p) = v(i,y, y_p).$$
where $v(i,y, y_p) = P(x_i = y | pars(i) = y_p)$.

\st We claim $\Pi$ is consistent. Construct a tableau $T$ for $\Pi$ as follows: beginning with the root which has depth
0, for every node $n$ at depth $i$ and every possible value $y$ of $x_{i+1},$ add an arc leaving $n$, terminating in a
node labeled $x_{i+1}=y$. Label the arc with $P(x_{i+1} = y | pv(n)),$ where $pv(n)$ is the path value of $n$ as in
Definition \ref{path-value}.

\st We first claim that $T$ is unitary. This follows from the construction of $T$ and basic probability theory, since
the labels of the arcs leaving any node $n$ at depth $i$ are the respective conditional probabilities, given $p_T(n),$
of all possible values of $x_{i+1}$.

\st We now claim that $T$ represents $\Pi$. Each answer set of $\tau(\Pi)$, the translation of $\Pi$ into Answer Set
Prolog, satisfies $x_1=r_1, \dots, x_n=r_n$ for exactly one vector $r_1, \dots, r_n$ in $R_1\times \dots \times R_n$ ,
and every such vector is satisfied in exactly one answer set. For the answer set $S$ satisfying $x_1=r_1, \dots,
x_n=r_n$, let $M(S)$ be the leaf node $n$ of $T$ such that $p_T(n) = \{x_1=r_1, \dots, x_n=r_n \}$.  $M(S)$ represents
$S$ by Definition \ref{rep1}, since $\Pi$ has no non-random attributes. Since $M$ is a one-to-one correspondence, $T$
represents $\Pi$. Also, since (\ref{in-th2-2}) is also equal to $P(x_1=r_1, \dots, x_n=r_n)$, this proves that
($\ref{in-th2-1-b}$) holds.

\st To complete the proof we will use Theorem \ref{th1} to show that $\Pi$ is consistent. $\Pi$ trivially satisfies the
Unique selection rule. The Unique probability assignment rule is satisfied because $pars(x_i)$ cannot take on two
different values $y_p^1$ and $y_p^2$ in the same answer set. $\Pi$ satisfies Condition 1 of consistency because by
assumption $1 \leq n $ and $R_1$ is nonempty. $\Pi$ contains no {\em do} atoms, and we have shown that $T$ is unitary.
Hence $\Pi$ is consistent by Theorem \ref{th1}. $\Box$

\ENDCOMMENT

\st Finally we give proof of Proposition~\ref{eq-cond-1}.
\begin{proposition} \label{eq-cond-1}
{\rm Let $T$ be a P-log program over signature $\Sigma$ not containing $pr$-atoms, and $B$ a collection of
$\Sigma$-literals. If
\begin{enumerate}
\item all random selection rules of $T$ are of the form \st $random(a(\overline{t}))$,

\item $T \cup obs(B)$ is
    coherent, and

\item for every term $a(\overline{t})$ appearing in literals from $B$ program $T$ contains a random selection rule $random(a(\overline{t}))$,
\end{enumerate}
then for every formula $A$
$$ P_{T \cup B}(A) = P_{T \cup obs(B)}(A)$$
} \hfill $\Box$
\end{proposition}

\st {\bf Proof}: \\ We will need some terminology. Answer Set Prolog programs $\Pi_1$ and $\Pi_2$ are called {\em
equivalent} (symbolically, $\Pi_1 \equiv \Pi_2$) if they have the same answer sets; $\Pi_1$ and $\Pi_2$ are called
strongly equivalent (symbolically $\Pi_1 \equiv_s \Pi_2$) if for every program $\Pi$ we have that $\Pi_1 \cup \Pi \equiv
\Pi_2 \cup \Pi$. To simplify the presentation let us consider a program $T^\prime = T \cup B \cup obs(B)$. Using the
splitting set theorem it is easy to show that $W$ is a possible world of $T \cup B$ iff $W \cup obs(B)$ is a possible
world of $T^\prime$. To show

\st $(1) \ P_{T \cup B}(A) = P_{T \cup obs(B)}(A).$

\st we notice that, since $T^\prime$, $T \cup B$ and $T \cup obs(B)$ have the same probabilistic parts and the same
collections of $do$-atoms to prove (1) it suffices to show that

\st $(2)\ W \mbox{ is a possible world of } T^\prime \mbox{ iff } W \mbox{ is a possible world of } T \cup obs(B)$.

\st Let $P_{B} = \tau(T^\prime)$ and $P_{obs(B)} = \tau(T \cup obs(B))$. By definition of possible worlds (2) holds iff

\st $(3)\ P_{B} \equiv P_{obs(B)}$

\st To prove (3) let us first notice that the set of literals $S$ formed by relations $do$, $obs$, and $intervene$ form
a splitting set of programs $P_B$ and $P_{obs(B)}$. Both programs include the same collection of rules whose heads
belong to this splitting set. Let $X$ be the answer set of this collection and let $Q_B$ and $Q_{obs(B)}$ be partial
evaluations of $P_B$ and $P_{obs(B)}$ with respect to $X$ and $S$. From the splitting set theorem we have that (3) holds
iff

\st $(4) \ Q_B \equiv Q_{obs(B)}$.

\st To prove (4) we will show that for every literal $l \in B$ there are sets $U_1(l)$ and $U_2(l)$ such that for some
$Q$

\st $(5) \ Q_{obs(B)} = Q \cup \{r: r \in U_1(l) \mbox{ for some } l \in B\}$,

\st $(6) \ Q_{B} = Q \cup \{r: r \in U_2(l) \mbox{ for some } l \in B\}$,

\st $(7) \ U_1(l) \equiv_s U_2(l)$

\st which will imply (4).

\st Let literal $l \in B$ be formed by an attribute $a(\overline{t})$. Consider two cases:

\st Case 1: $intervene(a(\overline{t})) \not\in X$.

\st Let $U_1(l)$ consist of the rules

\st $(a)\ \ \neg a(\overline{t},Y_1) \leftarrow a(\overline{t},Y_2), Y_1\not=Y_2$.

\st $(b) \ \ a(\overline{t},y_1) \mbox{ or }\dots \mbox{ or } a(\overline{t},y_k)$.

\st $(c)\ \ \leftarrow \no l$.

Let $U_2(l) = U_1(l) \cup B$.

It is easy to see that due to the restrictions on random selection rules of $T$ from the proposition $U_1(l)$ belongs to
the partial evaluation of $\tau(T)$ with respect to $X$ and $S$. Hence $U_1(l) \subset Q_{obs(B)}$. Similarly $U_2(l)
\subset Q_{B}$, and hence $U_1(l)$ and $U_2(l)$ satisfy conditions (5) and (6) above. To show that they satisfy
condition (7) we use the method developed in \cite{LiPeVa}. First we reinterpret the connectives of statements of
$U_1(l)$ and $U_2(l)$. In the new interpretation $\neg$ will be a strong negation of Nelson \cite{n49}; $\no$,
$\leftarrow$, {\bf or} will be interpreted as intuitionistic negation, implication, and disjunction respectively; $,$
will stand for $\wedge$. A program $P$ with connectives reinterpreted in this way will be referred to as \emph{NL
counterpart} of $P$. Note that the NL counterpart of $\leftarrow \no l$ is $\no \no l$. Next we will show that, under
this interpretation, $U_1(l)$ and $U_2(l)$ are equivalent in Nelson's intuitionistic logic (NL). Symbolically,

\st $(8) \ \ U_1(l) \equiv_{NL} U_2(l)$.

\st (Roughly speaking this means that $U_1(l)$ can be derived from $U_2(l)$ and $U_2(l)$ from $U_1(l)$ without the use
of the law of exclusive middle.) As shown in \cite{LiPeVa} two programs whose NL counterparts are equivalent in NL are
strongly equivalent, which implies (7).

\st To show (8) it suffices to show that

\st $(9) \ \ U_1(l) \vdash_{NL} l$.

\st If $l$ is of the form $a(\overline{t},y_i)$ then let us assume $a(\overline{t},y_j)$ where $j \not= i$. This,
together with the NL counterpart of rule (a) derives $\neg a(\overline{t},y_i)$. Since in NL $\neg A \vdash \no A$ this
derives $\no a(\overline{t},y_i)$, which contradicts the NL counterpart $\no \no a(\overline{t},y_i)$ of (c). The only
disjunct left in (b) is $ a(\overline{t},y_i)$.

\st If $l$ is of the form $\neg a(\overline{t},y_i)$ then (9) follows from (a) and (b).

\st Case 2: $intervene(a(\overline{t})) \in X$

\st This implies that there is some $y_i$ such that $do(a(\overline{t})=y_i) \in T$.

\st If $l$ is of the form $a(\overline{t}) = y$ then since $T \cup obs(B)$ is coherent, we have that $y = y_i$, and thus $Q_B$ and $Q_{obs(B)}$ are identical.

\st If $l$ is of the form $a(\overline{t}) \not= y$ then, since $T \cup obs(B)$ is coherent, we have that $y \not=
y_i$.

\st Let $U_1(l)$ consist of rules:

\st $\neg a(\overline{t},y) \leftarrow a(\overline{t},y_i)$.

\st $a(\overline{t},y_i)$.

\st Let $U_2(l) = U_1(l) \cup \neg a(\overline{t},y)$.

\st Obviously $U_1(l) \subset Q_{obs(B)}$, $U_2(l) \subset Q_B$ and $U_1(l)$ entails $U_2(l)$ in NL. Hence we have (7)
and therefore (4).

\st This concludes the proof.

\section{Appendix II: Causal Bayesian Networks} \label{cbns}

This section gives a definition of causal Bayesian networks, closely following the definition of Judea Pearl and
equivalent to the definition given in \cite{pearl99b}. Pearl's definition reflects the intuition that causal influence
can be elucidated, and distinguished from mere correlation, by {\em controlled experiments}, in which one or more
variables are deliberately manipulated while other variables are left to their normal behavior. For example, there is a
strong correlation between smoking and lung cancer, but it could be hypothesized that this correlation is due to a
genetic condition which tends to cause both lung cancer and a susceptibility to cigarette addiction. Evidence of a
causal link could be obtained, for example, by a controlled experiment in which one randomly selected group of people
would be forced to smoke, another group selected in the same way would be forced not to, and cancer rates measured among both groups (not that we recommend such an experiment). The definitions below characterize causal links among a
collection $V$ of variables in terms of the numerical properties of probability measures on $V$ in the presence of
interventions. Pearl gives the name ``interventional distribution'' to a function from interventions to probability
measures. Given an interventional distributipn $P^*$, the goal is to describe conditions under which a set of causal
links, represented by a DAG, agrees with the probabilistic and causal information contained in $P^*$. In this case the
DAG will be called a causal Bayesian network compatible with $P^*$.

\st We begin with some preliminary definitions. Let $V$ be a finite set of variables, where each $v$ in $V$ takes values from some finite set $D(v)$. By an {\em assignment} on $V$, we mean a function which maps each $v$ in $V$ to some member of $D(v)$. We will let $A(V)$ denote the set of all assignments on $V$. Assignments on $V$ may also be called
\emph{possible worlds} of $V$.

\st A {\em partial assignment} on $V$ is an assignment on a subset of $V$. We will say two partial assignments are {\em
consistent} if they do not assign different values to the same variable. Partial assignments can also be called {\em
interventions}. Let $Interv(V)$ be the set of all interventions on $V$, and let $\{\ \}$ denote the empty intervention,
that is, the unique assignment on the empty set of variables.

\st By a {\em probability measure} on $V$ we mean a function $P$ which maps every set of possible worlds of $V$ to a
real number in $[0,1]$ and satisfies the Kolmogorov Axioms.

\st When $P$ is a probability measure on $V$, the arguments of $P$ are sets of possible worlds of $V$. However, these
sets are often written as constraints which determine their members. So, for example, we write $P(v=x)$ for the
probability of the set of all possible worlds of $V$ which assign $x$ to $v$.

\st The following definition captures when a DAG $G$ is an ``ordinary'' (i.e., not-necessarily-causal) Bayesian network
compatible with a given probability measure. The idea is that the graph $G$ captures certain conditional independence
information about the given variables. That is, given information about the observed values of certain variables, the
graph captures which variables are relevant to particular inferences about other variables. Generally speaking, this may fail to reflect the directions of causality, because the laws of probability used to make these inferences (e.g., Bayes
Theorem and the definition of conditional probability) do not distinguish causes from effects. For example if $A$ has a
causal influence on $B$, observations of $A$ may be relevant to inferences about $B$ in much the same way that
observations of $B$ are relevant to inferences about $A$.

\begin{definition} {[Compatible]} \\
{\rm Let $P$ be a probability measure on $V$ and let $G$ be a DAG whose nodes are the variables in $V$. We say that {\em
$P$ is compatible with $G$} if, under $P$, every $v$ in $V$ is independent of its non-descendants in $G$, given its
parents in $G$. } \hfill $\Box$
\end{definition}

\noindent We are now ready to define causal Bayesian networks. In the following definition, $P^*$ is thought of as a mapping from each possible intervention $r$ to the probability measures on $V$ resulting from performing $r$. $P^*$ is intended to capture a model of causal influence in a purely numerical way, and the definition relates this causal model to a DAG $G$.

\st If $G$ is a DAG and $v$ vertex of $G$, let $Parents(G,v)$ denote the parents of $v$ in $G$.
\begin{definition}{[Causal Bayesian network]} \\
{\rm Let $P^*$ map each intervention $r$ in $Interv(V)$ to a probability measure $P_r$ on $V$. Let $G$ be a DAG whose
vertices are precisely the members of $V$. We say that $G$ is a {\em causal Bayesian network} compatible with $P^*$ if
for every intervention $r$ in $Interv(V)$,
\begin{enumerate}
\item $P_r$ is compatible with $G$, \item $P_r(v=x) = 1$ whenever $r(v)=x$, and \item whenever $r$ does not assign a
    value to $v$, and $s$ is an assignment on $Parents(G,v)$ consistent with $r$, we have that for every $x \in
    D(v)$
\end{enumerate}

\st $P_r(v=x\; |\; u=s(u) \mbox{ for all } u \in Parents(G,v))$

$ \hspace{1in} = P_{\{\ \}}(v=x\; | \; u=s(u) \mbox{ for all } u \in Parents(G,v))$ } \hfill $\Box$
\end{definition}

\st Condition 1 says that regardless of which intervention $r$ is performed, $G$ is a Bayesian net compatible with the
resulting probability measure $P^*$.\footnote{This part of the definition captures some intuition about causality. It
entails that given complete information about the factors immediately influencing a variable $v$ (i.e., given the
parents of $v$ in $G$), the only variables relevant to inferences about $v$ are its effects and indirect effects (i.e.,
descendants of $v$ in $G$) --- and that this property holds regardless of the intervention performed.} Condition 2 says
that when we perform an intervention on the variables of $V$, the manipulated variables ``obey'' the intervention.
Condition 3 says that the unmanipulated variables behave under the influence of their parents in the usual way, as if no
manipulation had occurred.

\st For example, consider $V=\{a, d\}$, $D(a) = D(d) = \{true,false\}$, and $P^*$ given by the following table:

\begin{tabular}{l|llll}
  \hline
  intervention & $\{a, d\}$ & $\{a, \neg d\}$ & $\{\neg a, d\}$ & $\{\neg a, \neg d\}$ \\
  \hline
  $\{ \}$ & 0.32 & 0.08 & 0.06 & 0.54 \\
  $\{a \}$ & 0.8 & 0.2 & 0 & 0 \\
  $\{\neg a \}$ & 0 & 0 & 0.01 & 0.99 \\
  $\{d \}$ & 0.4 & 0 & 0.6 & 0 \\
  $\{\neg d \}$ & 0 & 0.4 & 0 & 0.6 \\
  $\{a, d \}$ & 1 & 0 & 0 & 0 \\
  $\{a, \neg d \}$ & 0 & 1 & 0 & 0 \\
  $\{\neg a, d \}$ & 0 & 0 & 1 & 0 \\
  $\{\neg a, \neg d \}$ & 0 & 0 & 0 & 1 \\
  \hline
\end{tabular}

\noindent
The entries down the left margin give possible interventions, and each row defines the corresponding probability measure by giving the probabilities of the four singleton sets of possible worlds. Intuitively, the table represents $P^*$
derived from Example~\ref{rat}, where $a$ represents that the rat eats arsenic, and $d$ represents that it
dies.

\st If $G$ is the graph with a single directed arc from $a$ to $d$, then one can verify that $P^*$ satisfies Conditions
1-3 of the definition of Causal Bayesian Network. For example, if $r = \{a=true\}$, $s = \{d=true\}$, $v = d$, and $x =
true$, we can verify Condition 3 by computing its left and right hand sides using the first two rows of the table:
$$LHS = P_{\{a\}}(d\ | \ a) = 0.8/(0.8 + 0.2) = 0.8$$
$$RHS = P_{\{\ \}}(d \ |\ a) = 0.32/(0.32 + 0.08) = 0.8$$

\st Now let $G^\prime$ be the graph with a single directed arc from $d$ to $a$. We can verify that $P^*$ {\em fails} to
satisfy Condition 3 for $G^\prime$ with $r = \{a=true\}$, $v = d$, $x=true$, and $s$ the empty assignment, viz.,
$$ LHS = P_{\{a\}}(d) = 0.8 + 0 = 0.8$$
$$ RHS = P_{\{ \ \}}(d) = 0.32 + 0.6 = 0.38$$
This tells us that $P^*$ given by the table is not compatible with the hypothesis that the rat's eating arsenic is {\em
caused by} its death.

\st Definition 36 leads to the following proposition that suggests a straightforward algorithm to compute probabilities
with respect to a causal Bayes network with nodes $v_1, \ldots, v_k$, after an intervention $r$ is done.

\begin{proposition}[\cite{pearl99b}]
{\rm Let $G$ be a causal Bayesian network, with nodes $V =v_1= x_1, \ldots ,v_k = x_k$, compatible with an
interventional distribution $P^*$. Suppose also that $r$ is an intervention in $Interv(V)$, and the possible world $v_1
= x_1, \ldots, v_k = x_k$ is consistent with $r$. Then
$$P_r(v_1 = x_1, \ldots, v_k = x_k) = \prod_{i : r(v_i) \mbox{ is
not defined }} P_{\{\ \}}(v_i = x_i | pa_i(r)(x_1, \ldots ,x_k))
$$
\noindent where $pa_i(x_1, \ldots ,x_k))$ is the unique assignment world on $Parents(G,v_i)$ compatible with $ v_1 =
x_1, \ldots, v_k = x_k$.} \hfill $\Box$
\end{proposition}

\begin{theorem}
{\rm Let  $G$ be a DAG with vertices $V=\{v_1, \ldots, v_k\}$ and P$^*$ be as defined in Definition 36. For an intervention $r$, let $do(r)$ denote the set $\{do(v_i = r(v_i)): r(v_i)$ is defined $\}$.

Then there exists a P-log program $\pi$ with random attributes $v_1, \ldots, v_k$ such that for any intervention $r$ in
$Interv(V)$ and any assignment $v_1 = x_1, \ldots, v_k = x_k$ we have

\begin{equation}\label{beq1}
P_r(v_1 = x_1, \ldots, v_k = x_k) = P_{\pi \cup do(r)}(v_1 = x_1, \ldots, v_k =
x_k) \hspace*{1in}
\end{equation} } \hfill $\Box$
\end{theorem}

\noindent {\bf Proof:} We will first give a road map of the proof. Our proof consists of the following four steps.

\medskip \noindent
{\bf (i)} First, given the antecedent in the statement of the theorem, we will construct a P-log program $\pi$ which, as we will ultimately show, satisfies (\ref{beq1}).

\medskip \noindent
{\bf (ii)} Next, we will construct a P-log program $\pi(r)$ and show that:

\begin{equation} \label{beq2}
P_{\pi \cup do(r)}(v_1 = x_1, \ldots, v_k = x_k) = P_{\pi(r)}(v_1 =
x_1, \ldots, v_k = x_k)
\end{equation}

\medskip \noindent
{\bf (iii)} Next, we will construct a finite Bayes net $G(r)$ that defines a probability distribution $P'$ and show
that:

\begin{equation} \label{beq3}
P_{\pi(r)}(v_1 = x_1, \ldots, v_k = x_k) = P'(v_1 = x_1, \ldots,
v_k = x_k)
\end{equation}

\medskip \noindent
{\bf (iv)} Then we will use Proposition 1 to argue that:

\begin{equation} \label{beq4}
P'(v_1 = x_1, \ldots, v_k = x_k) = P_r(v_1 = x_1, \ldots, v_k =
x_k)
\end{equation}

\noindent (\ref{beq1}) then follows from (\ref{beq2}), (\ref{beq3}) and (\ref{beq4}).

\medskip \noindent We now elaborate on the steps (i)-(iv).

\medskip \noindent \underline{\bf Step (i)}
Given the antecedent in the statement of the theorem, we will construct a P-log program $\pi$ as follows:

\medskip \noindent
(a) For each variable $v_i$ in $V$, $\pi$ contains:

\medskip \noindent
$random(v_i)$.\\ $v_i : D(v_i)$.

\medskip \noindent
where $D(v_i)$ is the domain of $v_i$.

\medskip \noindent
(b) For any $v_i \in V$, such that $parents(G,v_i) = \{v_{i_1}, \ldots, v_{i_m}\}$, any $y \in D(v_i)$, and any
$x_{i_1}, \ldots, x_{i_m}$ in $D(v_{i_1}), \ldots, D(v_{i_m})$ respectively, $\pi$ contains the pr-atom:

\medskip \noindent
$pr(v_i$=$y \ |_c \ v_{i_1} $=$ x_{i_1}, \ldots v_{i_m} $=$ x_{i_m} ) = P_{\{\ \}}(v_i$=$y | v_{i_1} $=$ x_{i_1}, \ldots
v_{i_m} $=$ x_{i_m} ).$

\medskip \noindent \underline{\bf Step (ii)}
Given the antecedent in the statement of the theorem, and an intervention $r$ in $Interv(V)$ we will now construct a
P-log program $\pi(r)$ and show that (\ref{beq2}) is true.

\medskip \noindent
(a) For each variable $v_i$ in $V$, if $r(v_i)$ is not defined, then $\pi(r)$ contains $random(v_i)$ and $v_i : D(v_i)$, where $D(v_i)$ is the domain of $v_i$.

\medskip \noindent
(b) The pr-atoms in $\pi(r)$ are as follows. For any node $v_i$ such that $r(v_i)$ is not defined let $\{v_{i_{j_1}},
\ldots, v_{i_{j_k}} \}$ consists of all elements of $parents(G, v_i) = \{v_{i_1}, \ldots, v_{i_m} \}$ where $r$ is not
defined. Then the following pr-atom is in $\pi(r)$.

\medskip \noindent
$p(v_i = x \ | \ v_{i_{j_1}} = y_{i_{j_1}}, \ldots, v_{i_{j_k}} = y_{i_{j_k}}) = P_{\{\ \}}(v_i = x \ | \ v_{i_1} =
y_{i_1}, \ldots, v_{i_m} = y_{i_m}).$, where for all $v_{i_p} \in parents(G, v_i)$, if $r(v_{i_p})$ is defined then
$y_{i_p} = r(v_{i_p})$.

\medskip \noindent Now let us compare the P-log programs
$\pi \cup do(r)$ and $\pi(r)$. Their pr-atoms differ. In addition, for a variable $v_i$, if $r(v_i)$ is defined then
$\pi \cup do(r)$ has $do(v_i = r(v_i))$ and $random(v_i)$ while $\pi(r)$ has neither. For variables, $v_j$, where
$r(v_j)$ is not defined both $\pi \cup do(r)$ and $\pi(r)$ have $random(v_i)$. It is easy to see that there is a
one-to-one correspondence between possible worlds of $\pi \cup do(r)$ and $\pi(r)$; for any possible world $W$ of $\pi
\cup do(r)$ the corresponding possible world $W'$ for $\pi(r)$ can be obtained by projecting on the atoms about
variables $v_j$ for which $r(v_j)$ is not defined. For a $v_i$ for which $r(v_i)$ is defined, $W$ will contain
$intervene(v_i)$, and will not have an assigned probability. The default probability $PD(W, v_i = r(v_i))$ will be
$\frac{1}{|D(v_i)|}$. Now it is easy to see that the unnormalized probability measure associated with $W$ will be
$$\prod_{v_i \ : \ r(v_i) \mbox{ is defined }} \frac{1}{|D(v_i)|}$$ times the unnormalized probability measure
associated with $W'$ and hence their normalized probability measures will be the same. Thus $P_{\pi \cup do(r)}(v_1 =
x_1, \ldots, v_k = x_k) = P_{\pi(r)}(v_1 = x_1, \ldots, v_k = x_k)$.

\medskip \noindent \underline{\bf Step (iii)} Given $G$, $P^*$ and any
intervention $r$ in $Interv(V)$ we will construct a finite Bayes net $G(r)$. Let $P'$ denote the probability with
respect to this Bayes net.

\medskip \noindent The nodes and edges of $G(r)$ are as follows.
All vertices $v_i$ in $G$ such that $r(v_i)$ is not defined are the only vertices in $G(r)$. For any edge from $v_i$ to
$v_j$ in $G$, only if $r(v_j)$ is not defined the edge from $v_i$ to $v_j$ is also an edge in $G(r)$. No other edges are in $G(r)$. The conditional probability associated with the Bayes net $G(r)$ is as follows: For any node $v_i$ of $G(r)$, let $parents(G(r), v_i) = \{v_{i_{j_1}}, \ldots, v_{i_{j_k}} \} \subseteq parents(G, v_i) = \{v_{i_1}, \ldots, v_{i_m}
\}$. We define the conditional probability $p(v_i = x \ | \ v_{i_{j_1}} = y_{i_{j_1}}, \ldots, v_{i_{j_k}} =
y_{i_{j_k}}) = P_{\{\ \}}(v_i = x \ | \ v_{i_1} = y_{i_1}, \ldots, v_{i_m} = y_{i_m})$, where for all $v_{i_p} \in
parents(G, v_i)$, if $r(v_{i_p})$ is defined (i.e., $v_{i_p} \not \in parents(G(r), v_i)$) then $y_{i_p} = r(v_{i_p})$.

\medskip \noindent
From Theorem 2 which shows the equivalence between a Bayes net and a representation of it in P-log, which we will denote by $\pi(G(r))$ , we know that $P'(v_1 = x_1, \ldots, v_k = x_k) = P_{\pi(G(r))}(v_1 = x_1, \ldots, v_k = x_k)$. It is
easy to see that $\pi(G(r))$ is same as $\pi(r)$. Hence (\ref{beq3}) holds.

\medskip \noindent \underline{\bf Step (iv)} It is easy to see that $P'(v_1 = x_1, \ldots, v_k = x_k)$ is equal to the right hand side of Proposition 1. Hence (\ref{beq4}) holds.

%\st INSERT

\section{Appendix III: Semantics of ASP}\label{app3}
In this section we review the semantics of ASP. Recall that an ASP \emph{rule} is a statement of the form
\begin{equation}\label{asp-rule}
l_0 \oor \dots \oor l_k \leftarrow l_{k+1}, \dots ,l_m,\no l_{m+1}, \dots, \no l_n
\end{equation}
where the $l_i$'s are ground literals over some signature $\Sigma$. An ASP \emph{program}, $\Pi$, is a collection of
such rules over some signature $\sigma(\Pi)$, and a \emph{partial interpretation} of $\sigma(\Pi)$ is a consistent set
of ground literals of the signature. A program with variables is considered shorthand for the set of all ground
instantiations of its rules. The answer set semantics of a logic program $\Pi$ assigns to $\Pi$ a collection of
\emph{answer sets} --- each of which is a partial interpretation of $\sigma(\Pi)$ corresponding to some possible set of
beliefs which can be built by a rational reasoner on the basis of rules of $\Pi$. As mentioned in the introduction, in
the construction of such a set, $S$, the reasoner should satisfy the rules of $\Pi$ and adhere to the \emph{rationality
principle}\index{rationality principle} which says that \emph{one shall not believe anything one is not forced to
believe}. A partial interpretation $S$  \emph{satisfies} Rule \ref{asp-rule} if whenever $l_{k+1},\dots,l_{m}$ are in
$S$ and none of $l_{m+1},\dots,l_{n}$ are in $S$, the set $S$ contains at least one $l_i$ where $0 \leq i \leq k$. The
definition of an answer set of a logic program is given in two steps:

\st First we consider a program  $\Pi$  not containing default negation $not$.
\begin{definition}\label{as1}{(Answer set -- part one)}\\
{\rm A partial interpretation $S$ of the signature $\sigma(\Pi)$ of $\Pi$  is an \emph{answer set} for $\Pi$ if $S$ is minimal (in the sense of set-theoretic inclusion) among the partial interpretations of $\sigma(\Pi)$ satisfying the rules of $\Pi$.} \hfill $\Box$
\end{definition}

\noindent
The rationality principle is captured in this definition by the minimality requirement.

\st To extend the definition of answer sets to arbitrary programs, take any program $\Pi$, and let $S$ be a partial
interpretation of $\sigma(\Pi)$. The \emph{reduct} $\Pi^S$ of $\Pi$ relative to $S$ is obtained by

\begin{enumerate}
\item removing from $\Pi$ all rules containing $\no l$ such that $l \in S$, and then

\item removing all literals of the form $\no l$ from the remaining rules.
\end{enumerate}

\st Thus $\Pi^S$ is a program without default negation.
\begin{definition}\label{as2}{(Answer set -- part two)}\\
{\rm
A partial interpretation $S$ of $\sigma(\Pi)$ is an answer set for $\Pi$ if $S$ is an answer set for $\Pi^S$.}
\hfill $\Box$
\end{definition}

\noindent
The relationship between this fix-point definition and the informal principles which form the basis for the notion of
answer set is given by the following proposition.

\begin{proposition}\label{msl}{Baral and Gelfond, \cite{bg94a}}\\
{\rm Let $S$ be an answer set of ASP program $\Pi$.\\ (a) $S$ satisfies the rules of the ground instantiation of $\Pi$.\\ (b) If literal $l \in S$ then there is a rule $r$ from the ground instantiation of $\Pi$ such that the body of $r$ is satisfied by $S$ and $l$ is the only literal in the head of $r$ satisfied by $S$.} \hfill $\Box$
\end{proposition}

\noindent
The rule $r$ from (b) ``forces'' the reasoner to believe $l$.

\st It is easy to check that program $p(a) \oor p(b)$ has two answer sets, $\{p(a)\}$ and $\{p(b)\}$, and program $p(a)
\leftarrow \no p(b)$ has one answer set, $\{p(a)\}$. Program $P_1$ from the introduction indeed has one answer set
$\{p(a),\neg p(b), q(c)\}$, while program $P_2$ has two answer sets, $\{p(a),\neg p(b), p(c),\neg q(c)\}$ and
$\{p(a),\neg p(b), \neg p(c),\neg q(c)\}$.

\st Note that the left-hand side (the head) of an ASP rule can be empty. In this case the rule is often referred to as a
\emph{constraint} or \emph{denial}. The denial $\leftarrow B$ prohibits the agent associated with the program from
having a set of beliefs satisfying $B$. For instance, program $p(a) \oor \neg p(a)$ has two answer sets, $\{p(a)\}$ and
$\{\neg p(a)\}$. The addition of a denial $\leftarrow p(a)$ eliminates the former; $\{\neg p(a)\}$ is the only answer
set of the remaining program. Every answer set of a consistent program $\Pi \cup \{l.\}$ contains $l$ while a program
$\Pi \cup \{\leftarrow \no l.\}$ may be inconsistent. While the former tells the reasoner to believe that $l$ is true
the latter requires him to find support of his belief in $l$ from $\Pi$. If, say, $\Pi$ is empty then the first program
has the answer set $\{l\}$ while the second has no answer sets. If $\Pi$ consists of the default $\neg l \leftarrow \no
l$ then the first program has the answer set $l$ while the second again has no answer sets.

\st Some additional insight into the difference between $l$ and $\leftarrow \no l$ can also be obtained from the
relationship between ASP  and intuitionistic or constructive logic \cite{fer05e} which distinguishes between $l$ and $\neg \neg l$. In the corresponding mapping the denial corresponds to the double negation of $l$.

\st To better understand the role of denials in ASP one can view a program $\Pi$ as divided into two parts: $\Pi_r$
consisting of rules with non-empty heads and $\Pi_d$ consisting of the denials of $\Pi$. One can show that $S$ is an
answer set of $\Pi$ iff it is an answer set of $\Pi_r$ which satisfies all the denials from $\Pi_d$. This property is
often exploited in answer set programming where the initial knowledge about the domain is often defined by $\Pi_r$ and
the corresponding computational problem is posed as the task of finding answer sets of $\Pi_r$ satisfying the denials
from $\Pi_d$.

%\bibliography{C:/chitta/bib/gelfond-home}

\end{document}